\documentclass[twoside]{article}

\usepackage[accepted]{aistats2023}
\usepackage{defns_learnlabel}
\usepackage{amssymb,amsmath,amsthm,amsfonts,latexsym}
\usepackage{amsmath,graphicx,bm,url}
\usepackage[dvipsnames]{xcolor}
\usepackage{algorithm}
\usepackage{mathrsfs, bbm}
\usepackage{dsfont}

\usepackage{algpseudocode}
\usepackage{nicefrac}
\usepackage{enumitem}
\usepackage{hyperref}
\hypersetup{
    colorlinks = true,
    citecolor = blue,
}

\catcode`~=11 \def\UrlSpecials{\do\~{\kern -.15em\lower .7ex\hbox{~}\kern .04em}} \catcode`~=13 

\allowdisplaybreaks[1]

\newcommand{\calA}{\mathcal{A}}

\newcommand{\calE}{\mathcal{E}}

\newcommand{\calG}{\mathcal{G}}
\newcommand{\calH}{\mathcal{H}}

\newcommand{\calN}{\mathcal{N}}
\newcommand{\calO}{\mathcal{O}}
\newcommand{\calP}{\mathcal{P}}

\newcommand{\calX}{\mathcal{X}}

\newcommand{\bx}{\mathbf{x}}

\newcommand{\rmM}{\mathrm{M}}

\newcommand{\bbE}{\mathbb{E}}

\newcommand{\bbN}{\mathbb{N}}

\newcommand{\bbP}{\mathbb{P}}

\newcommand{\bbR}{\mathbb{R}}

\DeclareMathAlphabet{\mathbsf}{OT1}{cmss}{bx}{n}
\DeclareMathAlphabet{\mathssf}{OT1}{cmss}{m}{sl}

\DeclareSymbolFont{bsfletters}{OT1}{cmss}{bx}{n}  
\DeclareSymbolFont{ssfletters}{OT1}{cmss}{m}{n}
\DeclareMathSymbol{\bsfGamma}{0}{bsfletters}{'000}
\DeclareMathSymbol{\ssfGamma}{0}{ssfletters}{'000}
\DeclareMathSymbol{\bsfDelta}{0}{bsfletters}{'001}
\DeclareMathSymbol{\ssfDelta}{0}{ssfletters}{'001}
\DeclareMathSymbol{\bsfTheta}{0}{bsfletters}{'002}
\DeclareMathSymbol{\ssfTheta}{0}{ssfletters}{'002}
\DeclareMathSymbol{\bsfLambda}{0}{bsfletters}{'003}
\DeclareMathSymbol{\ssfLambda}{0}{ssfletters}{'003}
\DeclareMathSymbol{\bsfXi}{0}{bsfletters}{'004}
\DeclareMathSymbol{\ssfXi}{0}{ssfletters}{'004}
\DeclareMathSymbol{\bsfPi}{0}{bsfletters}{'005}
\DeclareMathSymbol{\ssfPi}{0}{ssfletters}{'005}
\DeclareMathSymbol{\bsfSigma}{0}{bsfletters}{'006}
\DeclareMathSymbol{\ssfSigma}{0}{ssfletters}{'006}
\DeclareMathSymbol{\bsfUpsilon}{0}{bsfletters}{'007}
\DeclareMathSymbol{\ssfUpsilon}{0}{ssfletters}{'007}
\DeclareMathSymbol{\bsfPhi}{0}{bsfletters}{'010}
\DeclareMathSymbol{\ssfPhi}{0}{ssfletters}{'010}
\DeclareMathSymbol{\bsfPsi}{0}{bsfletters}{'011}
\DeclareMathSymbol{\ssfPsi}{0}{ssfletters}{'011}
\DeclareMathSymbol{\bsfOmega}{0}{bsfletters}{'012}
\DeclareMathSymbol{\ssfOmega}{0}{ssfletters}{'012}

\newcommand{\tilk}{\tilde{k}}

\newcommand{\tilx}{\tilde{x}}

\newcommand{\defeq}{\stackrel{\cdot}{=}}

\newtheorem{theorem}{Theorem} 
\newtheorem{lemma}[theorem]{Lemma}

\newtheorem{corollary}[theorem]{Corollary}

\newcommand{\qednew}{\nobreak \ifvmode \relax \else
      \ifdim\lastskip<1.5em \hskip-\lastskip
      \hskip1.5em plus0em minus0.5em \fi \nobreak
      \vrule height0.75em width0.5em depth0.25em\fi}

\usepackage[round]{natbib}

\begin{document}

\twocolumn[

\aistatstitle{Active Cost-aware Labeling of Streaming Data}

\aistatsauthor{ Ting Cai \And Kirthevasan Kandasamy }

\aistatsaddress{ University of Wisconsin-Madison \And  University of Wisconsin-Madison } ]

\newcommand{\insertDiscreteAlgorithm}{%
\begin{algorithm}
\caption{Algorithm for $K$ discrete types}\label{alg:mab}
\textbf{Input}: confidence parameter $\delta$, labeling cost $B$ 
\begin{algorithmic}[1]
\For{time $t = 1, 2, \cdots$ }
    \State $x_t$ appears, $M_{t,\xt} \leftarrow M_{t-1, x_t} + 1$
    \State $ \forall k\neq x_t$, $M_{t,k} \leftarrow M_{t-1,k}$
    \If{$\Utmoxt\left(\delta/\Nt^3\right) > B^{\nicefrac{1}{3}} \Mtxt^{\nicefrac{-1}{3}}$}
        \newline
    \vphantom{T}\hspace{0.8in} \Comment{See~\eqref{eqn:Utk} for $\Utk$,~\eqref{eqn:MtkNtk} for
$\Nt$}
    \label{lin:ifthreshcondn}
        \State Label $x_t$ and observe $y_t$
        \State $\Ntxt \leftarrow \Nttkk{t-1}{\xt} +1$
        \State Update $\muhattxt, \Utmoxt$ according to~\eqref{eqn:muhat},~\eqref{eqn:Utk}.
        \State $\forall k\neq \xt$,
        $\;\Ntk \leftarrow \Nttkk{t-1}{k}$,
        $\;\;\widehat{\mu}_{t,k} \leftarrow \widehat{\mu}_{t-1, k}$, \newline
        \vphantom{T} \hspace{0.9in} $\;\Utk\leftarrow \Utmok$.
        \label{lin:updates}
    \Else
        \State $\forall k$,
        $\;\Ntk \leftarrow \Nttkk{t-1}{k}$,
        $\;\;\widehat{\mu}_{t,k} \leftarrow \widehat{\mu}_{t-1, k}$, \newline
        \vphantom{T} \hspace{0.6in} $\;\Utk\leftarrow \Utmok$.
    \EndIf
    \State Output prediction $p_t \leftarrow \widehat{\mu}_{t, x_t}$
    \label{lin:predict}
\EndFor
\end{algorithmic}
\end{algorithm}
}

\newcommand{\insertGPAlgorithm}{
\begin{algorithm}
\caption{Algorithm for $f$ in an RKHS}\label{alg:gpucb}
\textbf{Input}: GP Prior $\mu_0$, $\kappa_0$,  labeling cost $B$
\begin{algorithmic}[1]
\For{time $t = 1, 2, \cdots, T$ }
    \If{$\sigma_{t-1}(x_t)> \tau(t)$}
    \label{lin:gpifthreshcondn}
        \State Label $x_t$. Observe $y_t = f(x_t) + \epsilon_t$
        \State Update $\mu_{t}$ and $\sigma_{t}$ with $(\xt, \yt)$.
            \Comment{See~\eqref{eqn:gp_update_sigma}}
    \Else
        \State $\mu_{t} \leftarrow \mu_{t-1}$, \quad $\sigma_t \leftarrow \sigma_{t-1}$
    \EndIf
    \State Output $ p_t \gets \mu_{t}(x_t)$
    \label{lin:gppredict}
    
\EndFor
\end{algorithmic}
\end{algorithm}
}

\newcommand{\insertFigIllus}{
\begin{figure}
    \centering
    \includegraphics[width = 0.5\columnwidth]{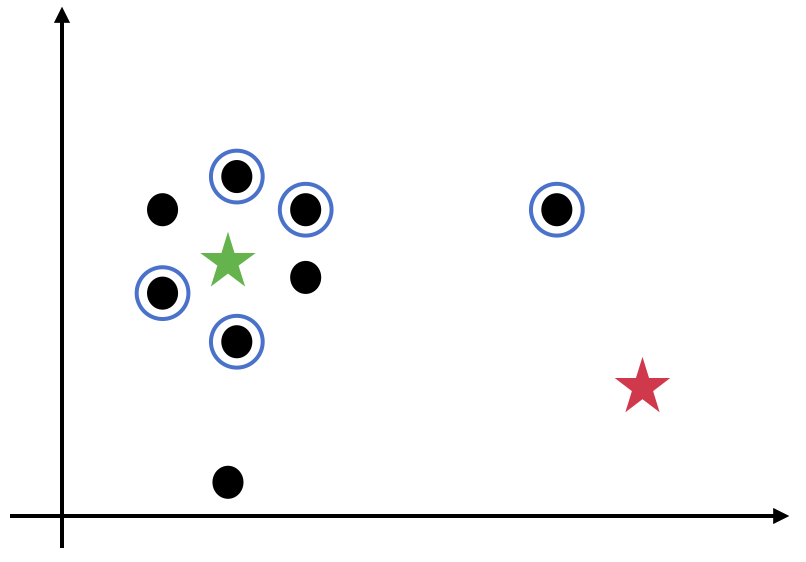}
\vspace{-0.1in}
    \caption{\small
An illustration of the problem set up when the inputs $\{\xt\}_t$ are in 2D.
Each black dot represents a data point that arrived in previous rounds.
If it is circled in blue, it indicates that it was chosen for labeling.
If the current input $\xt$ is similar to previously labeled inputs (e.g green star),
the algorithm may
choose to forego labeling and predict $f(\xt)$ based on these points to reduce labeling costs.
If it is dissimilar to previously labeled points (e.g red star),
it may choose to label it to improve the prediction.
}
\vspace{-0.15in}
    \label{fig:labelillus}
\end{figure}
}

\begin{abstract}
\vspace{-0.1in}
We study actively labeling streaming data, where an active learner is faced with a stream of data
points and must carefully choose which of these points to label via an expensive experiment.
Such problems frequently arise in applications such as healthcare and astronomy. We first study
a setting when the data's inputs belong to one
of  $K$ discrete distributions and
formalize this problem via a loss that captures the labeling cost and the prediction error.
When the labeling cost is $B$,
our algorithm, which chooses to label a point if the uncertainty is larger than a time
and cost dependent
threshold, achieves a
worst-case upper bound of $\tilde{\calO}(B^{\frac{1}{3}} K^{\frac{1}{3}} T^{\frac{2}{3}})$ on the loss after
$T$ rounds.
We also provide a more nuanced upper bound which demonstrates that the algorithm
can adapt to the arrival pattern, and
achieves better performance when the arrival pattern is more favorable.
We complement both upper bounds with matching lower bounds.
We next study this problem when the
inputs belong to a continuous domain and the output of the experiment is a smooth function
with bounded RKHS norm.
After $T$ rounds in $d$ dimensions, we show that the loss is bounded by $\tilde{\calO}(B^{\frac{1}{d+3}}
T^{\frac{d+2}{d+3}})$ in an RKHS with
a squared exponential kernel and by $\tilde{\calO}(B^{\frac{1}{2d+3}} T^{\frac{2d+2}{2d+3}})$ in an RKHS with a
Mat\'ern kernel.
Our empirical evaluation demonstrates that our method outperforms other baselines
in several synthetic experiments and two real experiments in medicine and astronomy.
\end{abstract}

\section{Introduction}
\label{sec:intro}

The success of real-world supervised learning methods, where we wish to learn a mapping from
inputs to outputs, often relies on the existence of abundant labeled training data, i.e
the outputs are known for the given inputs.
However, labeling data can be expensive, as it may require costly experimentation or human
effort.
The vast literature on active learning has focused on methods for reducing the labeling complexity
in such use cases.
The majority of such papers study \emph{pool-based active learning}, where there is
either a finite or infinite pool of data, and an algorithm should choose which samples from this
pool to label in order to learn the mapping well.
These methods have been applied successfully in real-world problems, such as text classification,
information extraction, materials discovery, astrophysics,
and agriculture~\citep{DBLP:conf/icml/McCallumN98,
DBLP:journals/tnn/Wu19, chandrasekaran2020exploring, kusne2020fly, yang2022dissimilarity,
kandasamy2017query}.

\insertFigIllus

\emph{Streaming} settings for active learning, where an algorithm should choose which data points to
label from a possibly infinite stream of data, has received comparatively less attention.
Unlike in pool-based settings, where a learner can choose any sample from this pool at any point for labeling,
here the learner goes through the data sequentially and has to make decisions to label it on a
per-instance basis.

In this work, we consider the following setting where our goal is to learn an underlying function $f$ from inputs to outputs.
On each round $t$, the algorithm receives a data's input $x_t$.
It may choose to observe a noisy label 
$\yt\in\RR$ for this point by incurring a
\emph{labeling cost}, or not
observe this label at no cost.
In either case, it must output a prediction $\pt$ at the end of the round.
The algorithm incurs a \emph{faulty prediction cost} proportional to $|f(x_t) - \pt|$ for choosing a
prediction that is far away from the true function.
The goal of the algorithm is to learn the mapping sufficiently well to minimize the sum of both types of costs
over a period of time.
If we label on each round we incur a large labeling cost; however, not labeling
a sufficient number of points incurs a large faulty prediction cost as we will not be able to estimate $f$ well.
To solve this problem effectively,
an algorithm should leverage information gained from past experiments to estimate the
cost of faulty prediction, and assess this against the cost of labeling to decide if
$\xt$ needs to be labeled.
We have illustrated this in Fig.~\ref{fig:labelillus}.

This problem is motivated by applications in medicine and healthcare.
For example, performing an expensive medical exam to assess the risk of a disease for a
patient can be costly.
In such cases, instead of performing this exam on all patients,
 a doctor may wish to quickly screen a patient based on their features,
past medical history, and cheap medical tests, to determine if an expensive exam is necessary.
In addition to saving costs,
this will also ensure that valuable medical resources are used prudently, reducing unnecessary
waiting times for patients to achieve efficiency. Other applications include reducing the experimentation costs in astronomy, such as when observing streams of astronomical events (e.g supernovae, comets),
and reducing investigation costs in finance, such as
when streams of customers are applying for large loans.

We study this problem in two different settings:
firstly, when each data point belongs to one of $K$ discrete types;
secondly, when $x_t\in\RR^d$ and $f$ belongs to a reproducing kernel Hilbert space (RKHS)
with bounded RKHS norm.
In both cases, we formalize this problem via a loss function capturing both costs at each time and describe a simple and efficient algorithm that chooses to label if
the uncertainty of $f(x_t)$ is larger than a threshold determined by the labeling cost
and the past arrival pattern.
\textbf{Our contributions} are:
\begin{itemize}[leftmargin=5mm]
\item First, in~\S\ref{sec:kdiscrete},
we study this problem when the input belongs to one of $K$ discrete types.
Algorithm~\ref{alg:mab} achieves a
$\tilde{\calO}(B^{\nicefrac{1}{3}}\sum_{t=1}^T M_{T,k}^{\nicefrac{2}{3}})$ upper bound on the loss
where $B$ is the cost of labeling and $M_{T,k}$ is the total number of arrivals for type $k$
after $T$ rounds.
We complement this upper bound with a matching (up to log factors) lower bound.
Under the worst possible arrival pattern, the upper and lower bounds become
$\Theta(B^{\nicefrac{1}{3}}K^{\nicefrac{1}{3}}T^{\nicefrac{2}{3}})$.

\item Second, in~\S\ref{sec:gp}, to model real-world use cases, we study the setting where $x_t\in\RR^d$
and the underlying function $f$ is in an RKHS.
Algorithm~\ref{alg:gpucb}, which is similar in spirit to Algorithm~\ref{alg:mab},
achieves a
$\tilde{\calO}(B^{\frac{1}{d+3}} T^{\frac{d+2}{d+3}})$ worst-case upper bound on the loss when the RKHS has a squared exponential kernel and a
$\tilde{\calO}(B^{\frac{1}{2d+3}} T^{\frac{2d+2}{2d+3}})$ worst-case upper bound when the RKHS has a Mat\'ern kernel. 

\item Third, in~\S\ref{sec:experiments}, we evaluate both algorithms empirically on several
synthetic examples and two real-world use cases in medicine and astronomy.
Our methods outperform baseline methods
under
different arrival patterns of input data points.
When compared to the next best baseline, 
our method achieves an average loss that is lower by a factor
of  $\sim$27\% in the medical application
and $\sim$57\% in the astronomical application. 
\end{itemize}
\vspace{-0.1in}

\subsection*{Related work}
The majority of the work in active learning studies the
\emph{pool-based} setting where a learner can choose to label a data point at any time in the
learning process.
In addition to the  works which focus on applications for the pool-based setting of active
learning~\citep{DBLP:conf/icml/McCallumN98,settles2009active,%
DBLP:journals/tnn/Wu19,chandrasekaran2020exploring,kusne2020fly,yang2022dissimilarity},
there is a rich body of theoretical work~\citep{zhang2015active, chaudhuri2015convergence, balcan2013statistical, balcan2010true, hanneke2019surrogate, hanneke2015minimax, hanneke2011rates, GentileWZ22}.
However, the \emph{streaming} setting, which we consider in this paper, is more challenging.
Since if we choose not to label the data point when it arrives, we may
never be able to label that point again in the future.

The majority of the works in the \emph{streaming} setting focus on applications.
\citet{DBLP:journals/tnn/ZliobaiteBPH14} explored active learning for streaming data when the
distribution of the incoming data shifts over time.
They choose to label each data point when the uncertainty is below a threshold,
but unlike us, this threshold does not depend on the labeling cost.
Moreover, their threshold is chosen in an ad-hoc manner while our threshold is chosen based on theoretical analysis.
\citet{lindstrom2010handling} considered a batch version of this problem when the incoming data
shifts with time. 
They group a stream of unlabelled data into batches and feed each batch into a classifier to select
a subset for labeling.
\citet{zhu2007active} studied a similar setting where they build a classifier on a subset of a data batch randomly and use uncertainty sampling, 
which queries the data that the learner is most uncertain about, to label more instances within this
batch. Our setting is different from these works
since we don't assume data comes in batches. 

Among works with theoretical guarantees in the streaming setting,
\citet{wang2021neural, ban2022improved, desalvo2021online} studied active learning to solve binary or multiple classification tasks. \citet{chu2011unbiased} studied a similar setting 
while focusing on unbiased online active learning. In addition to the technical and algorithmic differences, our work focuses more on regression problems and accounts for the cost of labeling.
\citet{werner2022online} aimed at finding a subset of the streaming data whose value is close to the same-sized subset that has the optimal value.
One important difference in our work,
when compared to both the applied and theoretical work above, is that
 we explicitly model the cost of labeling in our formalism and algorithm.
Moreover,
unlike the above theoretical work, our formalism is based on a loss term that encapsulates both
the labeling cost and prediction errors; the learner's goal is
to balance both effects out to minimize the loss.

There are works studying online learning and active learning under budget constraints~\citep{donmez2008proactive,vijayanarasimhan2010far,badanidiyuru2012learning}.
This is different from our formalism since we model the cost of labeling relative to
the prediction error without any budget and aim to minimize a loss composed of both effects.
In particular, our setting allows the learner to label more (possibly
infinite) points
when there is a benefit to doing so over an infinite time horizon.

Our method in the RKHS setting builds on a line of work on Bayesian active learning and adaptive decision-making
which use Gaussian processes (GPs).
In this flavor, there are several works specifically focused on applications such as large-scale
classification, preference learning and posterior approximation~\citep{pinsler2019bayesian,
houlsby2011bayesian, kandasamy2017query}.
Moreover, \citet{garnett2013active} study active learning for discovering
low-rank structure in high-dimensional GP. 
\citet{bui2017streaming} and~\citet{terry2020splitting} use GPs for approximation and regression with streaming data.
On the theoretical side, 
\citet{chen2017near} consider learning an unknown variable through a sequence of noisy tests and propose an algorithm to choose the test which maximizes the gain in a surrogate objective with a theoretical guarantee.
\citet{javdani2014near} consider a similar problem as~\cite{chen2017near} when the hypothesis tests
may have correlated regions.
The seminal work of~\citet{DBLP:conf/icml/SrinivasKKS10} provided the
first theoretical results for smooth bandits using GPs.
Our theoretical analysis in the RKHS setting uses some ideas from this paper.

\section{$K$ Discrete Types}
\label{sec:kdiscrete}

We will first study this problem in a setting where the input $\xt$ belongs to one of
$K$ discrete types $\{1,\dots, K\} \defeq [K]$
 (i.e similar to $K$ arms in $K$-armed bandits).
This simple setting will help us develop key intuitions and optimality properties
which we will then use to design an algorithm under more realistic assumptions in~\S\ref{sec:gp}.
On each round $t$, an \emph{input} data point $x_t\in[K]$ arrives.
Each type $k\in[K]$ is associated with a $\sigma$--subGaussian distribution with mean $\mu_k\in[0,1]$.
If the learner chooses to label $x_t$, they will observe a label $\yt\in\RR$ which is drawn
from this distribution. Hence, $\bbE[\yt] = \mu_{\xt}$.
We can therefore think of the function $f$ as a mapping from $[K]$ to $[0,1]$,
where $f(k) = \mu_k$, although we prefer the notation $\mu_k$ in this section.

At the end of round $t$, the learner should output a prediction $\pt$ for $\mu_{\xt}$.
The learner's \emph{faulty prediction cost} is simply the difference between the prediction
and the true mean of $x_t$, i.e $|\pt - \mu_{x_t}|$.
If the learner chooses to label a point to reduce the prediction cost, they incur a
\emph{labeling cost} $B$ (relative to $|\pt - \mu_{x_t}|$).
Therefore, the \emph{total loss} $L_T$ after $T$ rounds can be written as,
\begin{align*}
    L_T &= \sum_{t=1}^T \left(\;
    B \cdot \mathds{1}_{\{x_t \mbox{ is labeled}\}} + |\mu_{x_t} - p_t|
    \;\right) \\
    &= BN_T + \sum_{t=1}^T |\mu_{x_t} - p_t|,
    \numberthis
    \label{eqn:mabloss}
\end{align*}
where $N_T$ is the total number of points labeled after $T$ rounds.
To obtain sublinear loss, a learner should balance the cost of labeling against the faulty
prediction cost. If the learner chooses to label too frequently, the latter might be reduced,
but risks incurring a large labeling cost.

We will make no explicit assumption about the arrival pattern,
i.e order of arrival of $\xt$'s on each round.
They can even be chosen by an adversary.
Our theoretical results will show that our algorithm can naturally adapt to the difficulty
of the arrival pattern.
In practical applications,
the exact value for $B$ may be chosen based on the monetary cost of a labeling experiment,
the opportunity cost for using valuable experimental resources, and the risk of adverse
consequences if the true mean is not estimated correctly.

\subsection{Algorithm}
\label{sec:kdiscretealgo}

Our algorithm for this problem is outlined in Algorithm~\ref{alg:mab}.
On each round $t$, We will maintain an estimate $\muhattk$ for each mean $\muk$ along with confidence intervals.
If the uncertainty for $\muk$ as measured by the confidence interval is larger than a
certain threshold, we will label $x_t$; otherwise, we will not label it.
The exact value of the threshold depends on the number of times type $k$ has arrived 
and the cost $B$ of labeling.
As we will show in \S\ref{sec:kdiscretethm},
this algorithm, while simple and intuitive, is optimal for this problem.

To describe the algorithm formally, we will let $\labelt=1$ if we chose to label $\xt$ on round
$t$ and let  $\labelt=0$ otherwise.
Next,
$\Mtk$ and $\Ntk$ respectively will denote the number of times type $k$ arrived in $t$ rounds and
the number of times type $k$ arrived and was labeled.
Finally, let $\Nt$ denote the total number of rounds we labeled.
We have:
\begin{align*}
&\Mtk = \sum_{s=1}^t \mathds{1}_{\{\xs = k\}},
\numberthis
\label{eqn:MtkNtk}
\\
&\Ntk = \sum_{s=1}^t \mathds{1}_{\{\xs = k, \;\labels=1\}},
\hspace{0.2in}
\Nt = \sum_{k\in[K]} \Ntk.
\vspace{-0.05in}
\end{align*}
Next, for each $k\in[K]$, let 
$\muhattk$ denote the sample mean for type $k$ using samples obtained via the first $t$ rounds:

\begingroup
\allowdisplaybreaks
\begin{align*}
    \muhattk &= \frac{1}{N_{t,k}}\sum_{s=1}^t y_t \cdot \mathds{1}_{\{\xs = k, \labels=1\}}
\numberthis
\label{eqn:muhat}
\vspace{-0.05in}
\end{align*}
\endgroup
Using sub-Gaussian concentration, we can quantify the uncertainty of
this estimate via $\Utk$ defined below.
That is, $(\muhattk-\Utk(\delta), \muhattk+\Utk(\delta))$ is a confidence interval
which traps $\muk$ with probability at least $1-\delta$.
We have:
\begingroup
\allowdisplaybreaks
\begin{align}
    \Utk(\delta) &= \begin{cases}
     \;\sqrt{{2\sigma^2\log (2/\delta)}/\Ntk} \quad &\text{if $\Ntk > 0$}
        \\
      \;\infty &\text{otherwise}
    \end{cases}
\label{eqn:Utk}
\vspace{-0.05in}
\end{align}
\endgroup

We can now describe our algorithm.
When $\xt$ arrives, we choose to label it if the uncertainty $\Utk(\delta)$ is larger than the threshold
value $B^{\nicefrac{1}{3}} \Mtxt^{\nicefrac{-1}{3}}$
(line~\ref{lin:ifthreshcondn}),
which depends on the labeling cost $B$ and the number of times
$\xt$ has arrived (not observed) $\Mtxt$.
The threshold increases with $B$, discouraging frequent labeling when it is expensive,
and decreases with $\Mtxt$, encouraging frequent labeling the more times a type is observed to
reduce accumulating faulty prediction costs.
If we choose to label, we update the mean and confidence intervals for $\xt$ (line~\ref{lin:updates}).
Finally, we output the sample mean $\muhattxt$ as our prediction $\pt$ (line~\ref{lin:predict}). 

\insertDiscreteAlgorithm

\subsection{Theoretical Results}
\label{sec:kdiscretethm}

We now present our theoretical contributions.
We defer all proofs to Appendix~\S\ref{ax:proofs}, but
 outline the main proof intuitions at the end of this section.

\paragraph{Upper bounds:}
To state our first result,
recall from~\eqref{eqn:MtkNtk} that $\MTk$ denotes the number of arrivals of type
$k\in[K]$ in $T$ rounds.
Theorem~\ref{thm:mab} upper bounds the loss for Algorithm~\ref{alg:mab}
in terms of these $\MTk$ values.

\vspace{0.1in}
\begin{theorem}[Arrival-dependent Upper Bound] \label{thm:mab}
Assume $\muk\in[0,1]$ for all $k\in[K]$ and let $\delta, B$ be given.
Then, for Algorithm~\ref{alg:mab}, 
$L_T \in \tilde{\calO}\left(\sum_{k=1}^K B^{\frac{1}{3}}M_{T,k}^{\frac{2}{3}} \right)$.
Precisely, with probability at least $1-\delta\pi^2/12$, $\forall T \ge 1$,
\begin{align*}
    L_T\le \sum_{k=1}^K \left((B^{\frac{1}{3}}+B^{-\frac{2}{3}}) 2\sigma^2 \log\frac{2T^3}{\delta}
M_{T,k}^{\frac{2}{3}} + \frac{3}{2}  B^{\frac{1}{3}} M_{T,k}^{\frac{2}{3}} \right).
\end{align*}
\end{theorem}

The corollary below, which follows from Theorem~\ref{thm:mab}, states the 
upper bound under the worst possible arrival pattern for Algorithm~\ref{alg:mab}.
It follows via an application of the power mean inequality
and noting that $M_{T,1}+\cdots+M_{T,k} = T$.
\begin{align*}
    \Bigg(\frac{M_{T,1}^{\frac{2}{3}}+\cdots + M_{T,K}^{\frac{2}{3}}}{K}\Bigg)^{\frac{3}{2}} \le
\frac{M_{T,1}+\cdots+M_{T,k}}{K}
= \frac{T}{K}.
\end{align*}

\vspace{0.1in}
\begin{corollary}[Worst-case Upper Bound]
\label{col:worstupper}
Under the assumptions of Theorem~\ref{thm:mab},
 $L_T \in \tilde{\calO}\big(B^{\frac{1}{3}}T^{\frac{2}{3}} K^{\frac{1}{3}}\big)$
for Algorithm~\ref{alg:mab}.
\end{corollary}

To interpret these results,
observe that while the worst-case bound scales with $K$,
our algorithm can naturally adapt to the arrival pattern of the types.
In particular,
when the arrivals are such that
only  a few types appear frequently, the bound is significantly better.
For instance, in the easiest case, when only
one type $k\in[K]$ appears on all rounds, we obtain $L_T \in \tilde{\calO}(B^{\frac{1}{3}}
T^{\frac{2}{3}})$.
The worst case bound of Corollary~\ref{col:worstupper} occurs when all types appear an equal number
of times, i.e. $\Mtk = T/K$ for all $k$.
Intuitively, if only a few types appear most of the time, then the learner only needs to learn those
types well to obtain a small loss whereas if all types appear a large number of times then all their
mean values need to be estimated well to achieve a small loss.

\paragraph{Lower bounds:}
Next, we present arrival-dependent and worst-case hardness results to show that
the above observations and bounds are fundamental to this problem.
Our first result is a lower bound on the expected loss for any algorithm in terms of
the number of arrivals for each type.
Let $\calA$ be the class of all algorithms for this problem and $\calP$ denote the class of all
problems with $K$ types with $\sigma$-subGaussian observations.
We will write
$\LT(A, P)$ to explicitly denote the loss~\eqref{eqn:mabloss} of algorithm  $A$ on problem $P$.

\vspace{0.1in}
\begin{theorem}[Arrival-dependent Lower Bound] 
\label{thm:mablower}
Let $K > 0$, and $T \ge 1$ be given and
fix the number of arrivals $\{M_{T,k}\}_{k=1}^K$ for each $k\in[K]$.
Then,
\begin{align*}
    \inf_{A\in \calA} \sup_{P \in \calP} \bbE L_T(A,P) \ge \frac{1}{4}\sigma^{\frac{2}{3}}B^{\frac{1}{3}}\sum_{k=1}^K M_{T,k}^{\frac{2}{3}}
\end{align*}
\end{theorem}

Comparing this with the upper bound in Theorem~\ref{thm:mab}, we see that it 
 matches the lower
bound up to constant and logarithmic terms.
While the upper bound is in high probability, it is easy to obtain a bound in expectation by setting
$\delta=1/T$ and using the tower property of expectation.
Finally, we can obtain a
worst-case lower bound, stated formally below, by setting $\MTk=T/K$ for all $k$.
\begin{corollary}[Worst-case Lower Bound] \label{col:worstlower}
Let $K > 0$, and $T \ge 1$ be given.
Then,
\begin{align*}
    \inf_{A\in \calA} \sup_{P \in \calP} \sup_{\{\Mtk\}_k}
        \bbE L_T(A,P) \ge \frac{1}{4}\sigma^{\frac{2}{3}}B^{\frac{1}{3}} K^{\frac{1}{3}}
        T^{\frac{2}{3}}
\end{align*}
\end{corollary}

\paragraph{Proof Sketches:}
We will conclude this section by outlining our main proof ideas.
For the upper bound, we separately bound the prediction errors for rounds where we labeled
and for rounds we did not.
In the rounds that were not labeled, we used concentration results of sub-Gaussian random variables
and the fact that the uncertainty is bounded by the threshold $B^{\nicefrac{1}{3}}
\Mtxt^{\nicefrac{-1}{3}}$ to bound the sum of prediction errors.
To bound the total number of rounds that type $k$ was labeled, we used the fact that if
$\Ntk$ was too large,
that would reduce the uncertainty $\Utk$ sufficiently fast,
and hence it will fall below the threshold resulting in type $k$ not being labeled. 
The careful choice of the threshold gives the optimal trade-off between both terms in this
bound. 
For the lower bound, we use Le Cam's method~\citep{le1960approximation} to derive
a lower bound of the expected prediction error for type $k$ as a function  of $\Ntk$.
The lower bound is achieved 
by the best possible growth rate of $\Ntk$ against $\Mtk$.

\section{RKHS Setting}
\label{sec:gp}

While the setting in \S\ref{sec:kdiscrete} was useful for developing intuition and
optimality results, it does not reflect practical use cases for this problem.
Hence, in this section,
we will assume that each input is a $d$-dimensional (finite) feature vector and that
the output can be modeled using a smooth function.

\paragraph{Assumptions and problem set up:}
Specifically, we will assume $\xt\in [0,1]^d$
and that if we label $\xt$, we will observe $\yt = f(\xt) +
\epsilon_t$ where the noise $\epsilon_t$ is bounded, i.e $|\epsilon_t| \leq \sigma$,
and that $\EE[\epsilon_t] = 0$.
To make the problem tractable, we will assume that $f$ belongs to a
reproducing kernel Hilbert space (RKHS) with a kernel
$\kappa:[0,1]^d\times [0,1]^d \rightarrow \RR_+$, and moreover has bounded RKHS
norm~\citep{smola1998learning}.
Our loss for this problem takes a similar form to~\eqref{eqn:mabloss}:
\begin{align}
    L_T = N_T B + \sum_{t=1}^T |f(x_t) - p_t|,
    \label{eqn:gploss}
\end{align}
where recall that
$\pt$ is the prediction made by the learner for $f(\xt)$ and
$N_t$ is the total number of times we chose to label in $t$ rounds.
For what follows,
we will find it necessary to construct estimates along with confidence intervals for the unknown function
$f$ using past data.
We will do so via Gaussian processes (GPs).
While GPs are a Bayesian construct, they can be used to obtain  frequentist
confidence intervals when $f$ is in an RKHS.
Hence, we will begin with a brief review of GPs in the Bayesian setting.

\paragraph{Gaussian Process:} A GP, written as $f \sim
\calG\calP(\mu, \kappa)$, is characterized by a prior mean function $\mu : \calX
\to \bbR$ and prior covariance kernel $\kappa: \calX^2 \to \bbR$.
Some common choices for the kernel are the squared exponential (SE) 
kernel $\kappa_{\rm SE}$ and the Mat\'{e}rn kernel $\kappa_{\rm M}$, defined below:
\begin{align*}
\kappa_{\rm SE}(x, x') &= \exp\big(-(2l^2)^{-1}\|x-x'\|^2\big),
    \numberthis
    \label{eqn:kerneldefns}
\\
\kappa_{\rm M}(x, x') &= \frac{2^{1-\nu}}{\Gamma(\nu)}r^\nu B_\nu(r),
\quad r = (\sqrt{2\nu}/l)\|x-x'\|.
\end{align*}
Here $l$ is a lengthscale parameter, $\nu\in\RR_+$ is a smoothness parameter for Mat\'ern kernels and $B_{\nu}$ is a modified Bessel function.
If $f \sim
\calG\calP(\mu, \kappa)$, then $f(x)$ is distributed normally as $\calN(\mu(x),
\kappa(x,x))$ for all $x$.
Given $n$ observations $A = \{(x_i, y_i)\}_{i=1}^n$,
the posterior for $f$ is a GP with covariance $\kappa_A$,
standard deviation $\sigma_A$, and mean $\mu_A$ given by,
\begingroup
\allowdisplaybreaks
\begin{align*} 
    &\kappa_A(x,\tilx) = \kappa(x,\tilx) - k^\top(K+\sigma^2I_n)^{-1}\tilk,
    \numberthis
    \label{eqn:gp_update_sigma}
    \\
    &\sigma_A(x) = \sqrt{\kappa_A(x, x)},
    \hspace{0.2in}
    \mu_A(x) = k^\top(K + \sigma^2I_n)^{-1} Y,
\end{align*}
\endgroup
Here $Y \in \bbR^n$ such that $Y_i = y_i$, $k$, $\tilk \in \bbR^n$ are such that $k_i
= \kappa(x, x_i)$, $\tilk_i = \kappa(\tilx, x_i)$, $K \in \bbR^{n\times
n}$ is given by $K_{i,j} = \kappa(x_i, x_j)$, and $I_n \in \bbR^{n\times n}$ is the
identity matrix.
While GPs usually assume $y_i = f(x_i) + \epsilon_i$ and $\epsilon_i\sim\calN(0,\sigma^2)$
in the Bayesian setting,
when the noise $\epsilon_i$ is bounded as outlined in the beginning of this section, we
can use~\eqref{eqn:gp_update_sigma} to obtain frequentist confidence intervals for $f$.

\paragraph{Maximum Information Gain (MIG):}
One quantity of interest going forward
 will be the maximum information gain $\gamma_n$ for a GP after $n$
observations~\citep{DBLP:conf/icml/SrinivasKKS10}:
\begin{align}
\gamma_n =
\max_{\{x_1,\dots,x_n\}} \frac{1}{2}\log\left(1 + \sigma^{-2}
\sigma_{\{(x_j, y_j)\}_{j=1}^{i-1}}^2(x_i)\right).
\label{eqn:mig}
\end{align}
Here, $\sigma$ is bound on the noise and $\sigma_{\{(x_j, y_j)\}_{j=1}^{i-1}}$
is the posterior standard deviation using the first $i-1$ points~\eqref{eqn:gp_update_sigma}.
The MIG is used to quantify the
problem complexity in adaptive decision-making with
GPs~\citep{DBLP:conf/icml/SrinivasKKS10,kandasamy2017multi,chen2017near}.
In the Bayesian setting, the term inside the $\max$ operator is the mutual information
between $f$ and the $n$ observations, and $\gamma_n$ can be interpreted as the maximum
amount of information $n$ observations gives about $f$.
The following bounds are known for GPs with SE and Mat\'ern
kernels~\citep{DBLP:conf/icml/SrinivasKKS10}:
\begin{align} \label{eqn:migsematern}
   \mbox{SE kernel: } &\gamma_n \in \calO\left((\log n)^{d+1} \right),
    \\
   \nonumber \mbox{Mat\'ern kernel: } & \gamma_n \in
        \calO\left(n^{\frac{d(d+1)}{2\nu + d(d+1)}} \log n\right).
\end{align}

\subsection{Algorithm}
\label{sec:gpalgo}

Our algorithm for this formulation, outlined in Algorithm~\ref{alg:gpucb},
is similar in spirit to Algorithm~\ref{alg:mab},
but instead uses the GP posterior mean and standard deviation for the prediction and
the uncertainty.
For brevity, let us denote the GP mean using the observations in the first
 $t-1$ rounds by $\mu_{t-1}$.
Precisely, $\mu_{t-1}(x) = \mu_{A_{t-1}}(x)$, where
$A_{t-1} = \{(x_s, y_s); 1\leq s \leq t-1 \text{\;and\;} \ell_s=1\}$;
recall $\mu_A$ from~\eqref{eqn:gp_update_sigma} and that $\ell_s=1$ if we label on round
$s$.
Similarly, let $\sigma_{t-1}$ denote the GP standard deviation after $t-1$ rounds.

On each round, when point $\xt$ arrives, we will choose to label $\xt$ if
the GP standard deviation $\sigma_{t-1}(\xt)$, which quantifies the uncertainty of
$\mu_{t-1}(\xt)$,
 is larger than a cost and time-dependent threshold $\tau(t)$
(line~\ref{lin:gpifthreshcondn}).
This threshold is set separately for the SE and Mat\'ern kernels as follows:
%
%
\begin{align} \label{eqn:thresholdSE}
   \mbox{SE kernel: } &\tau(t) = \sqrt{2\sigma^2} B^{\frac{1}{d+3}} t^{-\frac{1}{d+3}}, \\
   \nonumber \mbox{Mat\'ern kernel: } & \tau(t) = \sqrt{2\sigma^2} B^{\frac{1}{2d+3}} t^{-\frac{1}{2d+ 3}} .
\end{align}
While these specific choices for $\tau$ were chosen to optimize for our final bound,
we see that, similar to the discrete case,
the threshold increases with cost $B$ to discourage frequent labeling when it is
expensive,
and decreases with the total number of arrivals $t$ to encourage labeling as time goes on
to minimize faulty prediction costs.
The resulting procedure is summarized in Algorithm~\ref{alg:gpucb}.

\insertGPAlgorithm

\subsection{Theoretical Results}

Our main theorem in this section is the following upper bound for the
loss~\eqref{eqn:gploss}
for Algorithm~\ref{alg:gpucb}.

\begin{theorem} \label{thm:gp-anytime}
Let $\delta \in (0,1)$ be given.
Assume that $f$ lies in an RKHS $\calH_\kappa(\calX)$ corresponding to the
kernel $\kappa(x, x')$, and denote its RKHS norm by $\|f\|_\kappa$.
Assume that the noise $\epsilon_t$ has a zero mean conditioned on the
history and is bounded by $\sigma$ almost surely.
Recall the MIG $\gamma_n$ be as defined in~\eqref{eqn:mig} and~\eqref{eqn:migsematern},
and denote $\beta_t = 2\|f\|_\kappa^2 + 300 \log^3(t/\delta)$.
Then,  with probability at least $1-\delta$, for all $T\geq 0$,
Algorithm~\ref{alg:gpucb} satisfies the following for the SE and Mat\'ern kernels.

For the SE kernel $L_T \in \tilde{\calO}\left(B^{\frac{1}{d+3}} T^{\frac{d+2}{d+3}} \right)$
with, 
    \begin{align*}
        L_T  &\le\; (C_1 + C_2 \beta_{T+1}^{1/2}) B^{1-d_1} T^{d_1}  \; +
        \\
        &\hspace{0.3in} \sqrt{C_3 \beta_{T+1} B^{-d_1} T^{d_1} \gamma_{C_1B^{-d_1} T^{d_1}}},
    \end{align*}
    where $d_1 = \frac{d+2}{d+3}$,
    $C_1 = (\sigma l)^{-d} d^{d/2}$, 
    $C_2 = \frac{ \sqrt{2}\sigma(d+3)}{d+2}$, $C_3 = \frac{2}{\log(1+\sigma^{-2})} C_1$,
    and $l$ is from~\eqref{eqn:kerneldefns}.
    
For a  Mat\'{e}rn kernel $L_T \in \tilde{\calO}(B^{\frac{1}{2d+3}} T^{\frac{2d+2}{2d+3}})$ with, 
    \begin{align*}
        L_T &\le ( C_3 + C_4\beta_{T+1}^{1/2}) B^{1-d_2}T^{d_2}  \; + \\
        &\hspace{0.3in} \sqrt{C_5 \beta_{T+1} B^{-d_2}T^{d_2}\gamma_{C_3 B^{-d_2}T^{d_2}}} 
    \end{align*}
    where $d_2 = \frac{2d+2}{2d+3}$,
    $C_3 = \sigma^{-2d} (2\sqrt{d}L_{\rm M})^{d}$, 
    $C_4 = \frac{\sqrt{2}\sigma(2d+3)}{2d+2}$, $C_5 = \frac{2}{\log(1+\sigma^{-2})} C_3$,
    and $L_{\rm M}$ is a Lipschitz constant for the Mat\'ern kernel~\eqref{eqn:kerneldefns}.
\end{theorem}

The theorem establishes that Algorithm~\ref{alg:gpucb} is able to achieve sublinear loss
for both kernels.
While the bound admittedly worsens with dimensionality $d$, this is to be expected in high
dimensions under nonparametric assumptions.
For instance, for bandit problems with a Mat\'ern kernel, the regret bound scales
at rate $\tilde{\calO}(T^{\frac{\nu + d(d+1)}{2\nu + d(d+1)}})$.
One key difference in this result, when compared to Theorem~\ref{thm:mab}, is that it only provides
a worst-case upper bound.
This is primarily because characterizing the number of arrivals in
different regions of $[0, 1]^d$ is not as straightforward as it was in the discrete setting.
Our empirical evaluation in \S\ref{sec:experiments} shows that Algorithm~\ref{alg:gpucb},
much like Algorithm~\ref{alg:mab}, does indeed perform
better when the arrivals are skewed than if they are uniform.


Finally,
it is worth mentioning that in both bounds, the second term of the RHS
is of a lower order than the first term;
after accounting for the MIG~\eqref{eqn:migsematern},
for the SE kernel, the exponent of $T$ in the second term is $\frac{d+2}{2(d+3)}$
 and for the
Mat\'ern kernel, it is
$\frac{2d+2}{2d+3}\frac{\nu + d(d+1)}{2\nu + d(d+1)}$ which is smaller than
$\frac{2d+2}{2d+3}$.

\paragraph{Proof sketch:}
The main conceptual difference between the RKHS setting and the $K$ discrete setting is that here
we use the GP standard deviation as our measure of uncertainty, 
which is not directly connected to the number of labeled points. 
Thus it is difficult to choose a threshold to 
balance with the number of labels. 
We instead use a discretization argument 
which allows us to bound the uncertainty in each bin by the number of labeled points in that bin,
and carefully choose the size of the discretization to optimize for the final bound.
For this proof, we also used a prior result
to obtain frequentist confidence intervals for $f$
from the GP updates~\citep{DBLP:conf/icml/SrinivasKKS10}.

\section{Experiments}
\label{sec:experiments}

\begin{figure*}[ht!]
\centering
\begin{tabular}{cccc}
    \includegraphics[width = 0.23\textwidth]{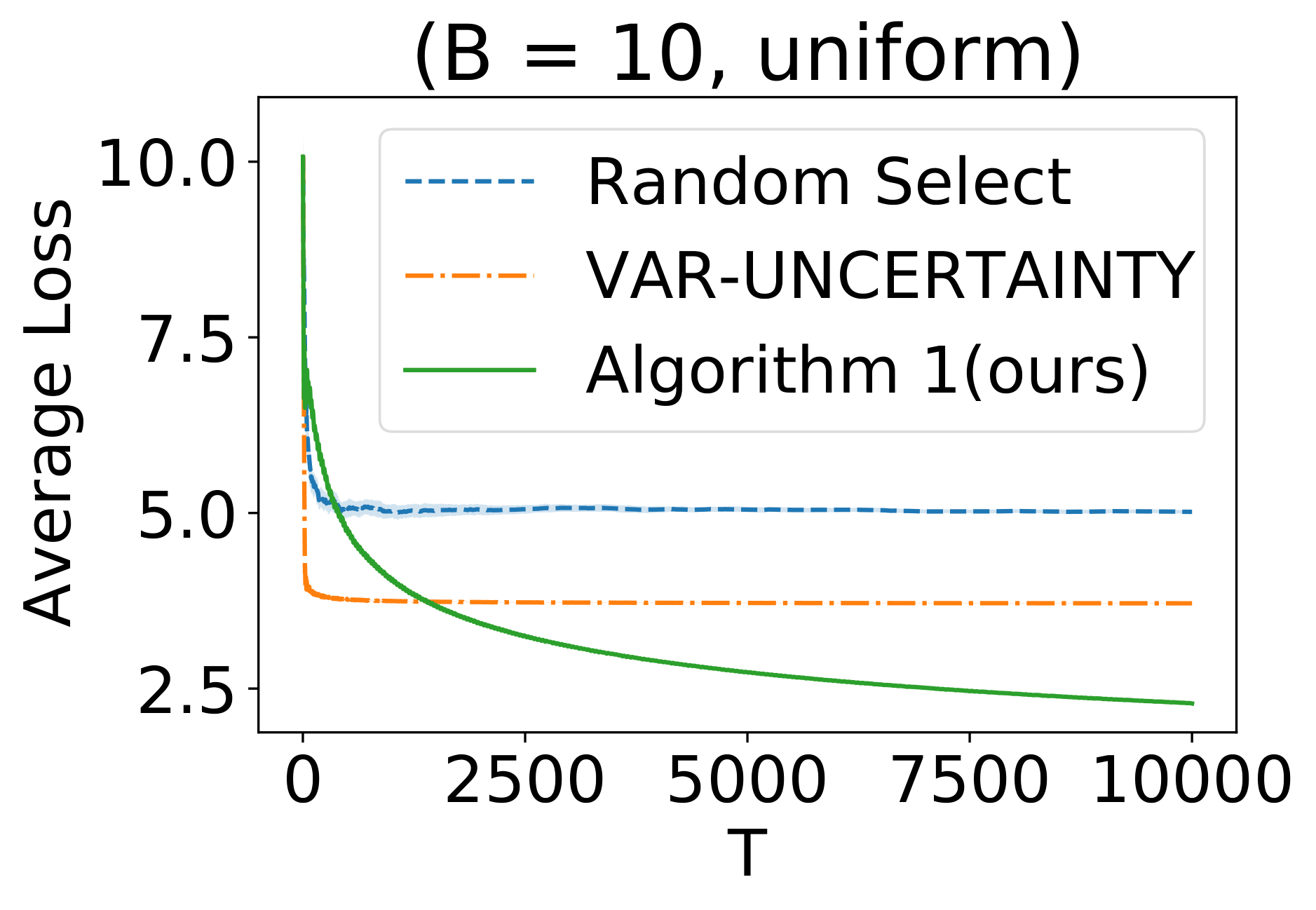} &
    \includegraphics[width = 0.22\textwidth]{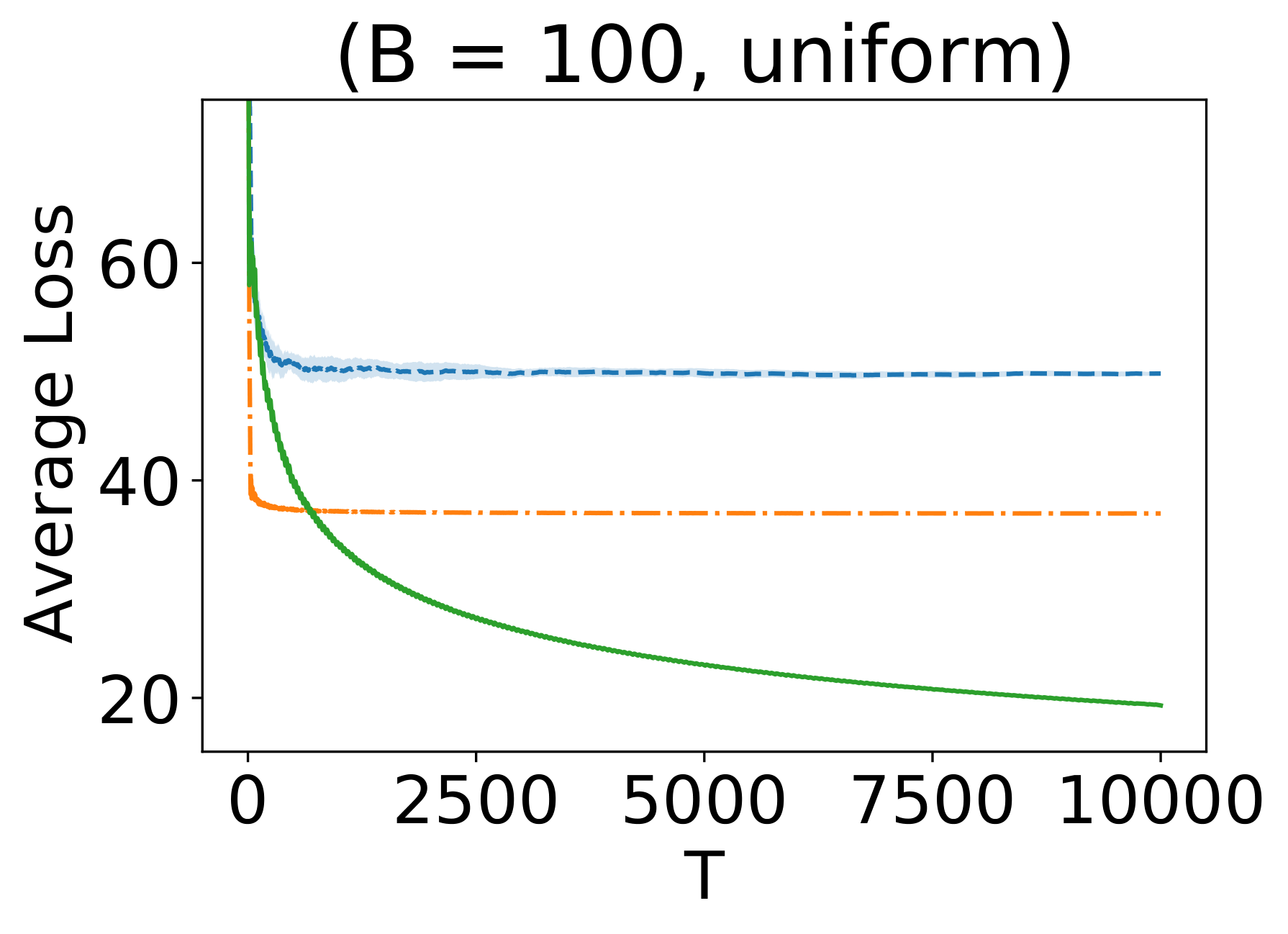} &
    \includegraphics[width = 0.22\textwidth]{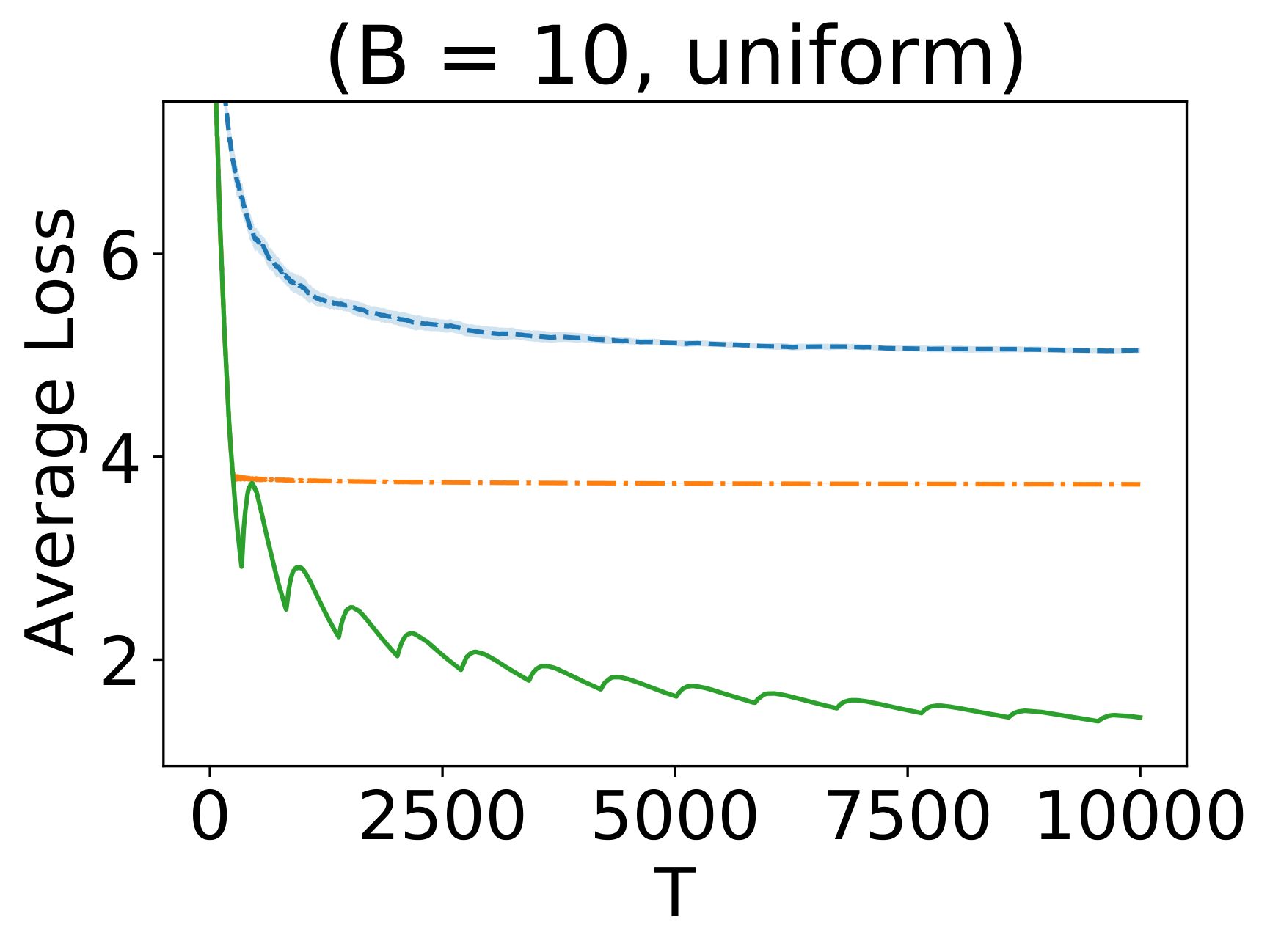} &
    \includegraphics[width = 0.23\textwidth]{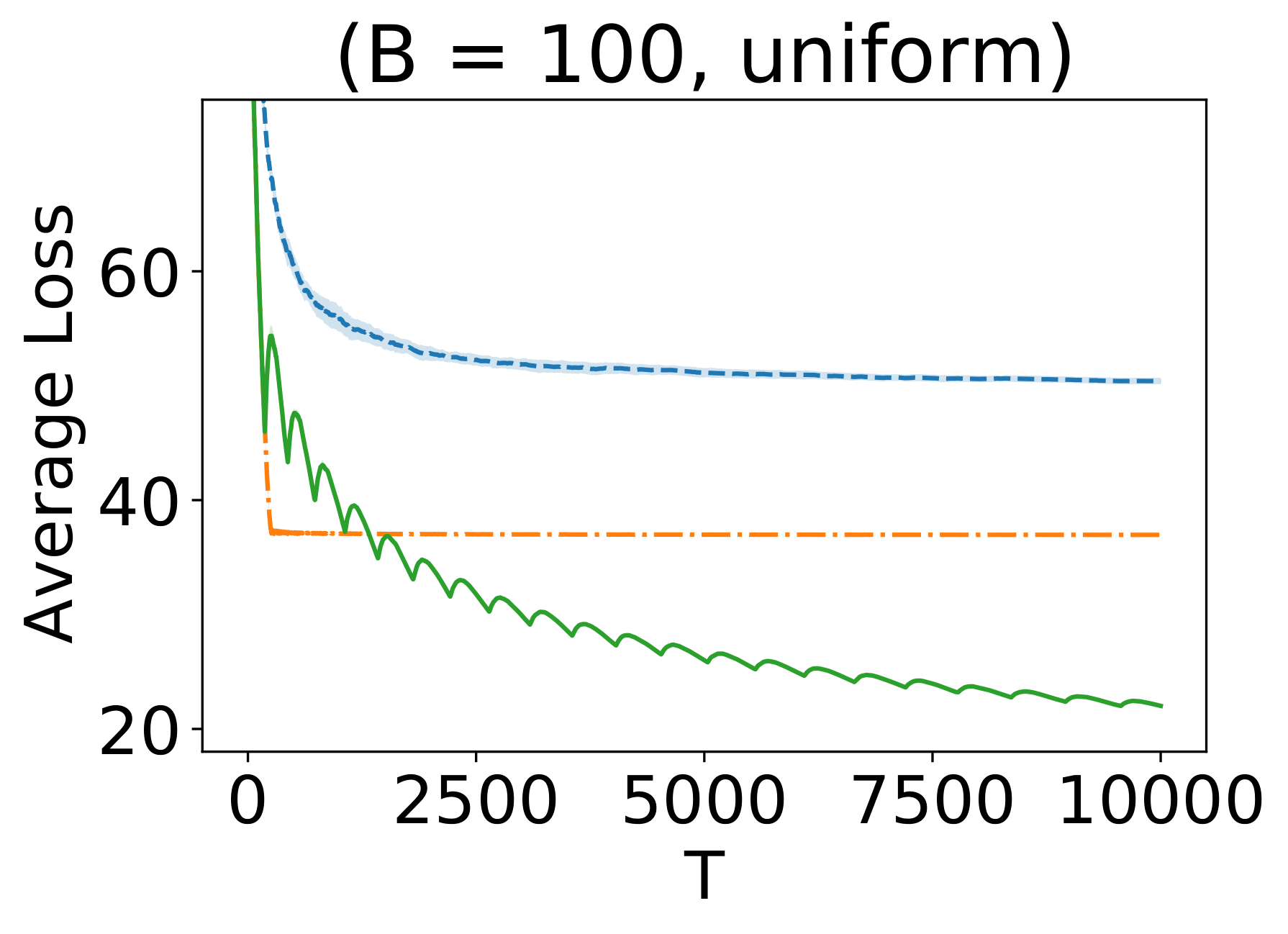}
     \\
     \includegraphics[width = 0.22\textwidth]{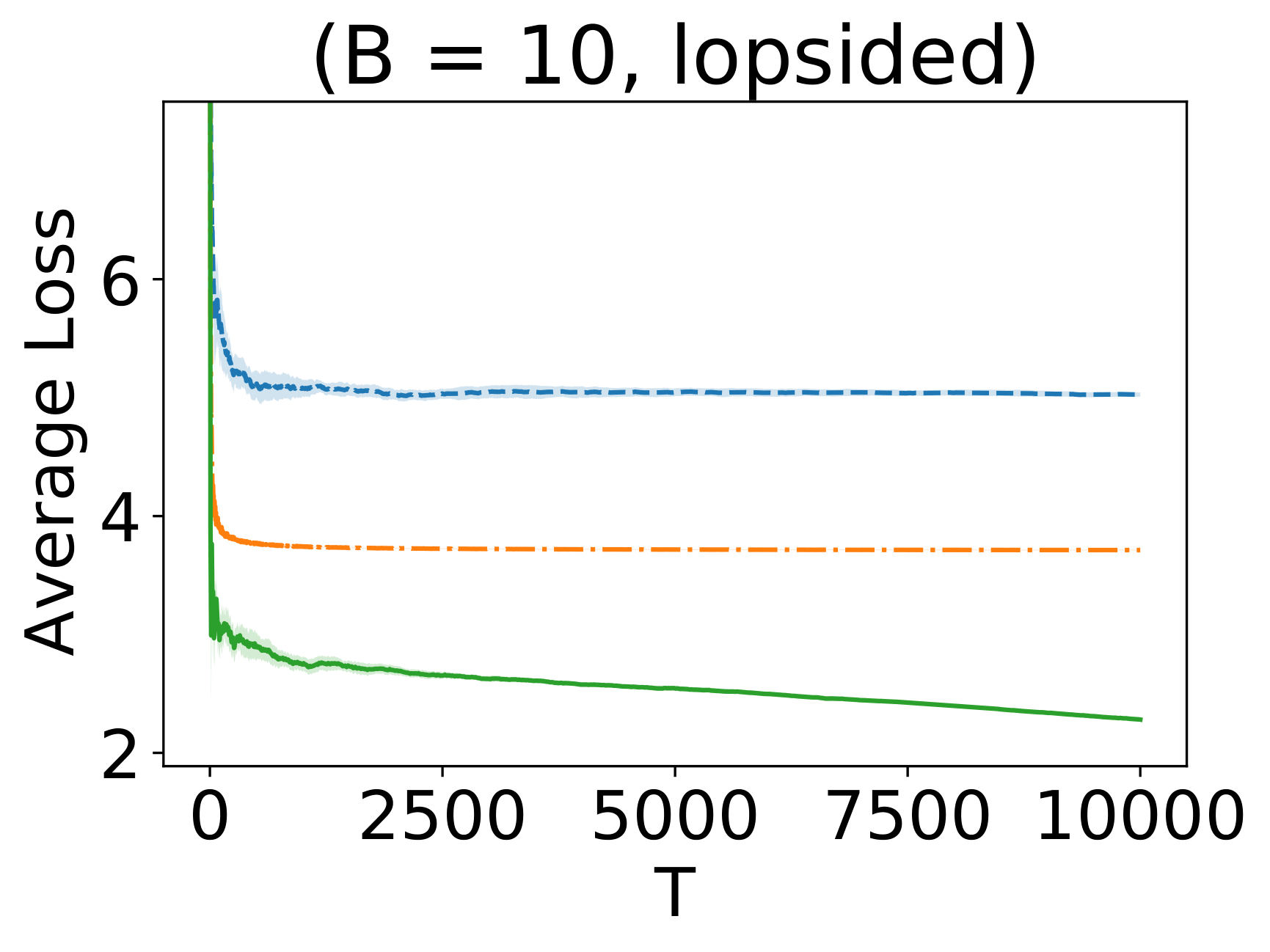}&
     \includegraphics[width = 0.223\textwidth]{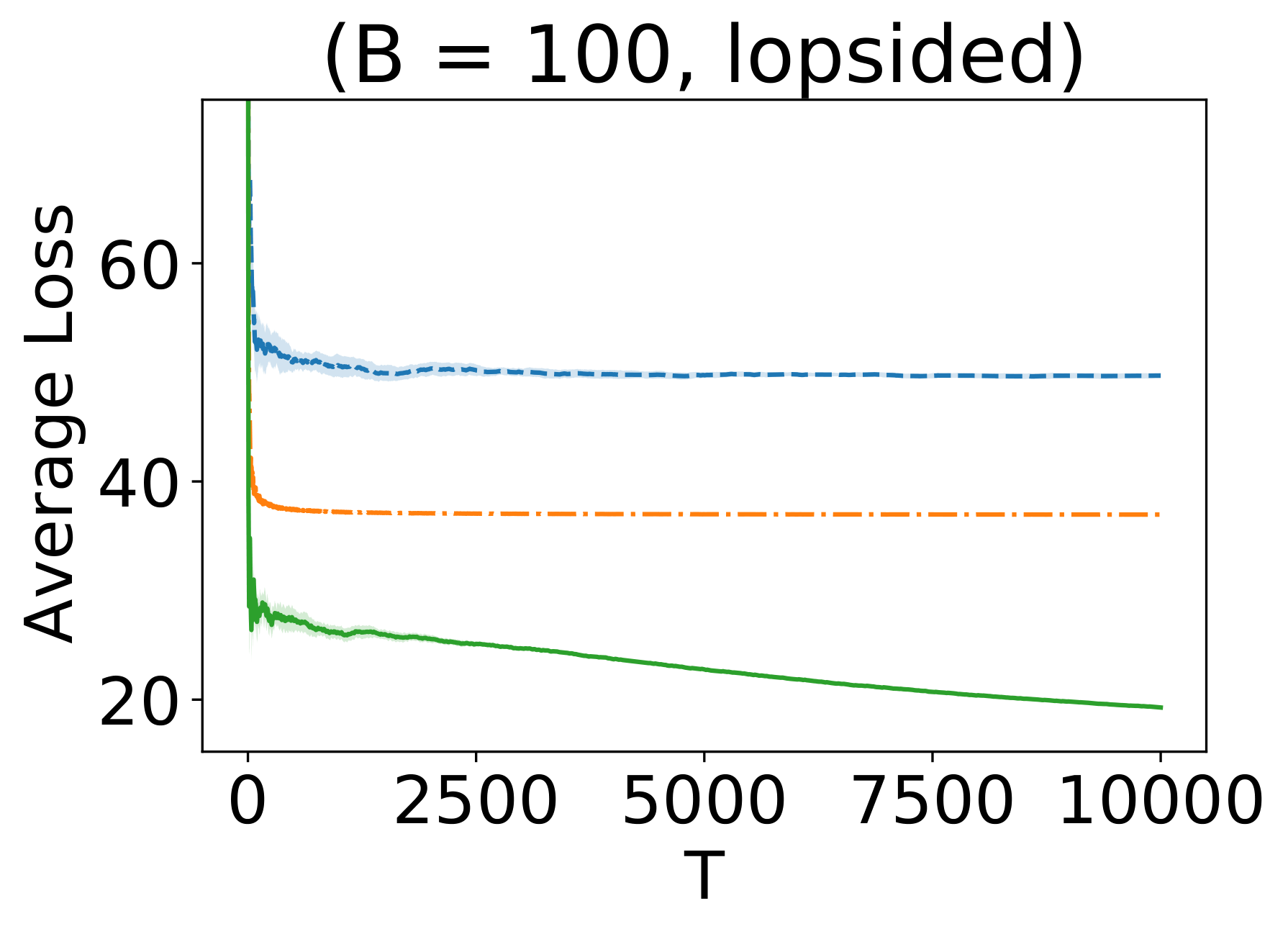}&
    \includegraphics[width = 0.22\textwidth]{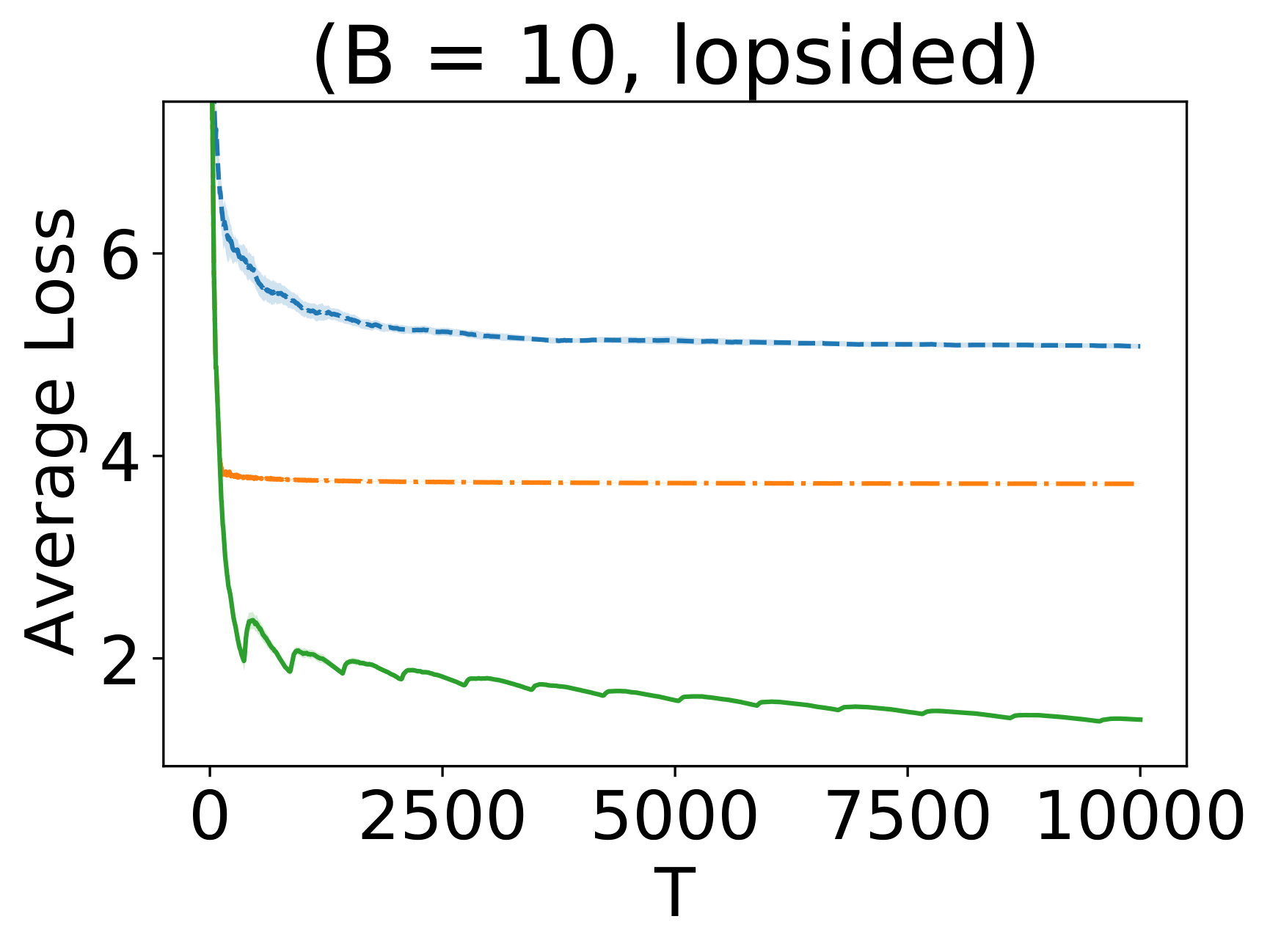} &
     \includegraphics[width = 0.23\textwidth]{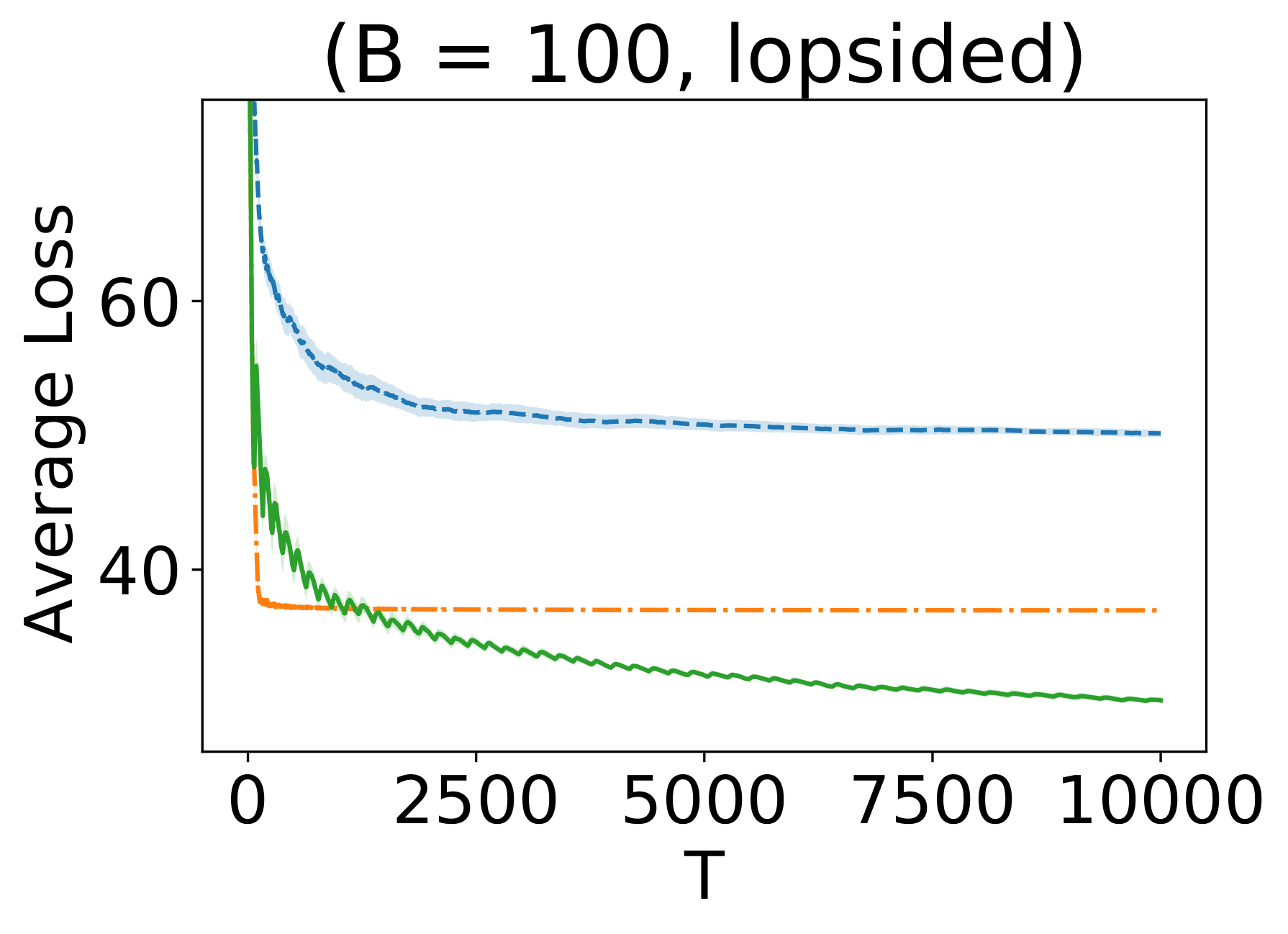}
\end{tabular}
\vspace{-0.1in}
\caption{\small Results comparing the three methods for $K$ discrete setting
of~\S\ref{sec:expsynK}. $K = 10$ for column 1 and
2. $K = 100$ for columns 3 and 4. The results are averaged over 10 trials, and we report the $95\%$
the confidence interval of the standard error.}
\label{fig:discrete}
\end{figure*}

\begin{figure*}[ht!] 
\centering
\begin{tabular}{cccc}
     \includegraphics[width = 0.23\textwidth]{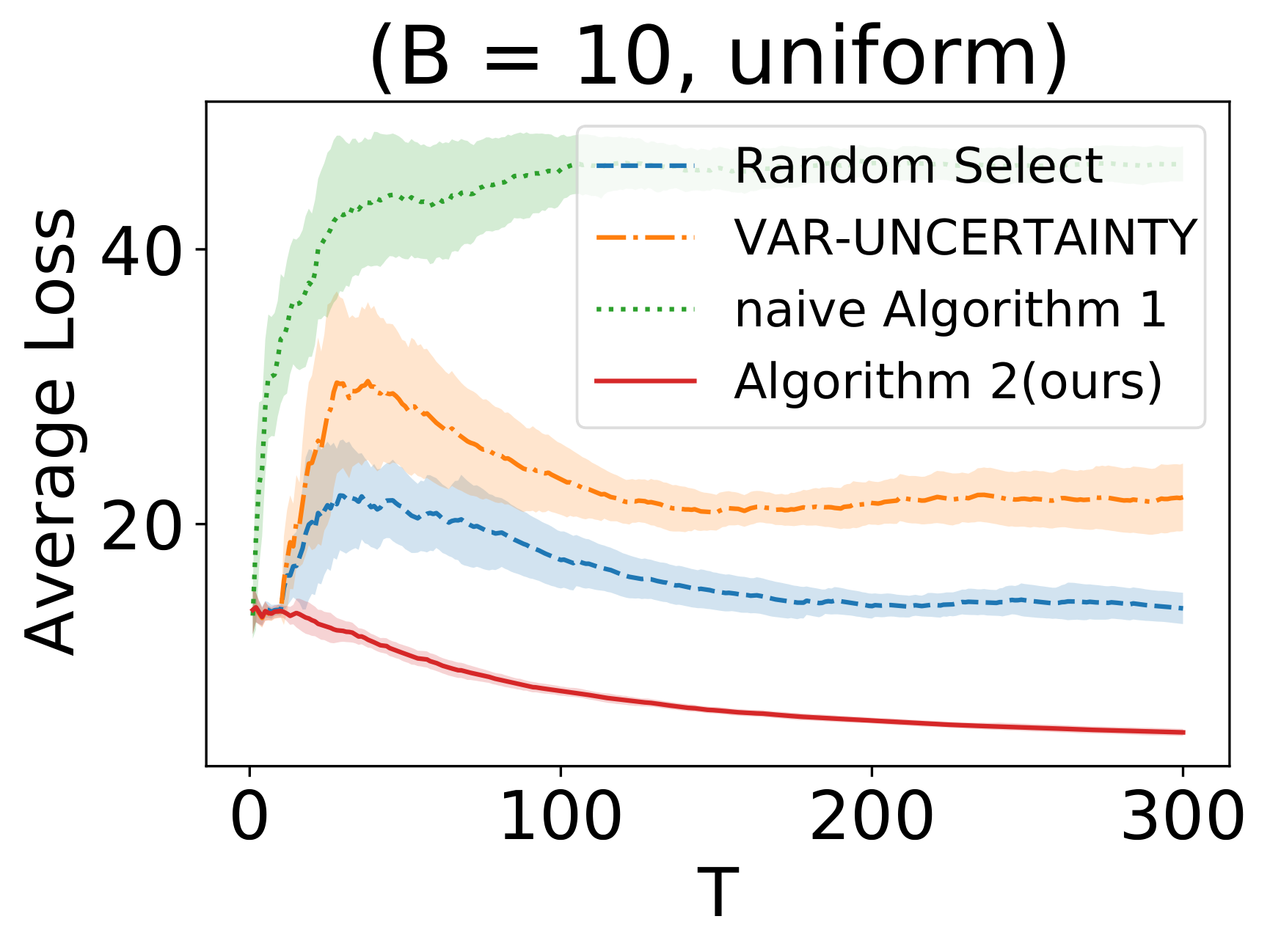} &
     \includegraphics[width = 0.23\textwidth]{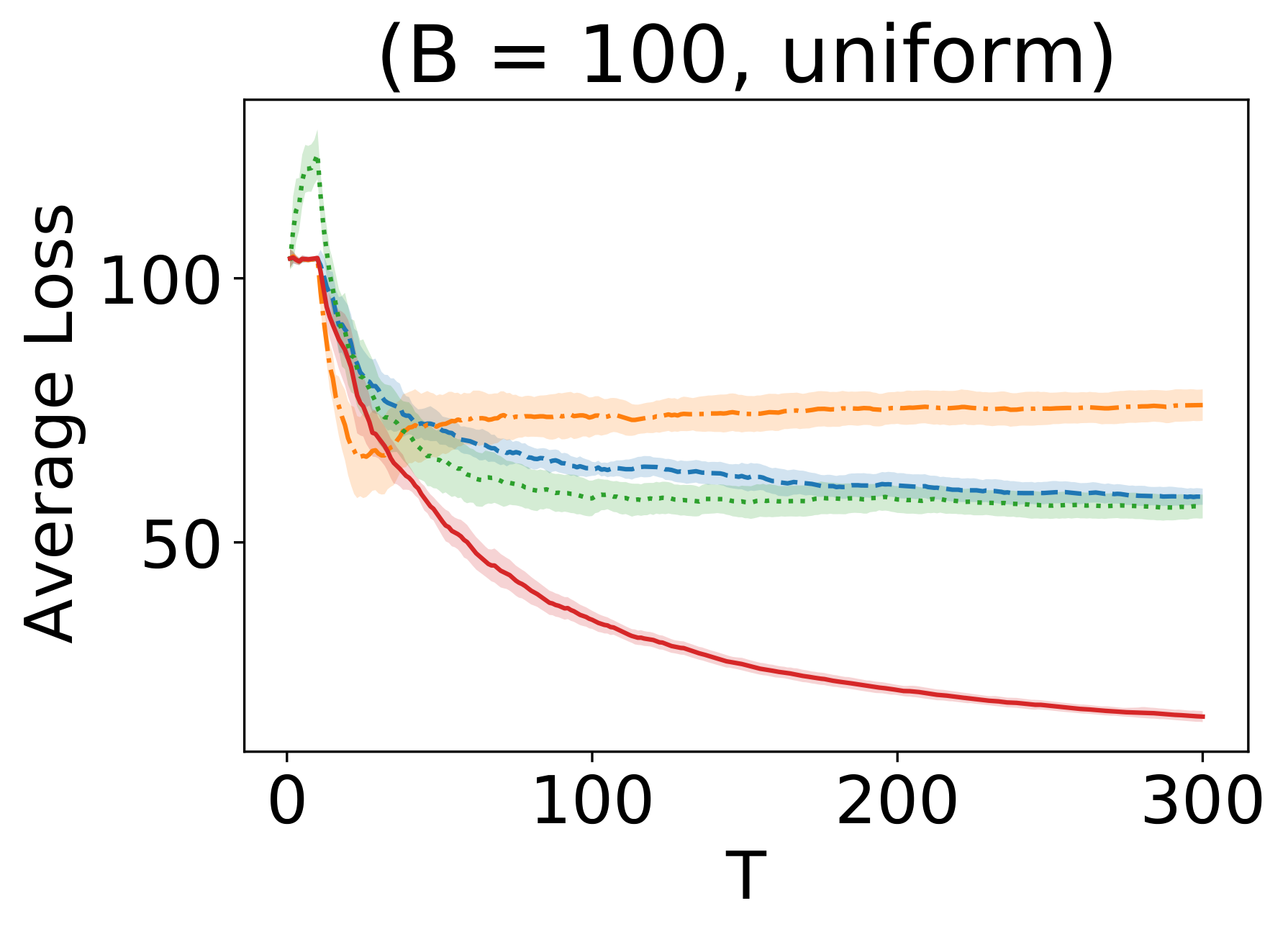} &
     \includegraphics[width = 0.23\textwidth]{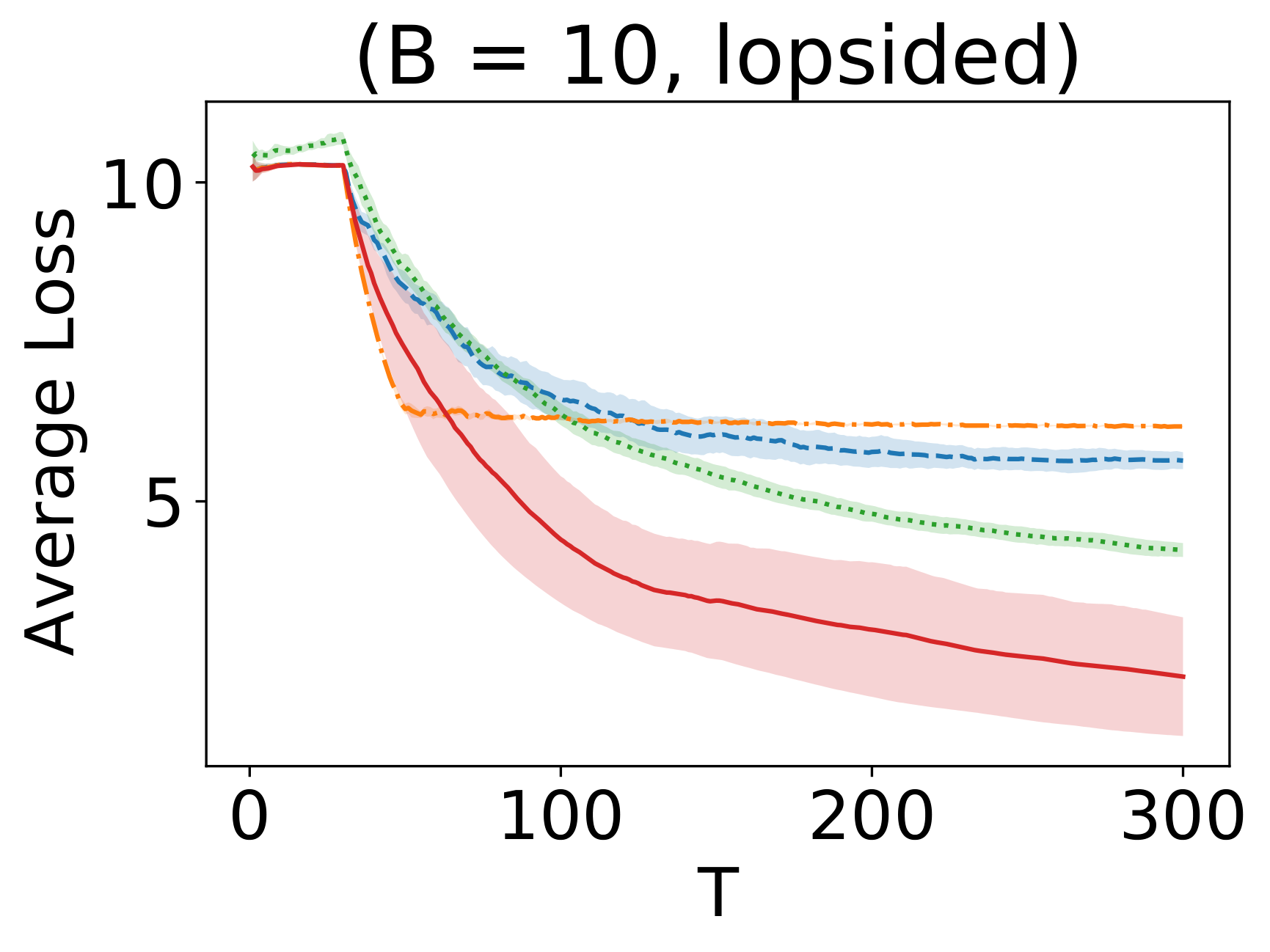} &
     \includegraphics[width = 0.23\textwidth]{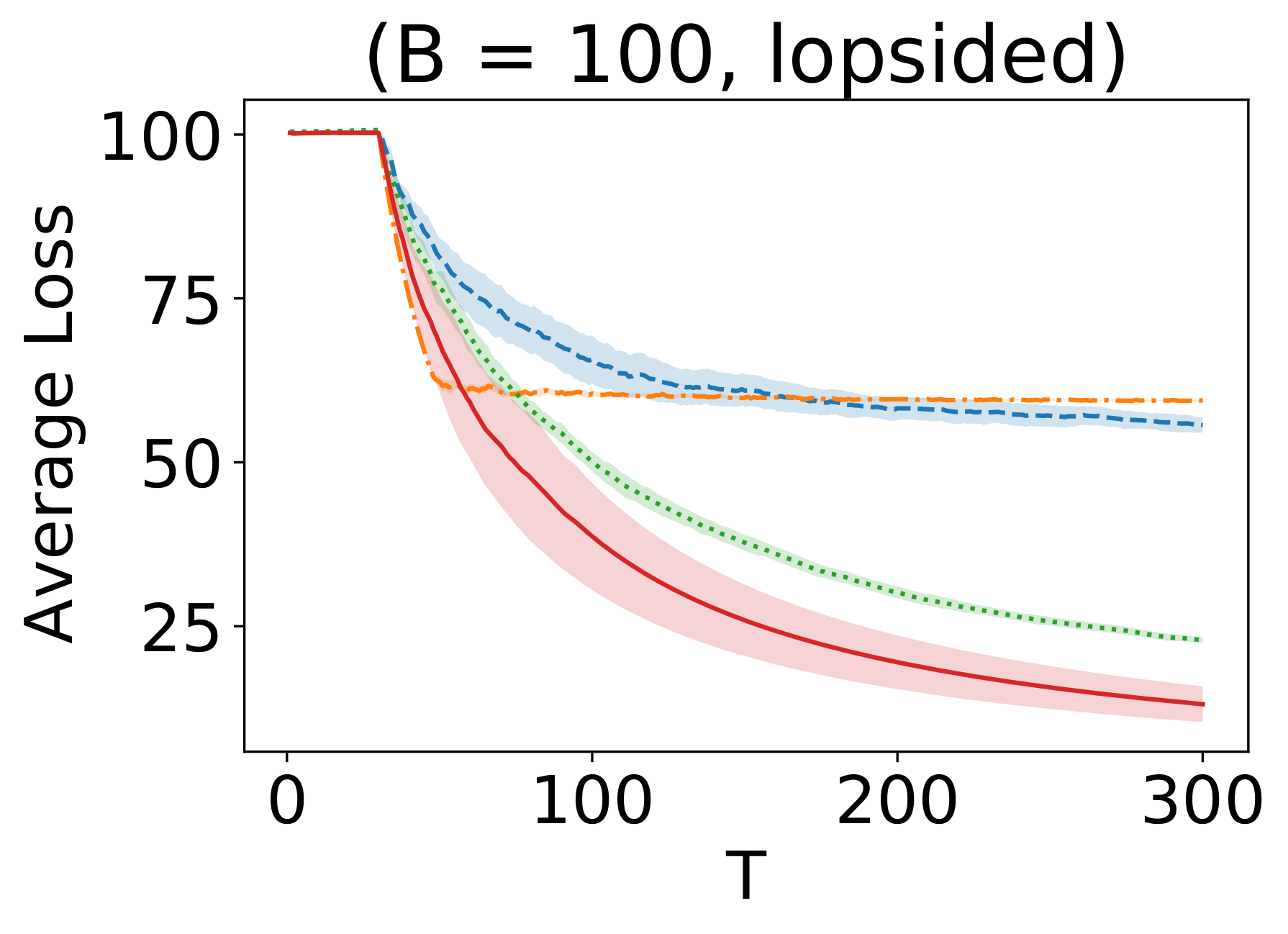}
\end{tabular}
\vspace{-0.1in}
\caption{\small Results comparing the four methods for synthetic functions in~\S\ref{sec:expsynsmooth}.
Columns 1 and 2 show the results
when the arrival pattern is uniform  for Branin function. Column 3 and 4 show the results when the
arrival pattern is lopsided for the Hartmann function. The results are averaged over 10 trials, and we
report the $95\%$ confidence interval of the standard error.}
\label{fig:gp_synthetic}
\end{figure*}

\begin{figure*}[ht!]
\centering
\begin{tabular}{cccc}
     \includegraphics[width = 0.31\textwidth]{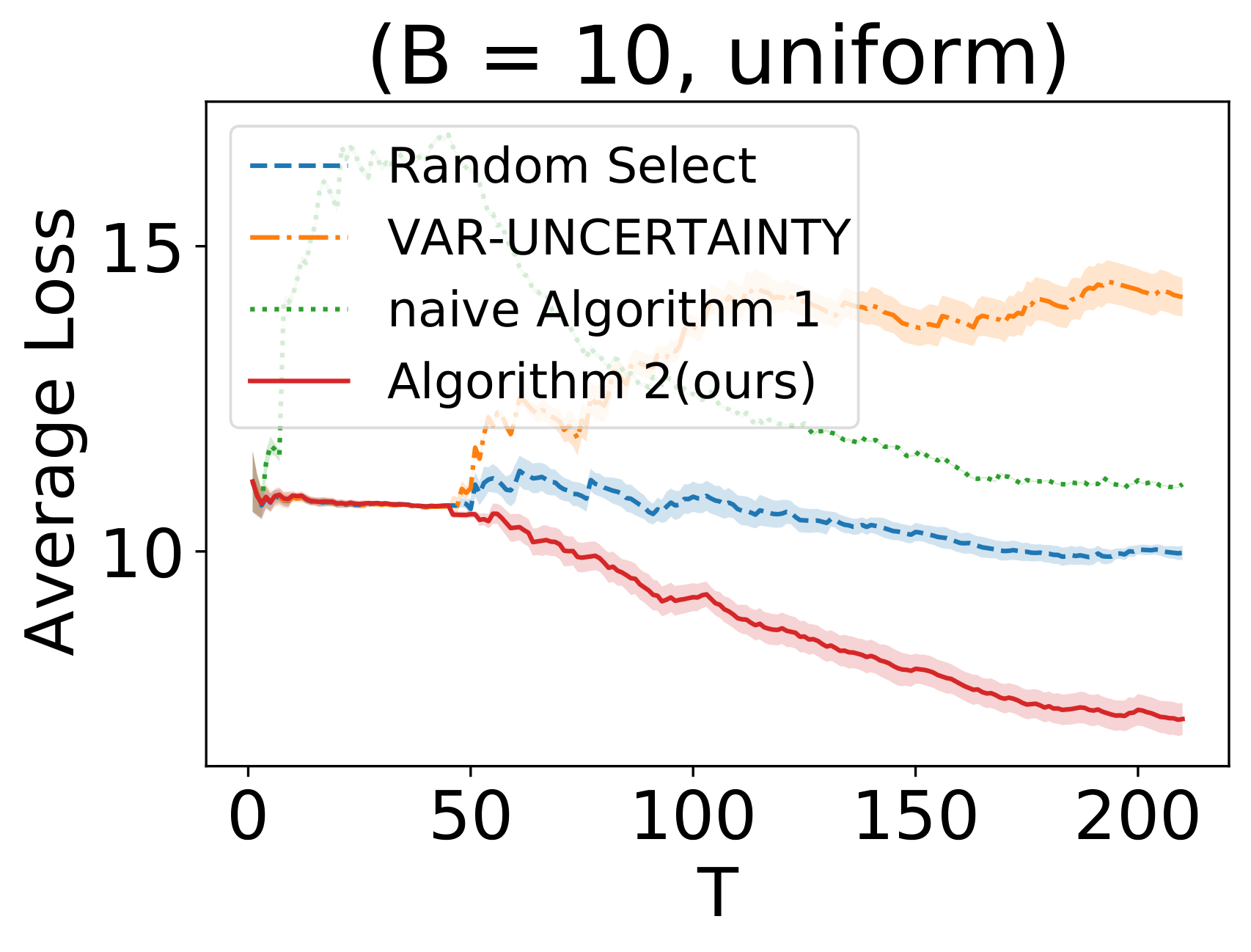} &
     \includegraphics[width = 0.31\textwidth]{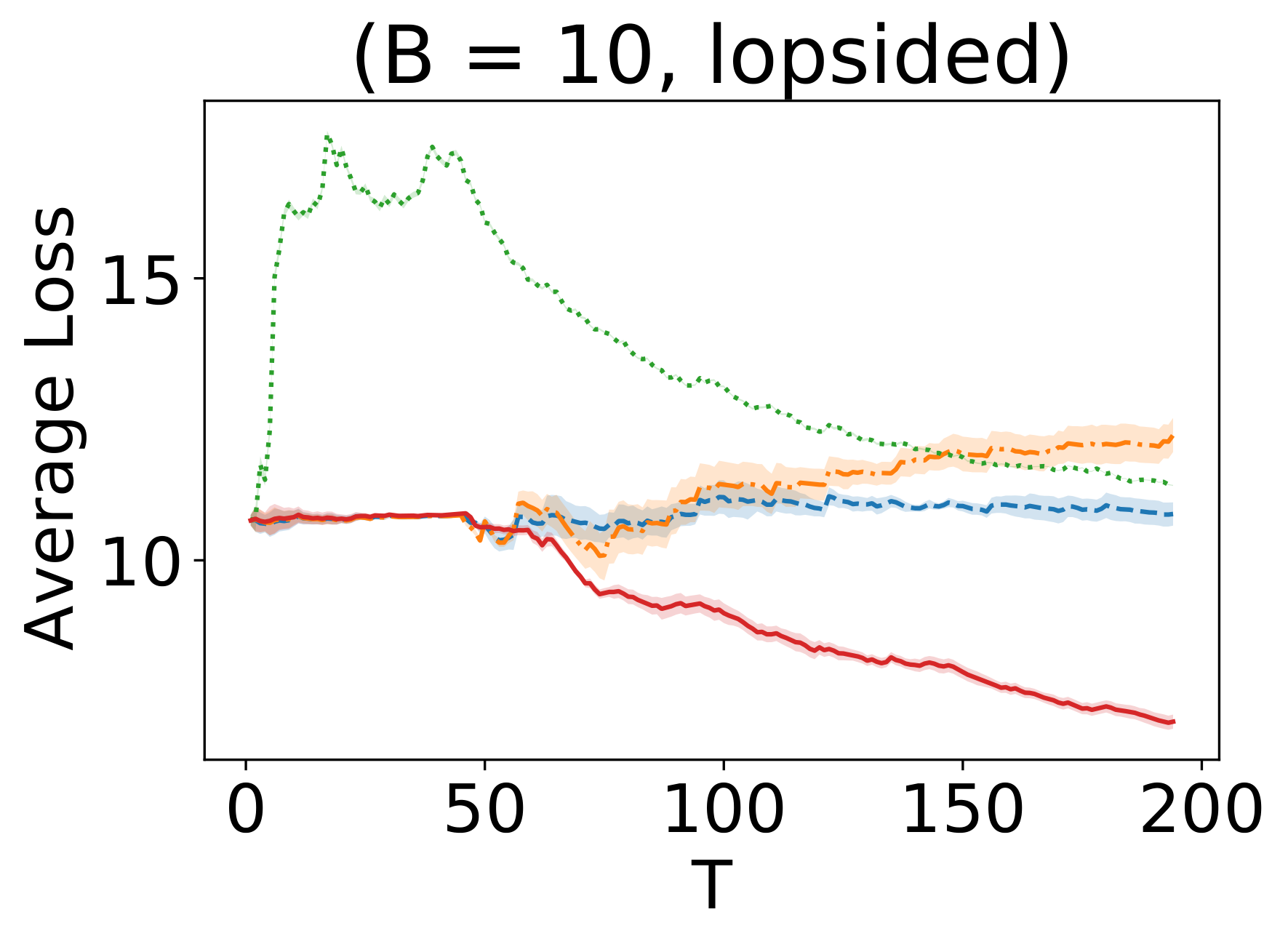} &
     \includegraphics[width = 0.31\textwidth]{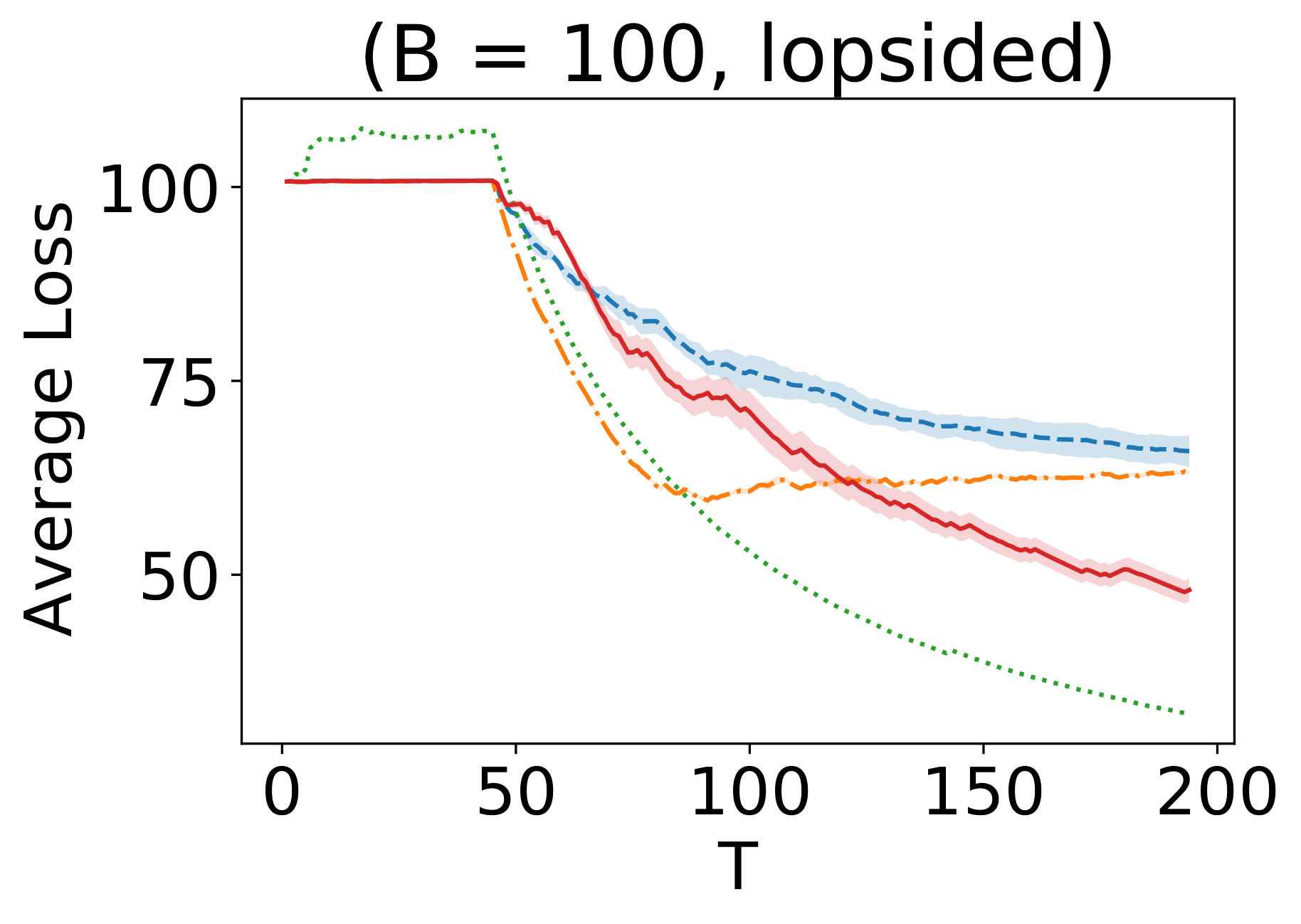}
\end{tabular}
\vspace{-0.1in}
\caption{\small Results comparing the four methods for Parkinsons dataset. The results are averaged over 5 trials and we report the $95\%$ confidence interval of the standard error.}
\label{fig:parkinsons}
\end{figure*}

\begin{figure*}[ht!]
\centering
\begin{tabular}{ccc}
     \includegraphics[width = 0.31\textwidth]{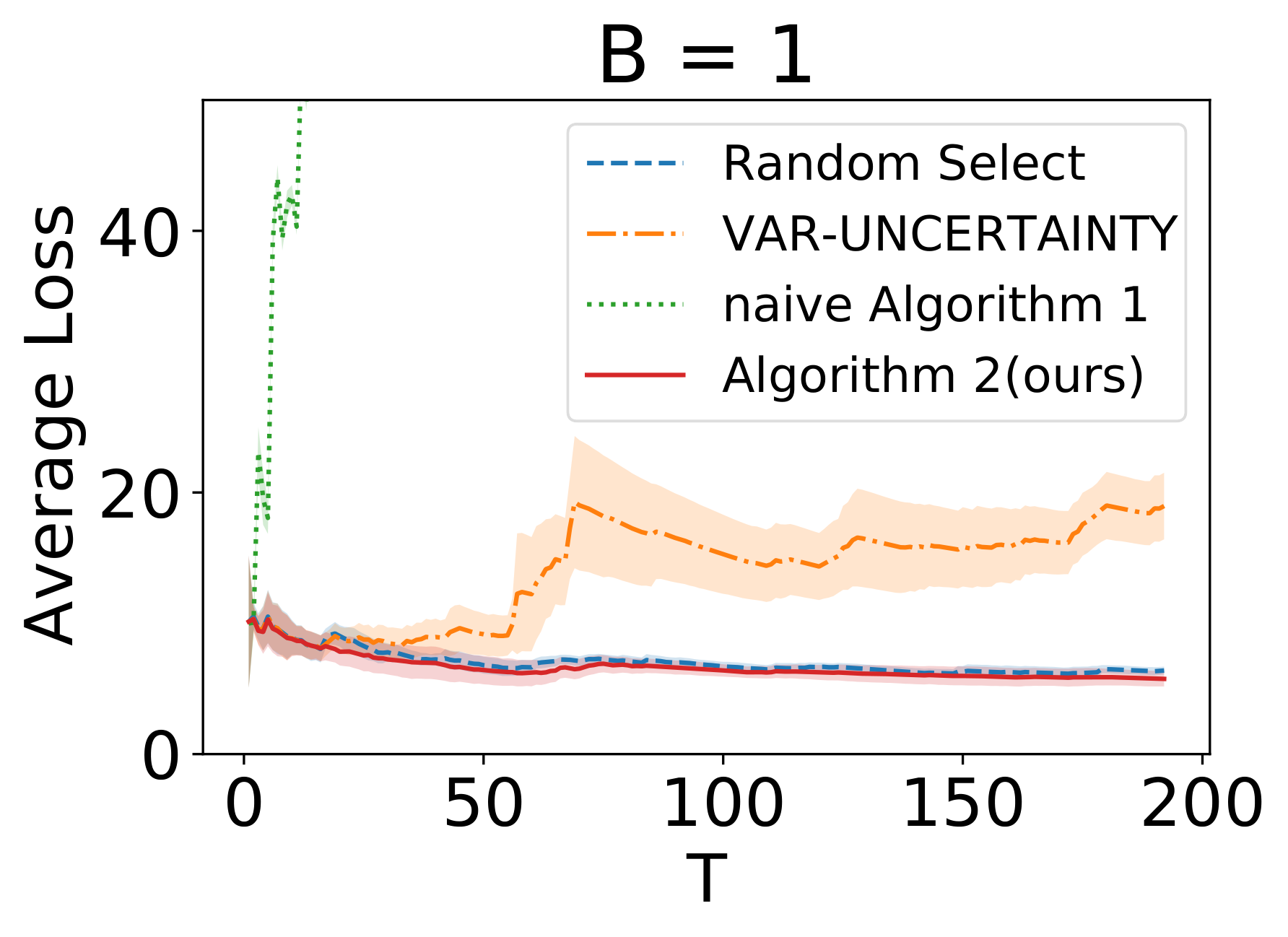} &
     \includegraphics[width = 0.31\textwidth]{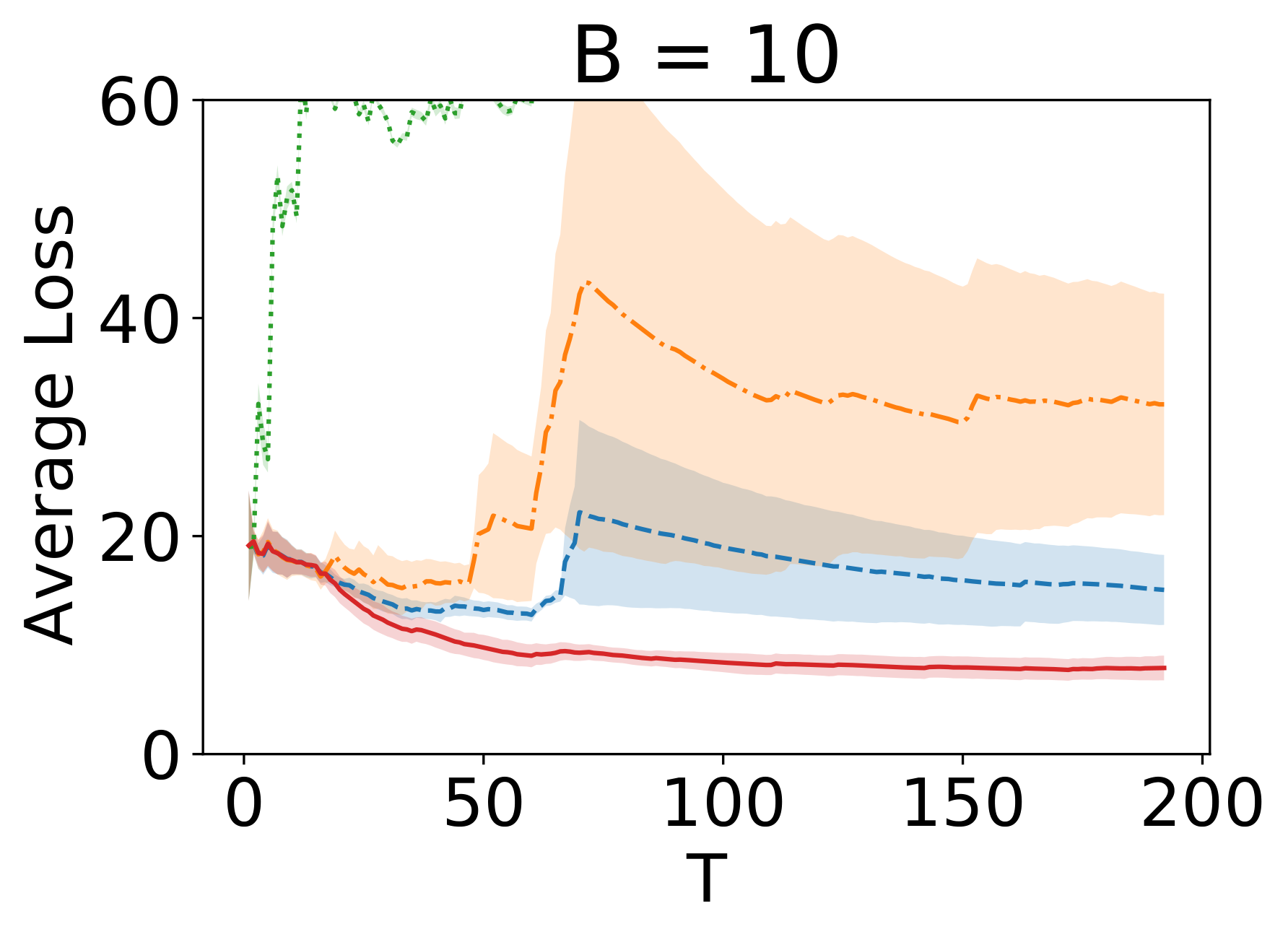} &
     \includegraphics[width = 0.31\textwidth]{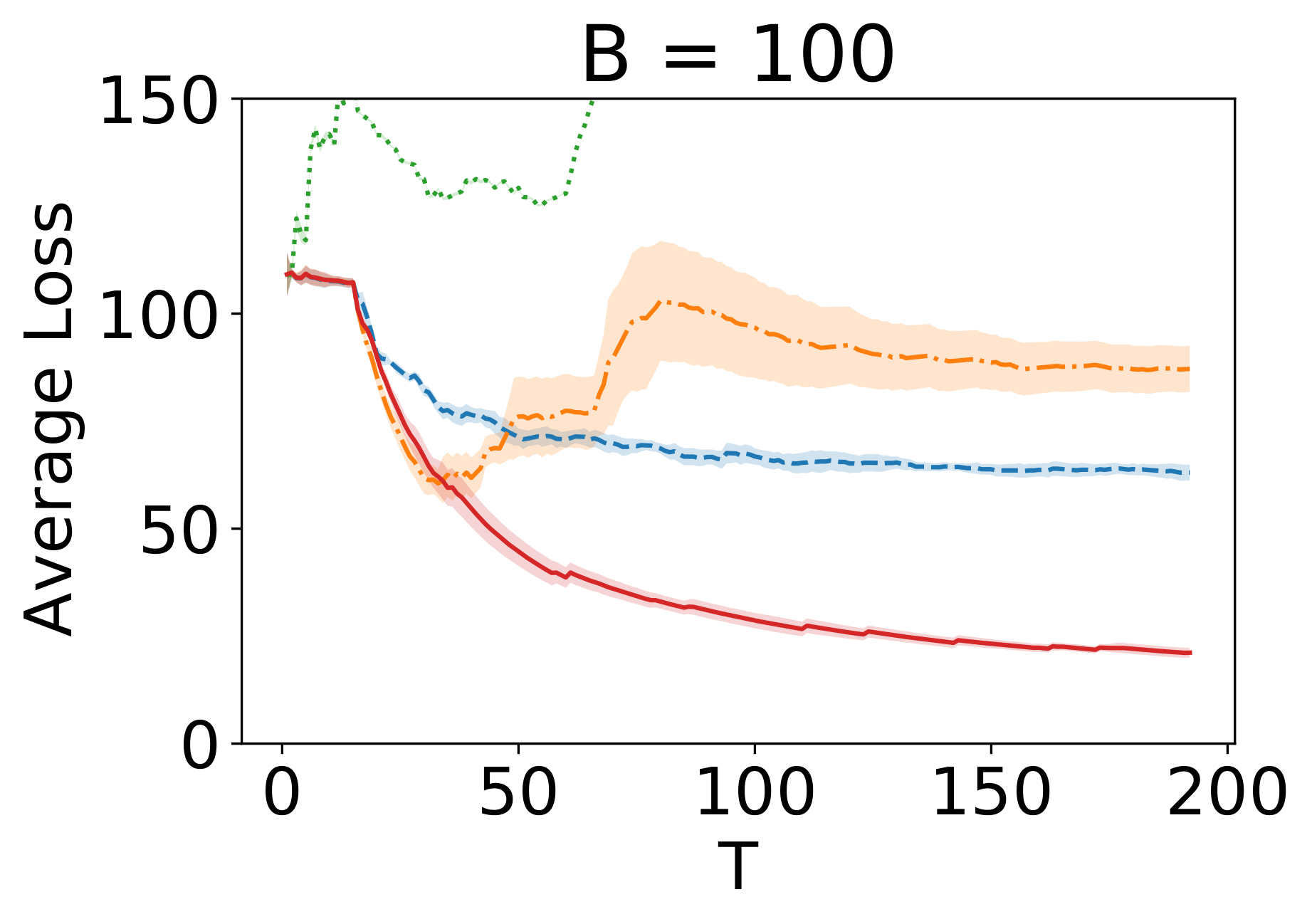}
\end{tabular}
\vspace{-0.1in}
\caption{\small
Results comparing the four methods for the supernova dataset. The results are averaged over 5 trials and
we report the $95\%$ confidence interval of the standard error.
The naive Algorithm~\ref{alg:mab} performs poorly so we
limit the $y$-axis to focus on the rest methods.}
\label{fig:supernova}
\end{figure*}

We evaluate our algorithms on synthetic and real
experiments, both in the discrete setting and when $f$ is a smooth function.
We compare against the following two baselines:
\emph{1. Random Select}: On each round, this baseline labels with probability $0.5$.
We experimented with different values in $[0,1]$ and found that $0.5$ worked best
across all our experiments. (Note that this includes, as a special case, the baseline which labels all points when the probability of $0.5$)

\emph{2. VAR-UNCERTAINTY~\citep{DBLP:journals/tnn/ZliobaiteBPH14}}:
This baseline chooses to label if the uncertainty is less than a threshold $\theta$, which
is decreased ($\theta\leftarrow \theta/2$) each time if it labels and
increased ($\theta\leftarrow 2\theta$) each time if not.
$\theta$ is initially set to be 1 as suggested by the authors.
We use the same uncertainty measure as we did for
our algorithms in both settings.

For both baselines, we determine $\pt$ via the sample mean of $\xt$'s labels for the
$K$ discrete setting and via the mean of a fitted GP model when $f$ is a smooth function. We also considered other methods outlined in the related work section but they could not be included
for the following reasons.
The batch active learning methods~\citep{lindstrom2010handling, zhu2007active} are hard to
adapt to our setting since we assume data comes one at a time.
Some methods~\citep{werner2022online,desalvo2021online} are tailored for
classification and have no straightforward generalization to our (regression) problem.
The remaining methods are for pool-based active learning and do not apply to our setting.

\vspace{-0.1in}
\subsection{Synthetic experiments with $K$ discrete types}
\label{sec:expsynK}

First, we compare the performance between Algorithm~\ref{alg:mab} with \emph{Random
Select} and \emph{VAR-UNCERTAINTY} when $K \in \{10, 100\}$. For each value of $K$, the actual mean values are generated by random sampling in the domain
$[0,1]$.
The labels $\yt$ are drawn from a Gaussian distribution with variance $0.01$ and the corresponding mean
value.

We design two different arrival patterns:
\emph{(i) uniform}: when all $K$ types appear uniformly at random,
 \emph{(ii) lopsided}: when $20\%$ of the types appear $80\%$ of the time and the remaining $80\%$ of
the types appear $20\%$ of the time.
The cost of labeling is set to
be $B = \{10, 100\}$ and the total number of rounds is set to be $T = 10^4$.  
We slightly modify our threshold function to $\lambda B^{1/3}M_{t,k}^{-1/3}$,
where $\lambda > 0$ is a hyperparameter tuned on a held-out synthetic dataset.
(Note that our rates do not change by adding a constant to the threshold function.) The
values of $\lambda$ used in each experiment can be found in Section~\ref{sup:sec_lambda} in the
Appendix. Figure~\ref{fig:discrete} shows the experimentation results when the arrival pattern is
uniform and lopsided for both $K = 10$ and $K = 100$.

\vspace{-0.1in}
\subsection{Synthetic experiments with smooth $f$}
\label{sec:expsynsmooth}

In this section, we compare Algorithm~\ref{alg:gpucb} with \emph{Random Select},
\emph{VAR-UNCERTAINTY}, and a naive version of Algorithm~\ref{alg:mab},
which splits each domain into $2^d$ discrete types by halving along each dimension.
For Algorithm~\ref{alg:gpucb}, 
we use a GP prior with zero mean and an SE kernel.
While our theoretical analysis assumes the kernel parameters (e.g length scales)
are known, they need to be tuned well for good empirical performance.
Hence, we initialize this method by choosing to label the first $5d$ points and tuning the
kernel parameters by maximizing the GP marginal likelihood.
We adopt the same procedure for the baselines since they also use a GP to output the prediction.
We run each experiment under different costs $B = \{10, 10^2\}$ and $T = 300$.  As we did for the $K$ discrete setting, we modified the threshold function to
    $\tau(t) = \lambda \cdot \sqrt{2\sigma^2} B^{\frac{1}{d+3}} C^{-\frac{1}{d+3}}$
where the hyper-parameter $\lambda > 0$ was tuned on a held-out set.

\textbf{Test functions:}
We use the Branin function ($d=2$) and Hartmann function ($d=6$) for our synthetic evaluation~\citep{dixon1978global}.
We form a uniform arrival pattern for the Branin function whose input domain is  $[-5, 10]\times[0, 15]$.
For the Hartmann function whose input domain is $[0,1]^6$, we form a lopsided arrival pattern where $80\%$ of the $\xt$'s coming from $[0, 0.2]\times[0,1]^5$ and
$20\%$ coming from $[0.2, 1]\times [0,1]^5$. 
For the Branin function, we set the standard deviation of the noise to be $\sigma = 5$
and for the Hartmann function, we set it to be $\sigma = 0.5$.

Figure~\ref{fig:gp_synthetic} shows the results for both of the synthetic functions.
Algorithm~\ref{alg:gpucb} performs the best among all four methods under different
arrival patterns and costs.

\vspace{-0.1in}
\subsection{Real experiments with smooth $f$}

In this section, we consider the following two real-world datasets in healthcare and astronomy:

\paragraph{Healthcare:} We use the Parkinsons telemonitoring dataset~\citep{tsanas2009accurate}. They are $5875$ biomedical voice recordings from $42$
people who have early-stage Parkinson's disease. Each recording contains
$20$ measurements (from voice and personal statistics), from which we choose $9$ different
measurements (\emph{age, sex, Jitter(\%), Shimmer, NHR, HNR, RPDE, DFA, PPE}) as the features for the input
$x_t$ and use the measurement \emph{total UPDRS} as the label $f(x_t)$ that needs to be predicted.  

As before, we create both uniform and lopsided arrival patterns for this experiment.
For uniform arrivals,
we randomly select $5$ recordings from each patient and shuffle the order
resulting in $210$ data points. For lopsided arrivals, we randomly select $20$ recordings
from patients whose \emph{subject\#} are from $1$ to $8$ and randomly select $1$ recording from each
of the rest patients. There is a total of $194$ data points. We manually add noise to each label,
the noises are independent and come from the distribution of $\calN(0, 1)$.

Figure~\ref{fig:parkinsons} shows the results of the Parkinsons dataset under the uniform and
lopsided arrival patterns. Algorithm~\ref{alg:gpucb} outperforms almost all the other methods.
It is interesting to note that
the naive implementation of Algorithm~\ref{alg:mab} performs better than Algorithm~\ref{alg:gpucb}
 when $B = 100$. 
We found that while the naive algorithm's prediction was worse than other methods,
it was able to achieve a lower overall cost by reducing the amount of labeling.

\textbf{Astronomy:} We use the supernova dataset from~\citet{davis2007scrutinizing}. There are in total
$192$ data points in $\bbR^3$. Each dimension of the data point represents the Hubble constant, dark
matter fraction, and dark energy fraction respectively.
We use prior simulators \citep{kandasamy2017multi, robertson1936interpretation,
shchigolev2017calculating} to get the
label for each data point
and manually add independent $\calN(0, 100)$ noise to each label
(we choose the noise variance to be large since the  range of the labels
is large $[-980, -129]$).  

Figure~\ref{fig:supernova} shows the results for all four methods. 
We observe that \emph{Random Select} and
Algorithm~\ref{alg:gpucb} perform relatively similarly when $B = 1$. However, when $B$ increases,
Algorithm~\ref{alg:gpucb} outperforms all the other methods.

This concludes our experiments.
In the Appendix, we have provided
additional experiments, plots of average prediction errors for the above methods,
and an ablation
study of the effect of the tuning parameter $\lambda$ in our algorithms.

\section{Conclusion}
We studied actively labeling streaming data, where we need to balance between the cost of
labeling and the cost of making faulty predictions with few labels.
We described two algorithms in two different settings based on the same principle: label a point if
its predictive uncertainty is larger than a threshold;
this threshold increases with the cost of labeling and decreases with the number of arrivals.
In the first setting,
 where the inputs belong to $K$ discrete distributions, we provided matching lower and
upper bounds on the loss in terms of the arrival pattern of queries.
In the second setting, where the expected output can be modeled using an RKHS, we showed that the
algorithm achieves sublinear loss in the worst case.
We corroborated these results with empirical results on real and synthetic experiments.
%

\bibliographystyle{plainnat}
\bibliography{biblio}

\onecolumn
\section*{Appendix}
\section{PROOFS} \label{ax:proofs}
In this section, we present the detailed proofs of the Theorems and Corollaries in the main paper.

\subsection{Proof of Theorem~\ref{thm:mab}} \label{sec:proof_1}
\begin{proof}
{The proof goes as follows: for each type, we will decompose the prediction errors into rounds that
are labeled and rounds that are not labeled and then bound the two parts separately. After this, we
will upper-bound $N_{T,k}$, the number of times type $k$ was labeled, to bound the total labeling
cost.}

The following decomposition of the loss divides the prediction errors into rounds that are labeled
($\ell_t = 1$) and not labeled ($\ell_t = 0$) separately,
\begin{align} \label{eqn:1}
    \nonumber L_T &= \sum_{k=1}^K \bigg (B \cdot N_{T,k} + \sum_{t=1}^{T} \big|\mu_k - \muhattk \big| \cdot \mathds{1}_{\{x_t = k\}} \bigg) \\
    &= \sum_{k=1}^K \bigg (B \cdot N_{T,k} + \underbrace{\sum_{x_t = k, \ell_t = 1} \big|\mu_k - \muhattk \big|}_{A_1} + \underbrace{\sum_{x_t = k, \ell_t = 0} \big|\mu_k - \muhattk \big|}_{A_2} \bigg ) 
\end{align}
We will bound $A_2$ using sub-Gaussian concentration and the threshold function.
For this, we will consider the following event 
where $|\mu_k - \muhattk|$ can be bounded $\forall t, \forall k$.
\begin{align*}
    \calE = \bigg\{ \muhattk - \sqrt{\frac{2\sigma^2\log(2N_t^3/\delta)}{N_{t-1,k}}} \le \mu_k \le \muhattk + \sqrt{\frac{2\sigma^2\log(2N_t^3/\delta)}{N_{t-1,k}}} \quad \forall t \ge 1,  \forall k \in [K]  \bigg\}
\end{align*}
Then we consider the probability that event $\calE$ is false,
\begin{align*}
    \bbP (\overline{\calE}) &= \bbP \left(\forall t, \forall k \in [K], \left|\frac{1}{N_{t,k}} \sum_{s=1}^t y_s \mathds{1}_{\{x_s = k, \ell_s = 1\}} - \mu_k \right| \ge \sqrt{\frac{2\sigma^2\log(2N_t^3/\delta)}{N_{t-1,k}}}\right) \\
    &= \bbP \left(\forall t, \left|\frac{1}{N_{t,x_t}} \sum_{s=1}^t y_s \mathds{1}_{\{x_s = x_t, \ell_s = 1\}} - \mu_{x_t} \right| \ge \sqrt{\frac{2\sigma^2\log(2N_t^3/\delta)}{N_{t-1,x_t}}}\right)
\end{align*}
where the second equality is because we can assume the sequence of the incoming data points is fixed. Then we take a union bound on $t$ and get
\begin{align*}
    \bbP (\overline{\calE}) \le \sum_{t=1}^\infty \bbP \left(\left|\frac{1}{N_{t,x_t}} \sum_{s=1}^t y_s \mathds{1}_{\{x_s = x_t, \ell_s = 1\}} - \mu_{x_t} \right| \ge \sqrt{\frac{2\sigma^2\log(2N_t^3/\delta)}{N_{t-1,x_t}}}\right)
\end{align*}
Since $N_{t-1, x_t}$ is a random variable that can take values in $\{1, \cdots, t-1\}$, we take a union bound on $N_{t-1, x_t}$. Also since $N_t \ge N_{t-1, x_t}$, we replace $N_t^3$ by $N_{t-1, x_t}^3$ in the probability and get
\begin{align*}
    \bbP (\overline{\calE}) &\le \sum_{t=1}^\infty \sum_{j=1}^{t-1} \bbP\left(\left|\frac{1}{N_{t, x_t}}\sum_{s=1}^{t}  y_s\mathds{1}_{\{x_s=x_t, \ell_s=1\}} - \mu_{x_t}\right| \ge \sqrt{\frac{2\sigma^2\log(2j^3/\delta)}{j}} \right) \\
    &\le \sum_{t=1}^\infty \sum_{j=1}^{t-1} 2\exp\left(-\frac{j}{2\sigma^2}\cdot \frac{2\sigma^2\log(2j^3/\delta)}{j}\right) & (\mbox{by subGaussian concentration})\\
    &\le \sum_{t=1}^\infty \sum_{j=1}^{t-1} \frac{\delta}{j^3} &(\mbox{by cancelling common terms}) \\
    &\le \sum_{t=1}^\infty \int_{0}^t \frac{\delta}{j^3}dj \\ 
    &= \sum_{t=1}^\infty \frac{\delta}{2t^2} \\
    &= \frac{\delta \pi^2}{12}.
\end{align*}
Thus $\bbP (\calE) \ge 1-\frac{\delta \pi^2}{12}$. Now, consider $ L_T$ under event $\calE$.
 Recall
our assumption that $\forall k$, $\mu_k \in [0,1]$ which allows us to bound part $A_1$ by replacing
each term by $1$. Thus we get, with probability at least $1-\frac{\delta\pi^2}{12}$,
\begin{align*}
    L_T&\le \sum_{k=1}^K \bigg (B \cdot N_{T,k} + \sum_{x_t = k, \ell_t = 1} 1 + \sum_{x_t = k, \ell_t = 0} \sqrt{\frac{2\sigma^2\log(2N_t^3/\delta)}{N_{t-1,k}}}  \bigg)\\
    &\le \sum_{k=1}^K \bigg( (B+1)\cdot N_{T,k} + \sum_{x_t = k, \ell_t = 0}   B^{\beta}M_{t,k}^\alpha \bigg) 
\end{align*}
where the second inequality is because: $\ell_t = 0$ (not labeled)
indicates our current confidence radius $\sqrt{\frac{2\sigma^2\log(2N_t^3/\delta)}{N_{t-1,k}}}$ is less
than the threshold 
$B^{\beta}M_{t,k}^\alpha $. Next we bound $\sum_{x_t = k, \ell_t = 0}  B^{\beta}M_{t,k}^\alpha$ by the integral from $t = 0$ to $M_{T,k}$,
\begin{align*}
    L_T &\le \sum_{k=1}^K \bigg( (B+1)\cdot N_{T,k} + \int_0^{M_{T,k}}  B^\beta t^\alpha dt \bigg) \\
    &= \sum_{k=1}^K \bigg( (B+1)\cdot N_{T,k} + \frac{ B^\beta }{\alpha+1}M_{T,k}^{\alpha+1} \bigg)
\end{align*}
To bound $N_{T,k}$, we use the same strategy as Theorem 2 in~\cite{Audibert+MS:2009}: $\forall u \in \bbN_+$, we define $S_k = \{t: x_t = k,  N_{t-1,k} \ge u\}$, recall the definition of $N_{T,k}$,
\begin{align} \label{eqn: n_tk}
    N_{T,k} = \sum_{t=1}^\infty \mathds{1}_{\{x_t = k, l_t = 1\}}
    \le u + \sum_{t: N_{t-1,k} \ge u}^\infty \mathds{1}_{\{x_t = k, \ell_t = 1\}}
    = u + \sum_{t \in S_k} \mathds{1}_{\bigg\{  \sqrt{\frac{2\sigma^2\log(2N_t^3/\delta)}{N_{t-1,k}}} > B^\beta M_{t,k}^\alpha\bigg\}} 
\end{align}
where the last equality is because $\ell_t=1$ indicates the confidence radius exceeds the threshold. Then by rearranging the terms in~\eqref{eqn: n_tk} and using the fact that $N_{t-1,k} \ge u$ in $S_k$, we get
\begin{align}\label{eqn:thm1_ntk}
    N_{T,k} \le u + \sum_{t \in S_k} \mathds{1}_{\bigg \{u <
\frac{2\sigma^2\log(2N_t^3/\delta)}{B^{2\beta}M_{t,k}^{2\alpha}} \bigg\}}.
\end{align}
We let $u = 2\sigma^2\log(2N_T^3/\delta) B^{-\frac{2}{3}}M_{T,k}^{\frac{2}{3}}$, $\beta = \frac{1}{3}$ and $\alpha = -\frac{1}{3}$, since we would like to choose $u$ such that $u \ge \frac{2\sigma^2\log(2N_t^3/\delta)}{B^{2\beta}M_{t,k}^{2\alpha}}$ and then we can bound $N_{T,k}$ by $u$. Recall that $M_{T,k} \ge M_{t,k}, N_T \ge N_t, \forall t\le T$, we get
\begin{align*}
    u B^{2\beta}M_{t,k}^{2\alpha} = 2\sigma^2\log(2N_T^3/\delta) B^{-\frac{2}{3}}M_{T,k}^{\frac{2}{3}} \cdot B^{\frac{2}{3}}M_{t,k}^{-\frac{2}{3}} \ge 2\sigma^2\log(2N_t^3/\delta)
\end{align*}
The inequality indicates that each term inside the summation of~\eqref{eqn:thm1_ntk} becomes $0$, thus $N_{T,k} \le u$. Since $N_{T,k} \le T$, finally we get
\begin{align*}
    L_T \le \sum_{k=1}^K \left( (B^{\frac{1}{3}} + B^{-\frac{2}{3}}) 2\sigma^2\log(2T^3/\delta) M_{T,k}^{\frac{2}{3}} + \frac{3}{2} B^{\frac{1}{3}} M_{T,k}^{\frac{2}{3}} \right)
\end{align*}

\end{proof}

\subsection{Proof of Theorem~\ref{thm:mablower}}
In this proof, we follow the standard procedure of Le Cam's method (Theorem~\ref{thm:lecam})
to derive the minimax lower bound. 
We include the proof here for completeness.
First, recall the formulation of our expected loss:
\begin{align*}
    \bbE L_T &=  \sum_{k=1}^K \bbE\bigg (B \cdot N_{T,k} + \sum_{x_t = k, t=1}^{T} \big|\mu_k - p_t\big| \bigg ).
\end{align*}
Since we are looking at the infimum over all problems, we focus on using Gaussian distributions for all the types. For each distribution $k \in [K]$, suppose there are $n_k$ i.i.d. samples $x_1, \cdots, x_{n_k}$ from $\nu_k = \calN(\mu_k, \sigma^2)$. In order to lower bound the minimax risk of estimating $\mu_k$ by any estimator $\widehat{\mu}_k$ under the metric $\rho  = |\mu_k - \widehat{\mu}_k|$, we consider two possible distributions for $\nu_k$: fix $\delta \in [0, \frac{1}{4}]$, $P_{1,k} = \calN(\mu_k'+\delta, \sigma^2)$ and $P_{2,k} = \calN(\mu_k'-\delta, \sigma^2)$ where $\mu_k' \sim \mbox{Uniform}(\frac{1}{4}, \frac{3}{4})$. It's easy to see that $\mu_k'+\delta$ and $\mu_k'-\delta$ are $2\delta$ apart.

Via Le Cam's Method, suppose we have an equal probability to choose between $P_{1,k}$ and $P_{2,k}$, we consider all tests $\Psi : \calX \to \{1,2\}$ and get
\begin{align*}
    \inf_{\widehat{\mu}_k} \sup_{\nu_k} \bbE |\mu_k - \widehat{\mu}_k| &\ge \delta \inf_{\Psi} \bigg\{\frac{1}{2}P_{1,k}(\Psi(x_1, \cdots, x_{n_k}) \neq 1) + \frac{1}{2}P_{2,k}(\Psi(x_1, \cdots, x_{n_k}) \neq 2)\bigg\} \\
    &= \frac{\delta}{2} \Big[1-\|P_{1,k}^{n_k} - P_{2,k}^{n_k}\|_{TV}\Big]
\end{align*}
where $P_{i,k}^{n_k}$ is the product distribution for $i = 1,2$ and   $\|P_{1,k}^{n_k} - P_{2,k}^{n_k}\|_{TV}$ is the total variation distance between $P_{1,k}^{n_k}$ and $P_{2,k}^{n_k}$. By Pinsker's inequality and the chain rule of KL-divergence~\cite{vershynin2018high},
\begin{align*}
    \|P_{1,k}^{n_k} - P_{2,k}^{n_k}\|_{TV}^2 \le \frac{1}{2} D_{KL}(P_{1,k}^{n_k}\|P_{2,k}^{n_k}) 
    = \frac{n}{2} D_{KL}(P_{1,k}^{n_k}\|P_{2,k}^{n_k}) 
    = \frac{n}{2} \cdot \frac{(2\delta)^2}{2\sigma^2} = \frac{n\delta^2}{\sigma^2}
\end{align*}
Thus $\|P_{1,k}^{n_k}-P_{2,k}^{n_k}\|_{TV} \le \frac{\sqrt{n_k}\delta}{\sigma}$.  Taking $\delta = \frac{\sigma}{2\sqrt{n_k}}$ guarantees that $\|P_{1,k}^{n_k}-P_{2,k}^{n_k} \|\le \frac{1}{2}$. Thus
\begin{align*}
    \inf_{\widehat{\mu}_k} \sup_{\nu_k} \bbE |\mu_k - \widehat{\mu}_k| \ge \frac{\delta}{2}(1-\frac{1}{2}) = \frac{\delta}{4} = \frac{\sigma}{8\sqrt{n_k}}.
\end{align*}
Hence when considering all $K$ types, we get
\begin{align*}
    \inf\sup \bbE L_T \ge \sum_{k=1}^K\bbE\bigg(B N_{T,k} + M_{T,k}\cdot\frac{\sigma}{8\sqrt{ N_{T,k}}}\bigg)
\end{align*}
We tune the values of $N_{T,k}$ by letting $BN_{T,k} = \frac{\sigma M_{T,k}}{8\sqrt{N_{T,k}}}$, so $N_{T,k} = (\frac{\sigma}{8B})^{\frac{2}{3}} M_{T,k}^{\frac{2}{3}}$. Thus we get
\begin{align*}
    \inf\sup\bbE L_T \ge \sum_{k=1}^K \frac{1}{4}\sigma^{\frac{2}{3}}B^{\frac{1}{3}} M_{T,k}^{\frac{2}{3}}.
\end{align*}

\subsection{Proof of Theorem~\ref{thm:gp-anytime}}
\begin{proof}
In this proof, similarly as the idea in Section~\ref{sec:proof_1}, we will first decompose
the prediction error into rounds that are labeled ($\ell_t = 1$) and rounds that are not labeled
($\ell_t = 0$) and then bound each part separately.
Finally, we will bound $N_T$ for the SE kernel and Mat\'ern kernel separately to bound the labeling cost.
Recall the definition of the cumulative loss,  $L_T$ is given by,
\begin{align} \label{gp:eqn1}
    L_T = BN_T + \sum_{t = 1}^T |f(x_t)-\mu_t(x_t)|.
\end{align}
where $p_t = \mu_{t}(x_t)$ in the GP setting. 
Theorem 6 in~\cite{DBLP:conf/icml/SrinivasKKS10} states that: let $\delta \in (0,1)$ and $\beta_t = 2\|f\|_k^2 + 300 \gamma_t \ln^3(t/\delta)$,
\begin{align*} 
    \bbP \left \{\forall T, \forall x \in \calX, |\mu_T(\bx) - f(\bx)| \le \beta_{T+1}^{1/2}\sigma_T(\bx) \right \} \ge 1-\delta.
\end{align*}
Thus with probability at least $1-\delta$, \eqref{gp:eqn1} is upper bounded by
\begin{align} \label{gp:eqn2}
    L_T &\le B N_T + \sum_{t=1}^T \beta_{t+1}^{1/2} \sigma_t(x_t) = \underbrace{B N_T}_{B_1} + \underbrace{\sum_{\ell_t = 1}\beta_{t+1}^{1/2} \sigma_t(x_t)}_{B_2} + \underbrace{\sum_{\ell_t = 0} \beta_{t+1}^{1/2} \sigma_t(x_t)}_{B_3},
\end{align}
where the equality comes from the decomposition of labeled rounds and not labeled rounds.  We consider bounding the prediction errors $B_2$ and $B_3$ first.
To bound part $B_2$, we follow the same idea and use part of the results as Lemma 5.4 in~\citep{DBLP:conf/icml/SrinivasKKS10},
\begin{align*}
    \beta_{t+1} \sigma_t^2(x_t) &\le \beta_{T+1} \sigma^2 (\sigma^{-2} \sigma_t^2(x_t)) \\
    &\le \beta_{T+1} \sigma^2 C_1 \log(1+\sigma^{-2} \sigma_t^2(x_t)) 
\end{align*}
with $C_1 = \sigma^{-2}/\log(1+\sigma^{-2}) \ge 1$, since $s^2 \le C_1 \log(1+s^2)$ for $s \in [0, \sigma^{-2}]$, and $\sigma^{-2} \sigma_{t}^2(x_t) \le \sigma^{-2} k(x_t, x_t) \le \sigma^{-2}$ where $k(x_t, x_t)$ is the kernel. Summarized over all the labeled rounds, we get
\begin{align*}
    \sum_{\ell_t = 1} \beta_{t+1} \sigma_t^2(x_t) \le \sum_{\ell_t = 1} \beta_{T+1} \sigma^2 C_1 \log(1+\sigma^{-2} \sigma_t^2(x_t)) 
\end{align*}
Recall the definition of MIG~\eqref{eqn:mig}, we get
\begin{align*}
    \sum_{\ell_t = 1} \beta_{T+1} \sigma^2 C_1 \log(1+\sigma^{-2} \sigma_t^2(x_t))  \le \frac{2\beta_{T+1}}{\log(1+\sigma^{-2})} \gamma_{N_T}.
\end{align*}
By the Cauchy-Schwarz inequality, 
\begin{align} \label{sup:labeled_bound}
    \sum_{\ell_t=1}\beta_{t+1}^{1/2}\sigma_t(x_t) \le \sqrt{N_T\sum_{\ell_t=1}\beta_{t+1}\sigma_t^2(x_t)} \le \sqrt{\frac{2\beta_{T+1} \gamma_{N_T} N_T}{\log(1+\sigma^{-2})} }. 
\end{align}
For parts $B_1$ and $B_3$, we analyze the SE kernel and Mat\'ern kernel separately since different kernels have different threshold functions.

\subsubsection*{Squared Exponential (SE) kernel}
Since we decide not to label when $\sigma_{t-1}(x_t) \le \tau(t)$ and recall that the threshold function for SE kernel is $\tau(t) = \sqrt{2\sigma^2} B^{\frac{1}{d+3}} t^{-\frac{1}{d+3}}$, we get to bound part $B_3$:
\begin{align*} 
    \sum_{\ell_t = 0} \beta_{t+1}^{1/2}\sigma_t(x_t) &\le \beta_{T+1}^{1/2}\sum_{\ell_t = 0} \sigma_{t-1}(x_t) & (\mbox{by definition of }\beta \mbox{ and~\eqref{eqn:gp_update_sigma}}) \\
    &\le \beta_{T+1}^{1/2} \int_{t=0}^T  \sqrt{2\sigma^2} B^{\frac{1}{d+3}} t^{-\frac{1}{d+3}} dt \\
    &= \frac{\beta_{T+1}^{1/2}\sqrt{2\sigma^2} B^{\frac{1}{d+3}}(d+3)}{d+2}T^{\frac{d+2}{d+3}}.
\end{align*}
In order to bound part $B_1$, we discretize $\calX = [0,1]^d$ into $K^d$ equal-size hypercubes $A_1, \cdots, A_{K^d}$ whose side lengths are $\frac{1}{K}$. $\forall A_i$, their $L_2$ diameters are $\rm{diam}(A_i)$ $=\frac{\sqrt{d}}{K}$. Let $N_{t,k}$ denote the number of labels after time $t$ in $A_k$ and $M_{t,k}$ denote the number of points after time $t$ in $A_k$. $\forall k \in \{1, \cdots, K^d\}$ and $\forall u \in \bbR_+$,
\begin{align} \label{eqn:gp_ntk}
    N_{T,k} \le u + \sum_{t \in S_k} \mathds{1}_{\{\sigma_{t-1}(x_t) > \tau(t)\}}
\end{align}
where $S_k = \{t:x_t\in A_k, N_{t-1,k} \ge u\}$. Let $u = B^{-\frac{2}{d+3}}T^{\frac{2}{d+3}}$ and
$K = \sqrt{\frac{d/l^2}{\sigma^2}} B^{-\frac{1}{d+3}} T^{\frac{1}{d+3}}$, where $l$ is
from~\eqref{eqn:kerneldefns}.
Using the results from Lemma~\ref{lemma} and we get
\begin{align*}
    \sigma_{t-1}^2(x_t) &\le  \frac{d/l^2}{K^2} + \frac{\sigma^2}{N_{t-1,k}} \\
    &\le \frac{d/l^2}{K^2} + \frac{\sigma^2}{u} & (\mbox{since } N_{t-1,k} \ge u)\\
    &\le \frac{2\sigma^2}{u} & (\mbox{by definitions of }K \mbox{ and } u) \\
    &= \frac{2\sigma^2}{B^{-\frac{2}{d+3}} T^{\frac{2}{d+3}}}, 
\end{align*}
which indicates $\sigma_{t-1}(x_t) \le \sqrt{2\sigma^2} B^{\frac{1}{d+3}} T^{-\frac{1}{d+3}} \le \tau(t)$. Thus the terms inside the summation of~\eqref{eqn:gp_ntk} becomes $0$ and we can bound $N_T$ by,
\begin{align*}
    N_T \le K^d u = \left(\sqrt{\frac{d/l^2}{\sigma^2}}\right)^d B^{-\frac{d+2}{d+3}}T^{\frac{d+2}{d+3}}.
\end{align*}
Combining the results for part $B_1$, $B_2$ and $B_3$ to~\eqref{gp:eqn2}, we get
\begin{align*}
    L_T &\le \left(\sqrt{\frac{d/l^2}{\sigma^2}}\right )^d B^{\frac{1}{d+3}} T^{\frac{d+2}{d+3}} + \sqrt{\frac{2\beta_{T+1} \gamma_{N_T}N_T}{\log(1+\sigma^{-2})} } 
    + \frac{\beta_{T+1}^{1/2} \sqrt{2}\sigma(d+3)}{d+2} B^{\frac{1}{3+d}} T^{\frac{d+2}{d+3}} \\
    &= \left(C_1 + \frac{\beta_{T+1}^{1/2} \sqrt{2}\sigma(d+3)}{d+2} \right)B^{\frac{1}{3+d}} T^{\frac{d+2}{d+3}} + \sqrt{\frac{2\beta_{T+1}}{\log(1+\sigma^{-2})} \gamma_{C_1 B^{-\frac{d+2}{d+3}}T^{\frac{d+2}{d+3}}} C_1 B^{-\frac{d+2}{d+3}}T^{\frac{d+2}{d+3}}}
\end{align*}
where $C_1 = \left (\sqrt{\frac{d/l^2}{\sigma^2}}\right )^d$.

\subsubsection*{Mat\'{e}rn Kernel}
Recall the threshold function for Mat\'{e}rn kernel, $\tau(t) = \sqrt{2\sigma^2} B^{\frac{1}{2d+3}} t^{-\frac{1}{2d+3}}$, we can bound part $B_3$ by,
\begin{align*} 
    \sum_{\ell_t = 0} \beta_{t+1}^{1/2}\sigma_t(x_t) &\le \beta_{T+1}^{1/2}\sum_{\ell_t = 0} \sigma_{t-1}(x_t) & (\mbox{by definition of }\beta \mbox{ and~\eqref{eqn:gp_update_sigma}}) \\
    &\le \beta_{T+1}^{1/2} \int_{t=0}^T  \sqrt{2\sigma^2} B^{\frac{1}{2d+3}} t^{-\frac{1}{2d+3}} dt \\
    &= \frac{\beta_{T+1}^{1/2}\sqrt{2\sigma^2} B^{\frac{1}{2d+3}}(2d+3)}{2d+2}T^{\frac{2d+2}{2d+3}}.
\end{align*}
We follow the same discretization argument in SE kernel but let $u = B^{-\frac{2}{2d+3}}T^{\frac{2}{2d+3}}$ and $K = \frac{2L_{\rmM} \sqrt{d}}{\sigma^2} B^{-\frac{2}{2d+3}}T^{\frac{2}{2d+3}}$, where $L_{\rmM}$ is from~\eqref{eqn:kerneldefns}. By Lemma~\ref{lemma}, we get
\begin{align*}
    \sigma^2_{t-1}(x_t) &\le \frac{2L_{\rmM}\sqrt{d}}{K} + \frac{\sigma^2}{N_{t-1,k}} \\ 
    &\le \frac{2L_{\rmM}\sqrt{d}}{K} + \frac{\sigma^2}{u} \\
    &\le \frac{2\sigma^2}{u} \\
    &=\frac{2\sigma^2}{B^{-\frac{2}{2d+3}}T^{\frac{2}{2d+3}}}
\end{align*}
which indicates $\sigma_{t-1}(x_t) \le \sqrt{2\sigma^2} B^{\frac{1}{2d+3}} T^{-\frac{1}{2d+3}} \le \tau(t)$. Thus the terms inside the summation of~\eqref{eqn:gp_ntk} becomes $0$ and we can bound $N_T$ by,
\begin{align*}
    N_T \le K^d u = \left(\frac{2L_{\rmM} \sqrt{d}}{\sigma^2}\right)^d B^{-\frac{2d+2}{2d+3}}T^{\frac{2d+2}{2d+3}}
\end{align*}
Combining the results for part $B_1$, $B_2$, and $B_3$ to~\eqref{gp:eqn2}, we get
\begin{align*}
    L_T &\le \left(\frac{2L_{\rmM} \sqrt{d}}{\sigma^2}\right)^d B^{\frac{1}{2d+3}}T^{\frac{2d+2}{2d+3}} + \sqrt{\frac{2\beta_{T+1} \gamma_{N_T} N_T}{\log(1+\sigma^{-2})} } + \frac{\beta_{T+1}^{1/2}\sqrt{2}\sigma (2d+3)}{2d+2}B^{\frac{1}{2d+3}}T^{\frac{2d+2}{2d+3}} \\
    &= \left(C_2 + \frac{\beta_{T+1}^{1/2} \sqrt{2}\sigma(2d+3)}{2d+2} \right) B^{\frac{1}{3+2d}} T^{\frac{2d+2}{2d+3}} + \sqrt{\frac{2\beta_{T+1}}{\log(1+\sigma^{-2})} \gamma_{C_2 B^{-\frac{2d+2}{2d+3}}T^{\frac{2d+2}{2d+3}} } C_2 B^{-\frac{2d+2}{2d+3}}T^{\frac{2d+2}{2d+3}} }
\end{align*}
where $C_2 = \left( \frac{2L_{\rmM} \sqrt{d}}{\sigma^2}\right)^d$.

\end{proof}

\section{Values of $\lambda$} \label{sup:sec_lambda}

Table~\ref{tbl:discrete}, \ref{tbl:synthetic}, \ref{tbl:parkinsons}, \ref{tbl:supernova} provide the values of $\lambda$ used for Algorithm~\ref{alg:mab} and Algorithm~\ref{alg:gpucb} in all the experiments.

\begin{table}[h!]
\centering
\begin{tabular}{ |c|c|c|c| } 
 \hline
 B = 10, uniform & B = 100, uniform & B = 10, uniform & B = 100, uniform \\
 \hline
 0.5 & 0.25 & 2 & 0.75 \\ 
 \hline
 B = 10, lopsided & B = 100, lopsided & B = 10, lopsided & B = 100, lopsided \\
 \hline
 0.5 & 0.25 & 2 & 0.5 \\ 
 \hline
\end{tabular}
\caption{$\lambda$ used for Fig~\ref{fig:discrete}. $K = 10$ for columns 1 and 2. $K = 100$ for columns 3 and 4.}
\label{tbl:discrete}
\end{table}

\begin{table}[h!]
    \centering
    \begin{tabular}{|c|c|c|c|c|}
        \hline 
         & B = 10, uniform & B = 100, uniform & B = 10, lopsided & B = 100, lopsided\\
        \hline
         Algorithm~\ref{alg:gpucb} & 1 & 1 & 0.5 & 0.5  \\
         \hline
         Algorithm~\ref{alg:mab} & 10 & 10 & 10 & 10 \\
         \hline
    \end{tabular}
    \caption{$\lambda$ used for Fig~\ref{fig:gp_synthetic}, Column 1 and 2 are used for the Branin function. Columns 3 and 4 are used for the Hartmann function.}
    \label{tbl:synthetic}
\end{table}

\begin{table}[h!]
    \centering
    \begin{tabular}{|c|c|c|c|}
        \hline
         & B = 10, uniform & B = 10, lopsided & B = 100, lopsided \\
        \hline
         Algorithm~\ref{alg:gpucb}& 5 & 5 & 5  \\
         \hline
         Algorithm~\ref{alg:mab}& 5 & 5 & 5 \\
         \hline
    \end{tabular}
    \caption{$\lambda$ used for Fig~\ref{fig:parkinsons}. }
    \label{tbl:parkinsons}
\end{table}

\begin{table}[h!]
    \centering
    \begin{tabular}{|c|c|c|c|}
        \hline
         & B = 1 & B = 10 & B = 100  \\
         \hline
         Algorithm~\ref{alg:gpucb}& 1 & 1 & 1 \\
         \hline
         Algorithm~\ref{alg:mab} & 50 & 30 & 10 \\
         \hline
     \end{tabular}
    \caption{$\lambda$ used for Fig~\ref{fig:supernova}}
    \label{tbl:supernova}
\end{table}

\section{Plots of error} \label{sup:sec_errorplots}

In this section, we provide the average prediction error plots for all the experiments.

\begin{figure}[ht!]
\centering
\begin{tabular}{cccc}
    \includegraphics[width = 0.23\textwidth]{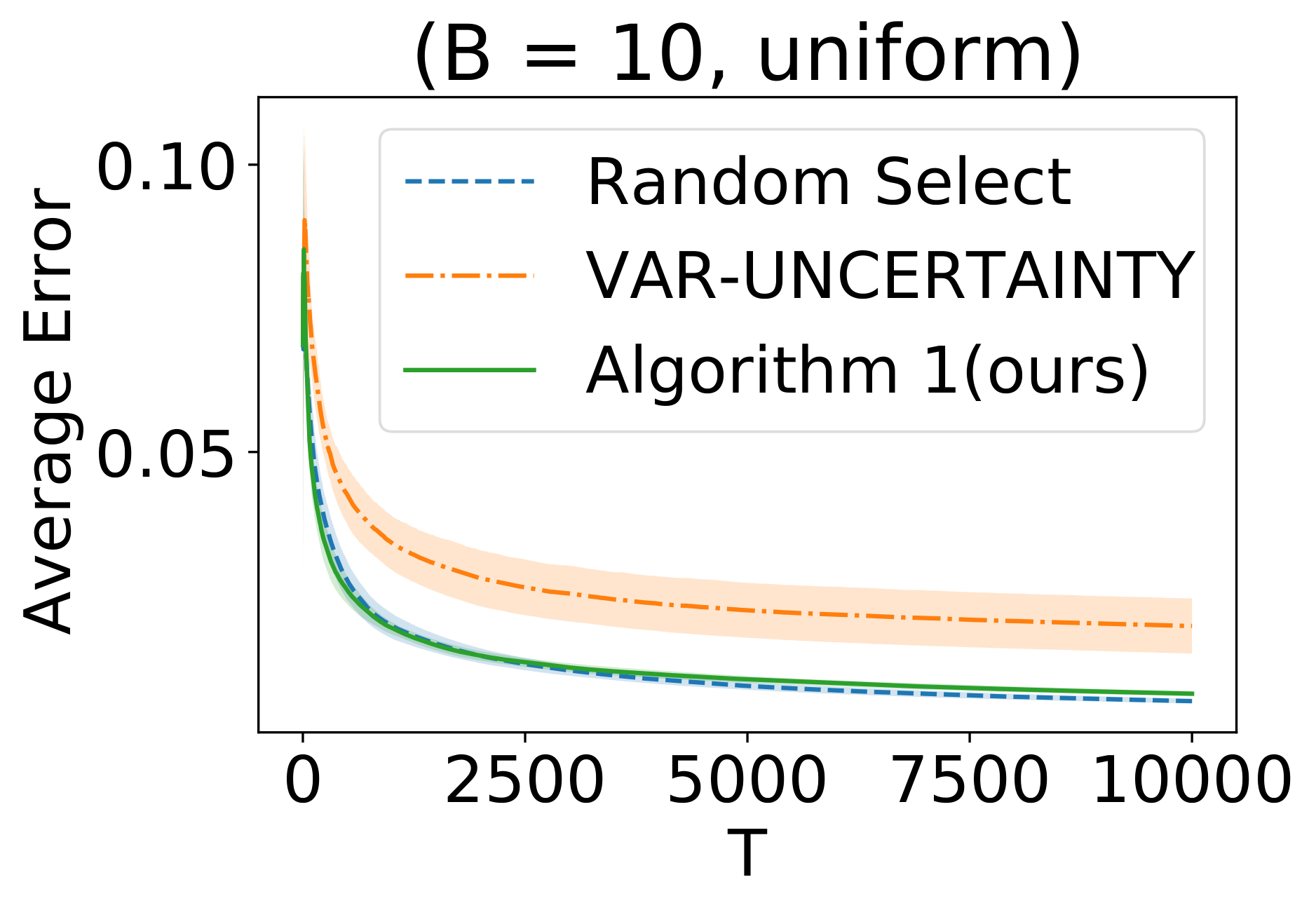} &
    \includegraphics[width = 0.22\textwidth]{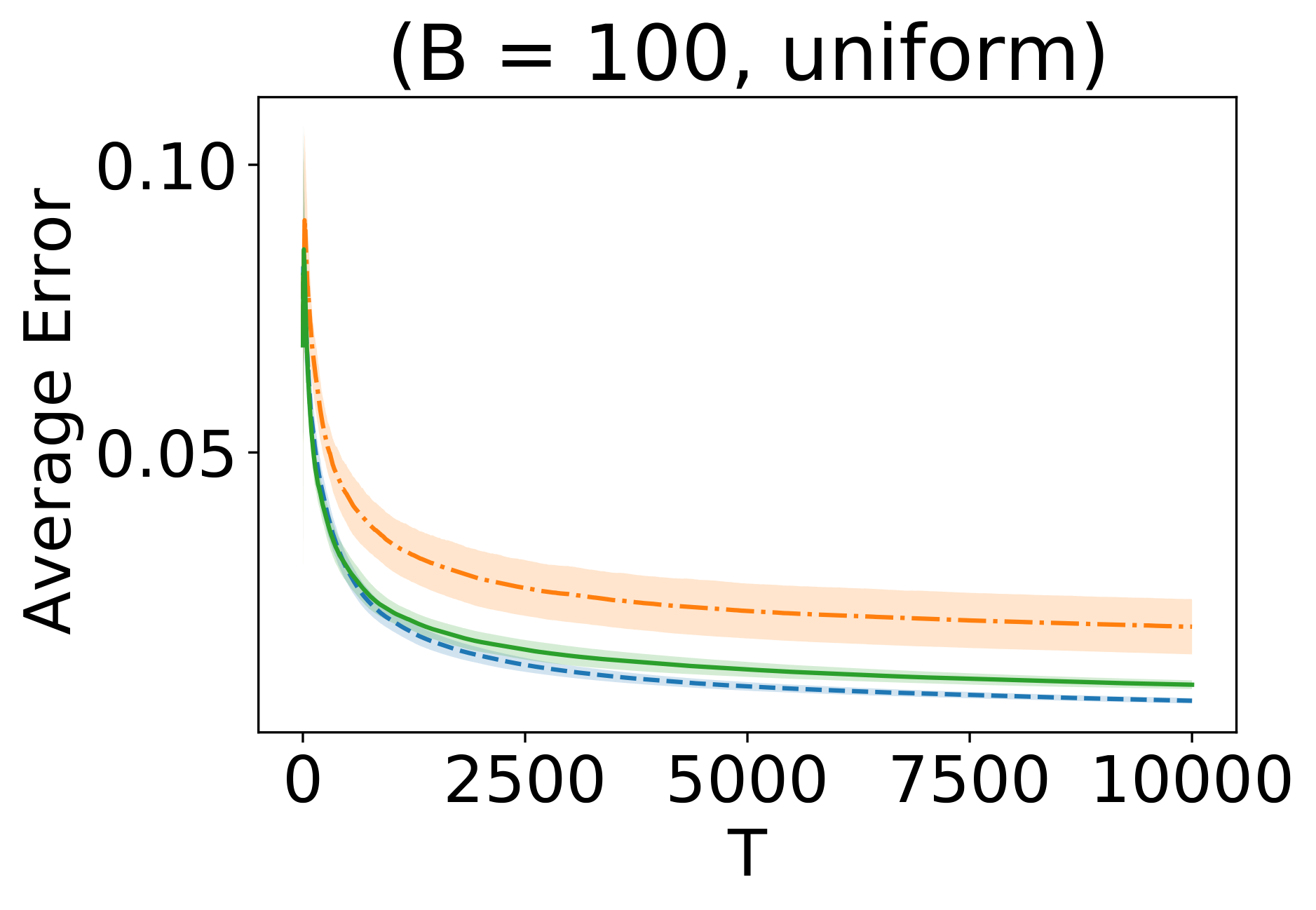} &
    \includegraphics[width = 0.22\textwidth]{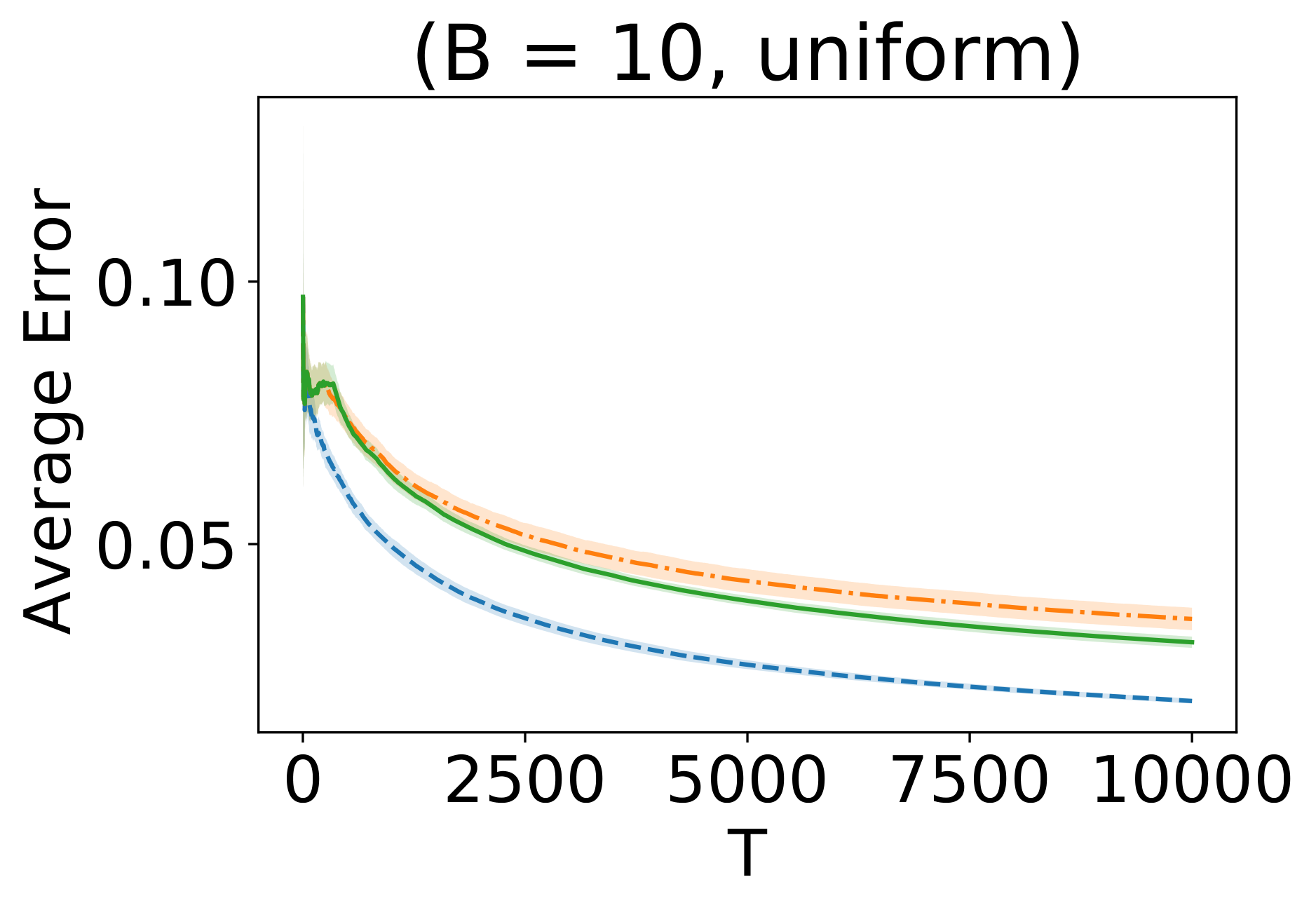} &
    \includegraphics[width = 0.23\textwidth]{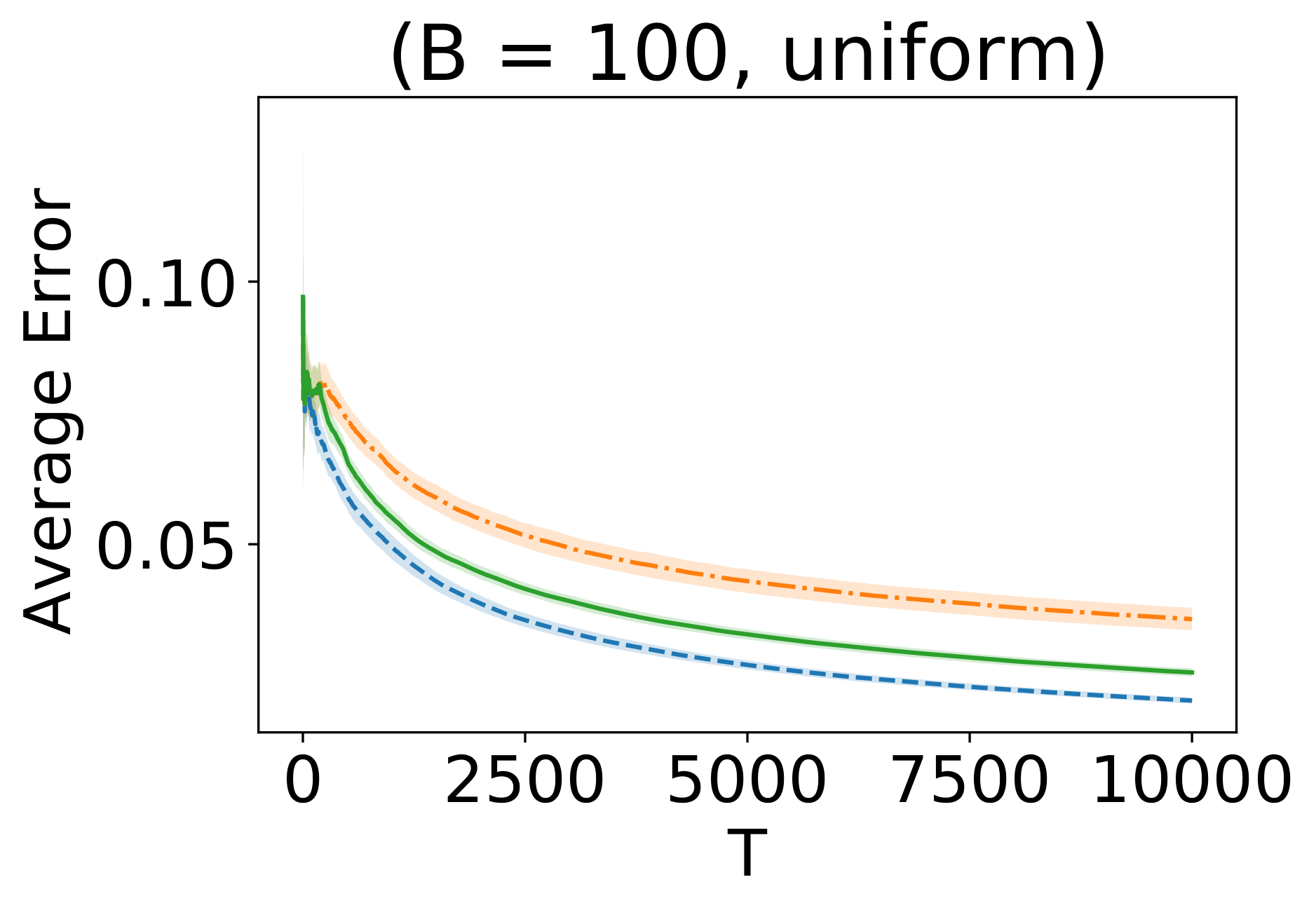}
     \\
     \includegraphics[width = 0.22\textwidth]{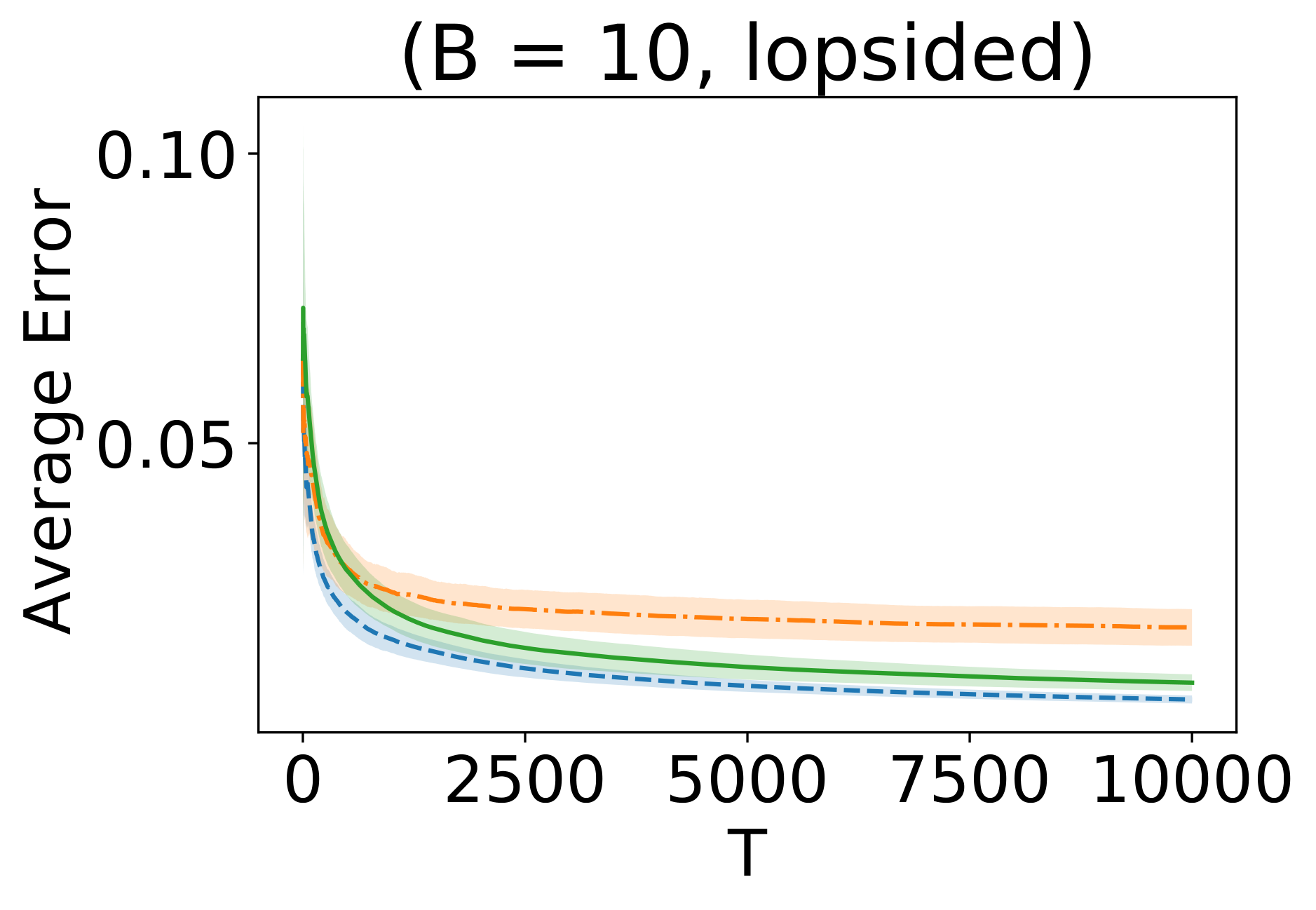}&
     \includegraphics[width = 0.223\textwidth]{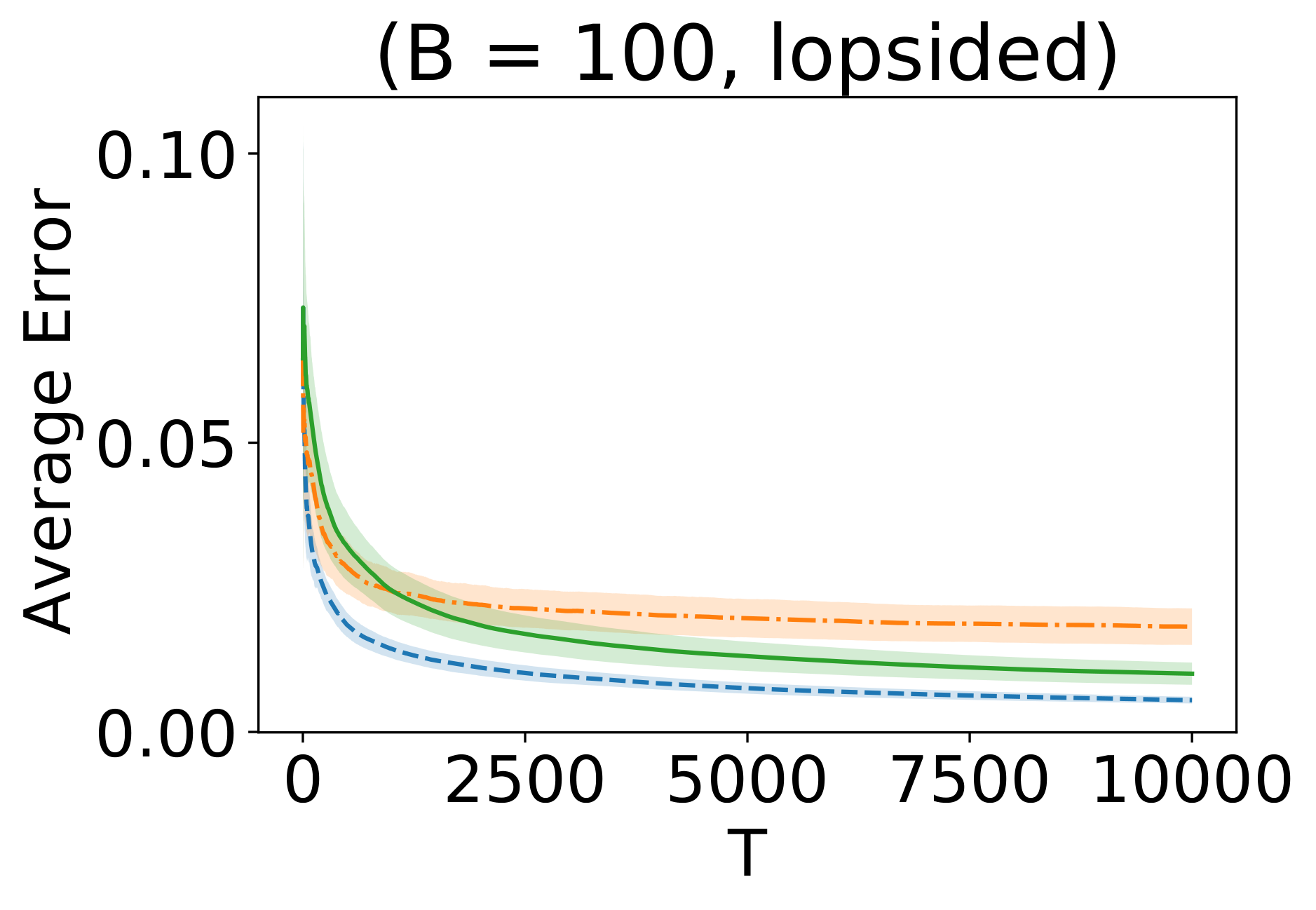}&
    \includegraphics[width = 0.22\textwidth]{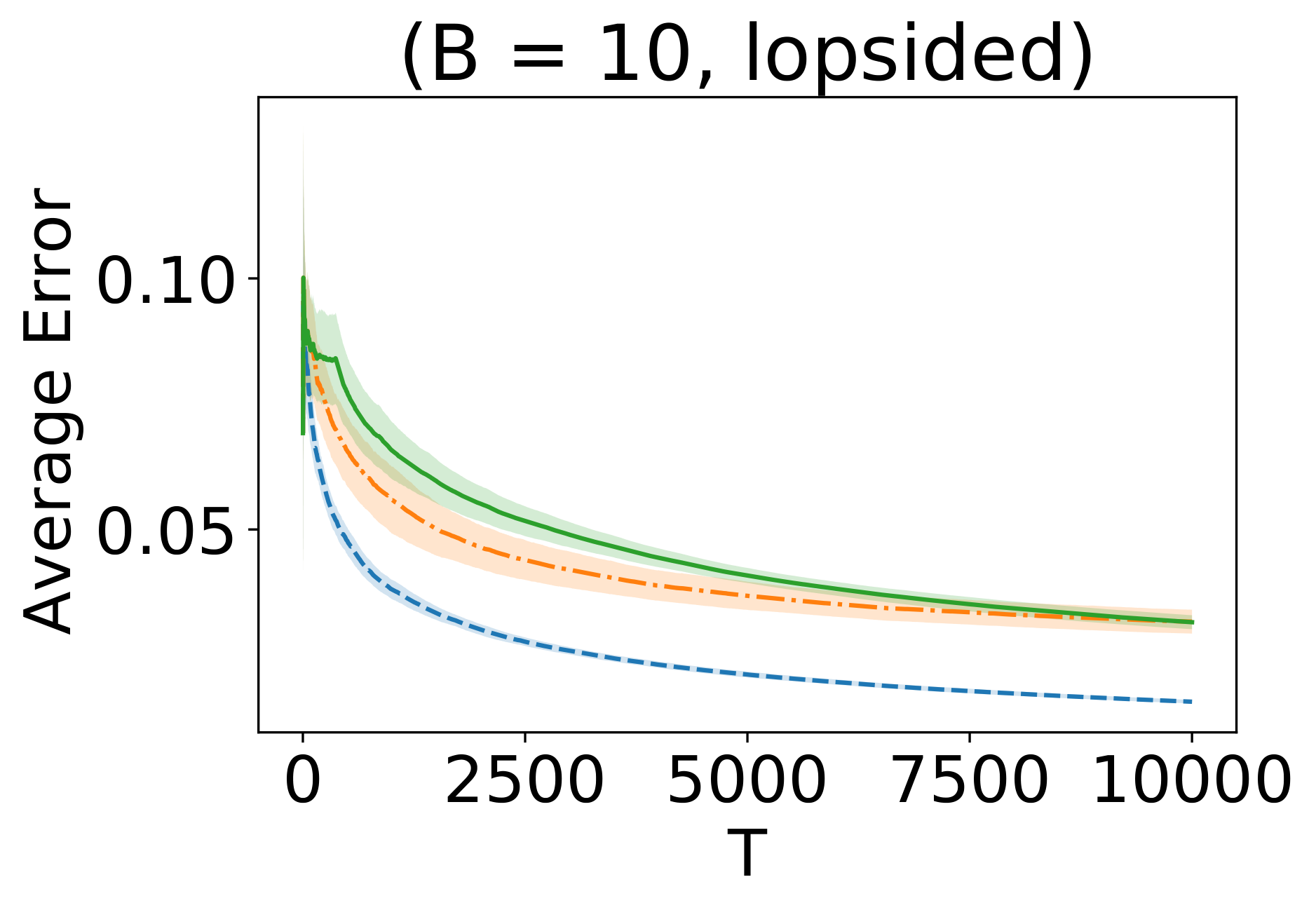} &
     \includegraphics[width = 0.23\textwidth]{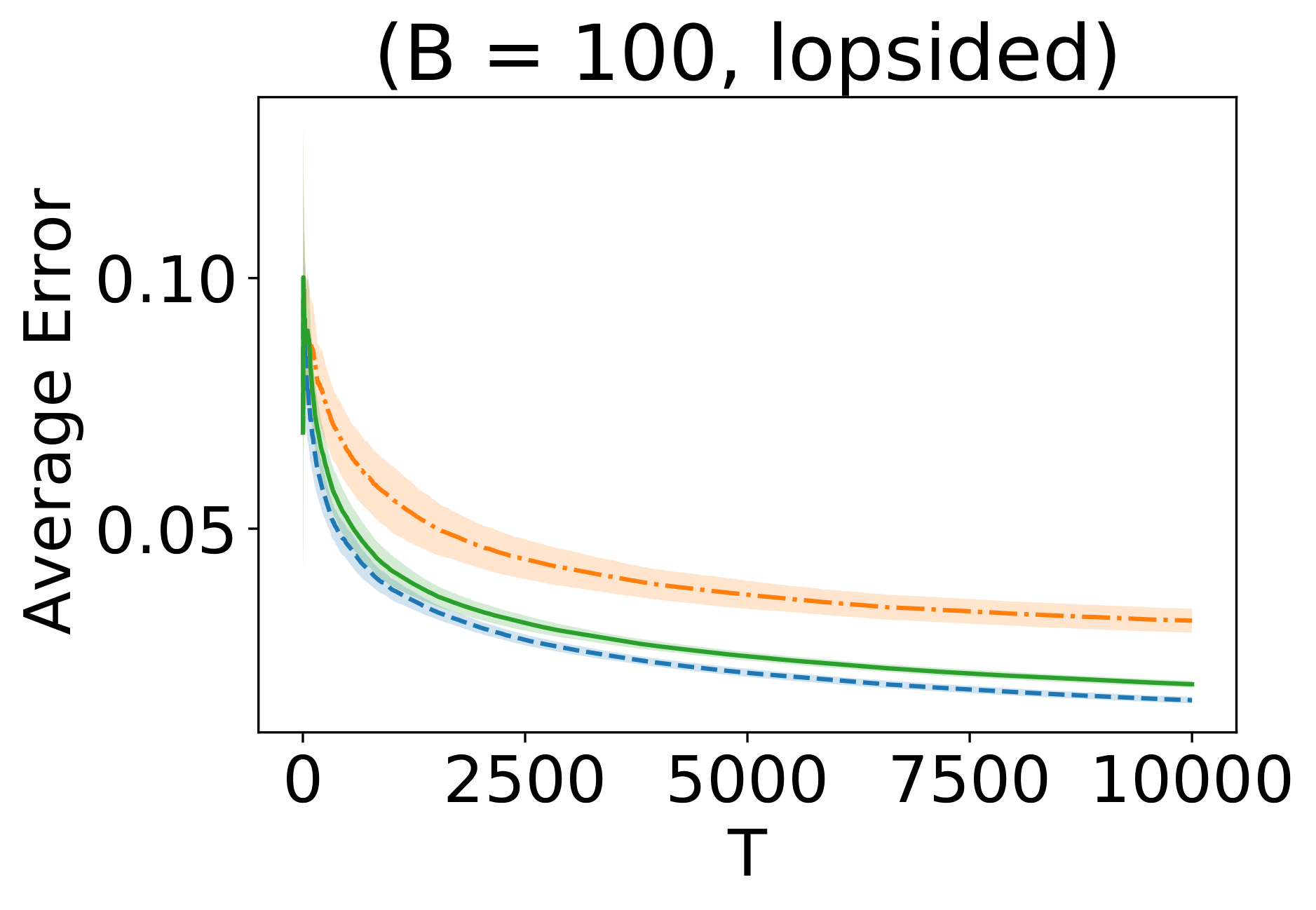}
\end{tabular}
\vspace{-0.1in}
\caption{\small Error plots comparing the three methods for $K$ discrete setting
of~\S\ref{sec:expsynK}. $K = 10$ for column 1 and
2. $K = 100$ for columns 3 and 4. The results are averaged over 10 trials, and we report the $95\%$
the confidence interval of the standard error.}
\label{error:discrete}
\end{figure}

\begin{figure}[ht!] 
\centering
\begin{tabular}{cccc}
     \includegraphics[width = 0.23\textwidth]{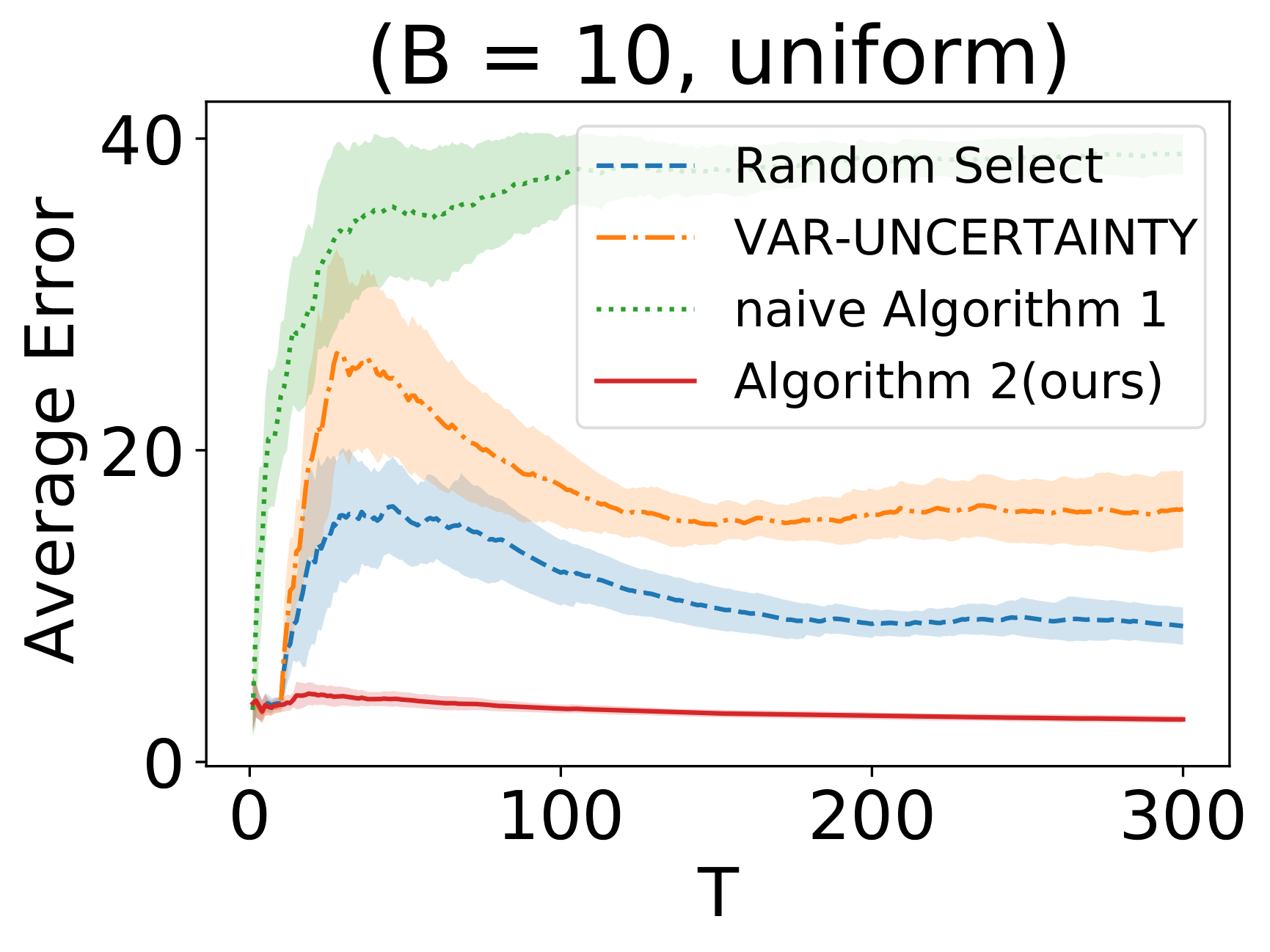} &
     \includegraphics[width = 0.23\textwidth]{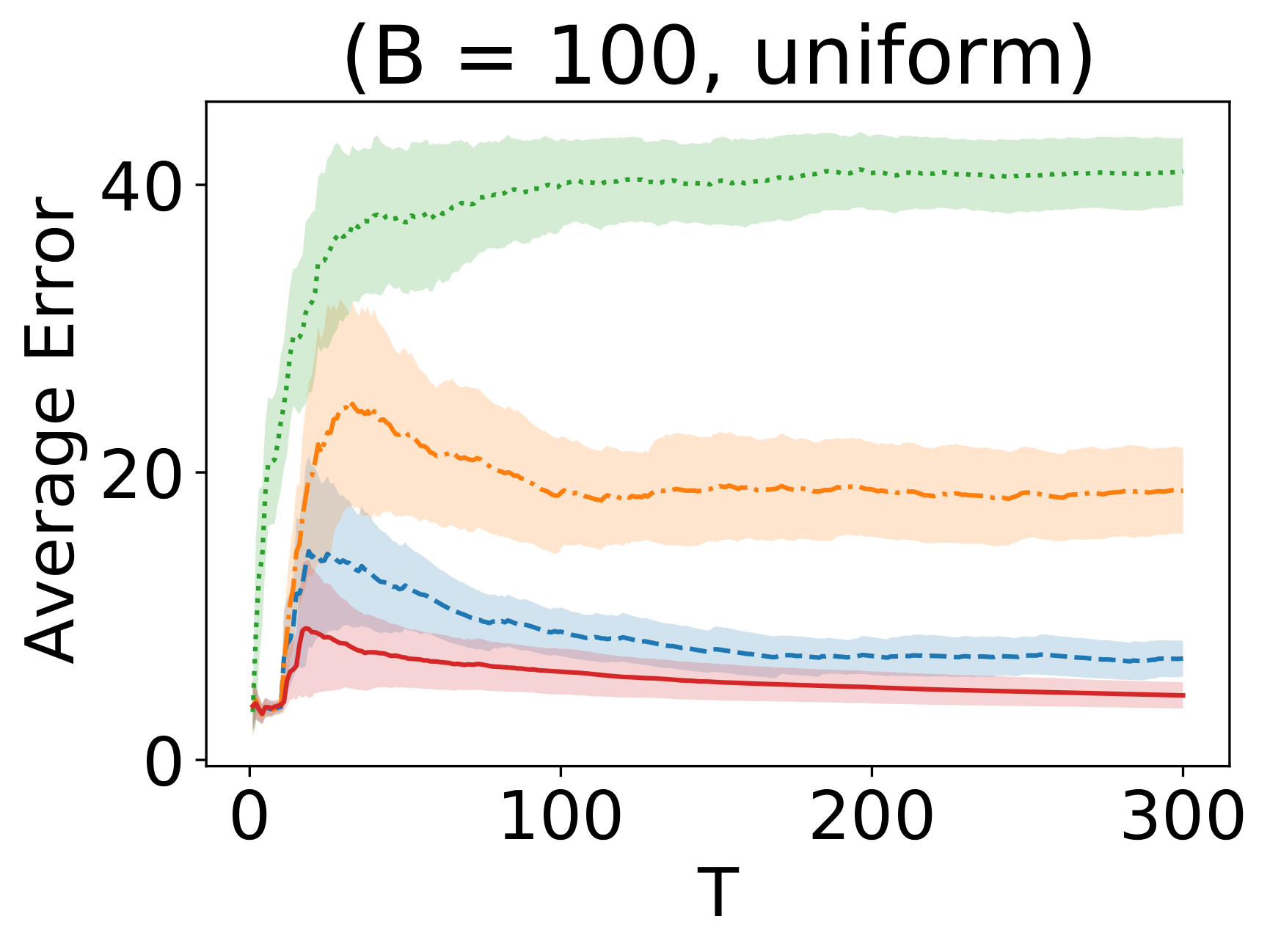} &
     \includegraphics[width = 0.23\textwidth]{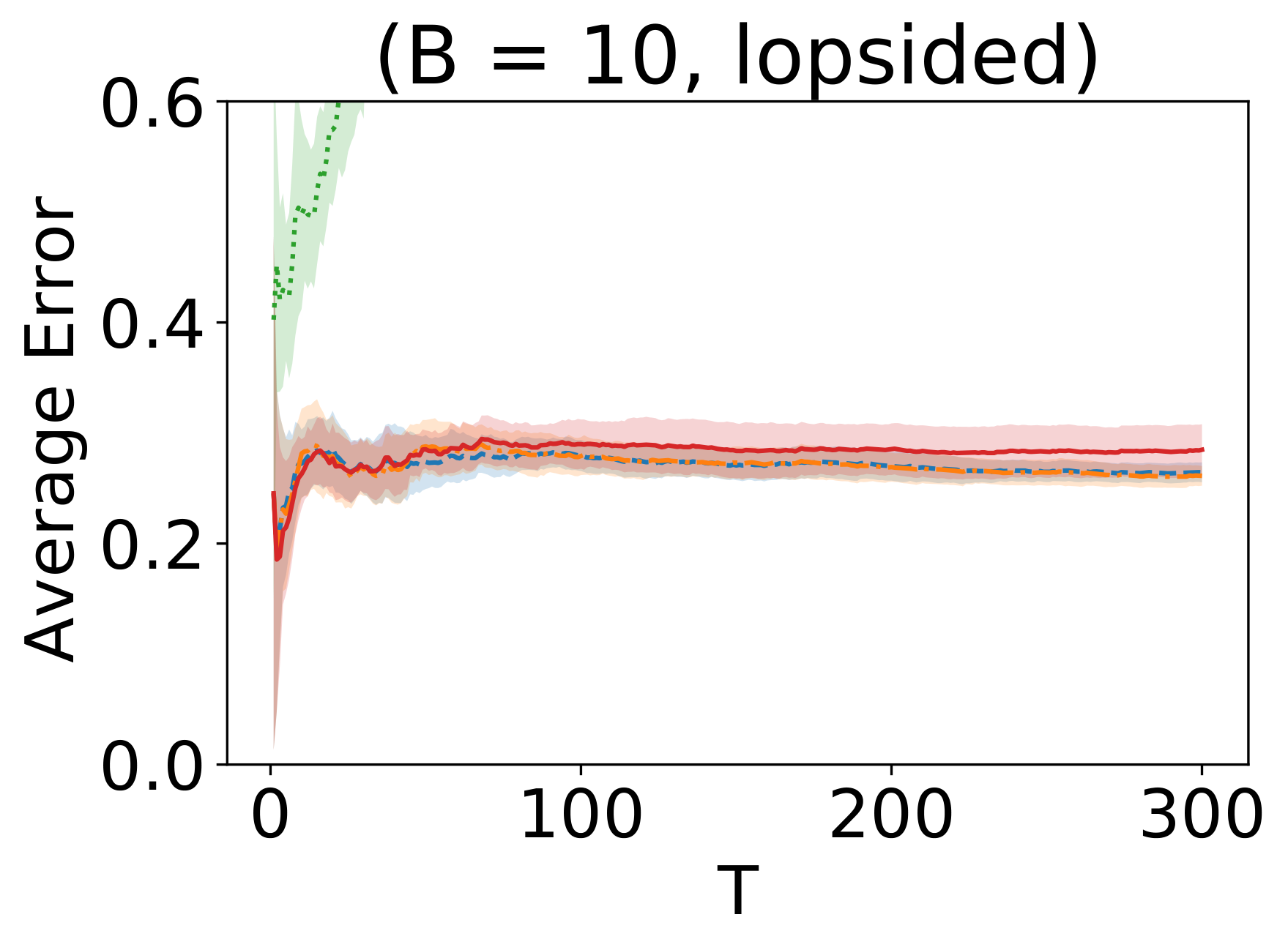} &
     \includegraphics[width = 0.23\textwidth]{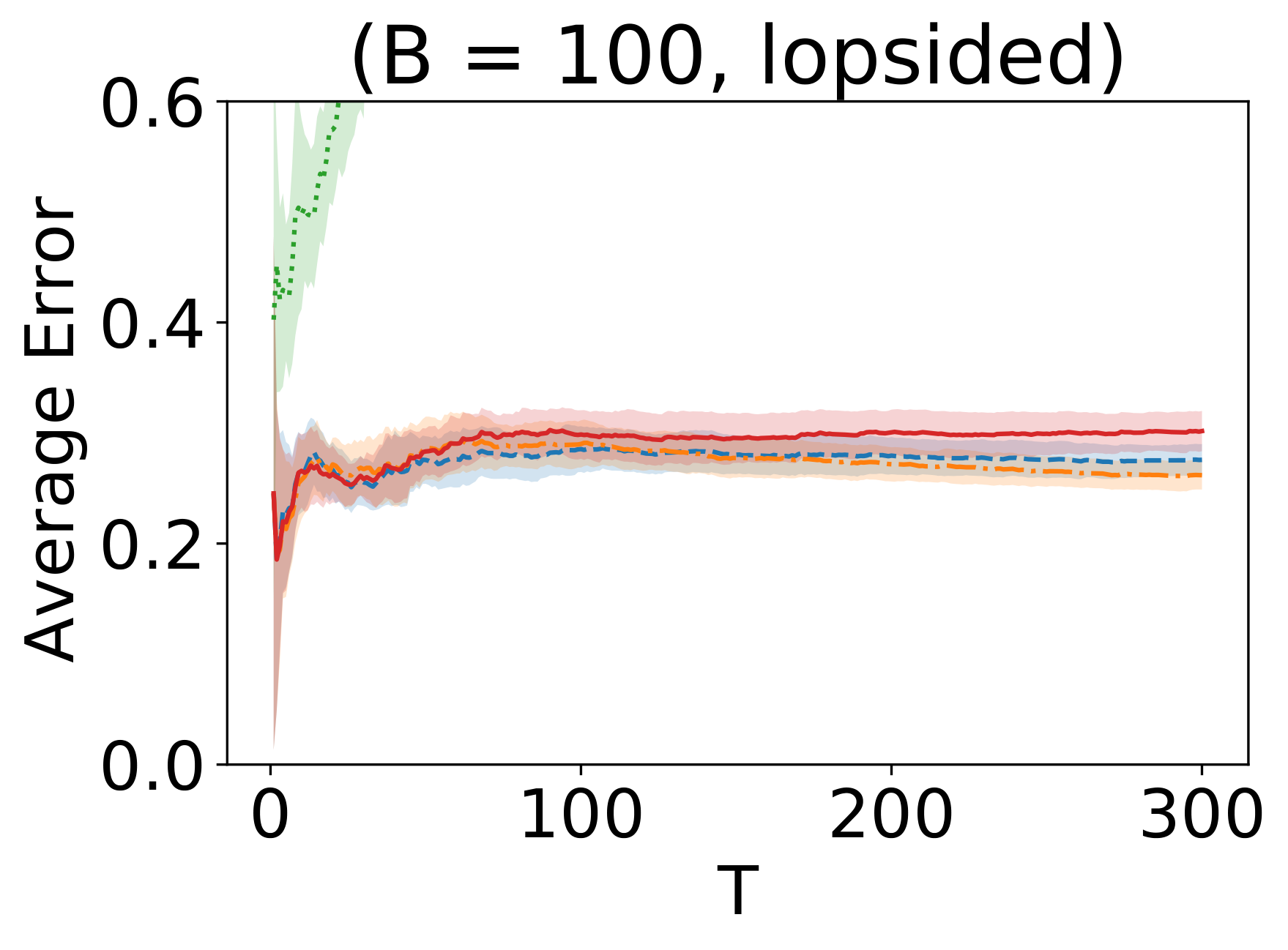}
\end{tabular}
\vspace{-0.1in}
\caption{\small  Error plots comparing the four methods for synthetic functions in~\S\ref{sec:expsynsmooth}.
Columns 1 and 2 show the results
when the arrival pattern is uniform for the Branin function. Columns 3 and 4 show the results when the
arrival pattern is lopsided for the Hartmann function. The results are averaged over 10 trials, and we
report the $95\%$ confidence interval of the standard error. The results of Algorithm~\ref{alg:mab} are poor for columns 3 and 4, so we limit the $y$-axis to focus on the rest methods.}
\label{error:gp_synthetic}
\end{figure}

\begin{figure}[ht!]
\centering
\begin{tabular}{ccc}
     \includegraphics[width = 0.31\textwidth]{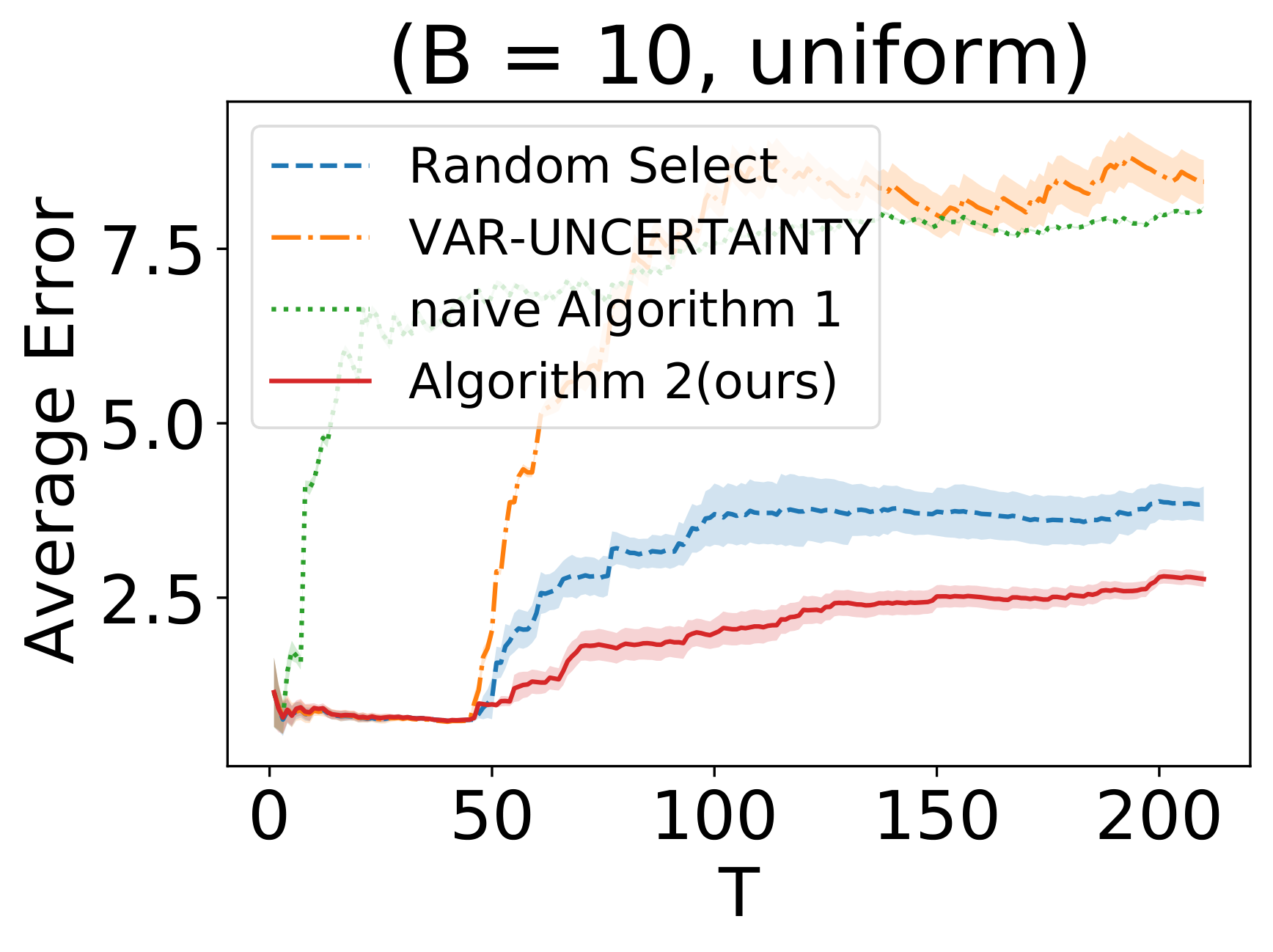} &
     \includegraphics[width = 0.31\textwidth]{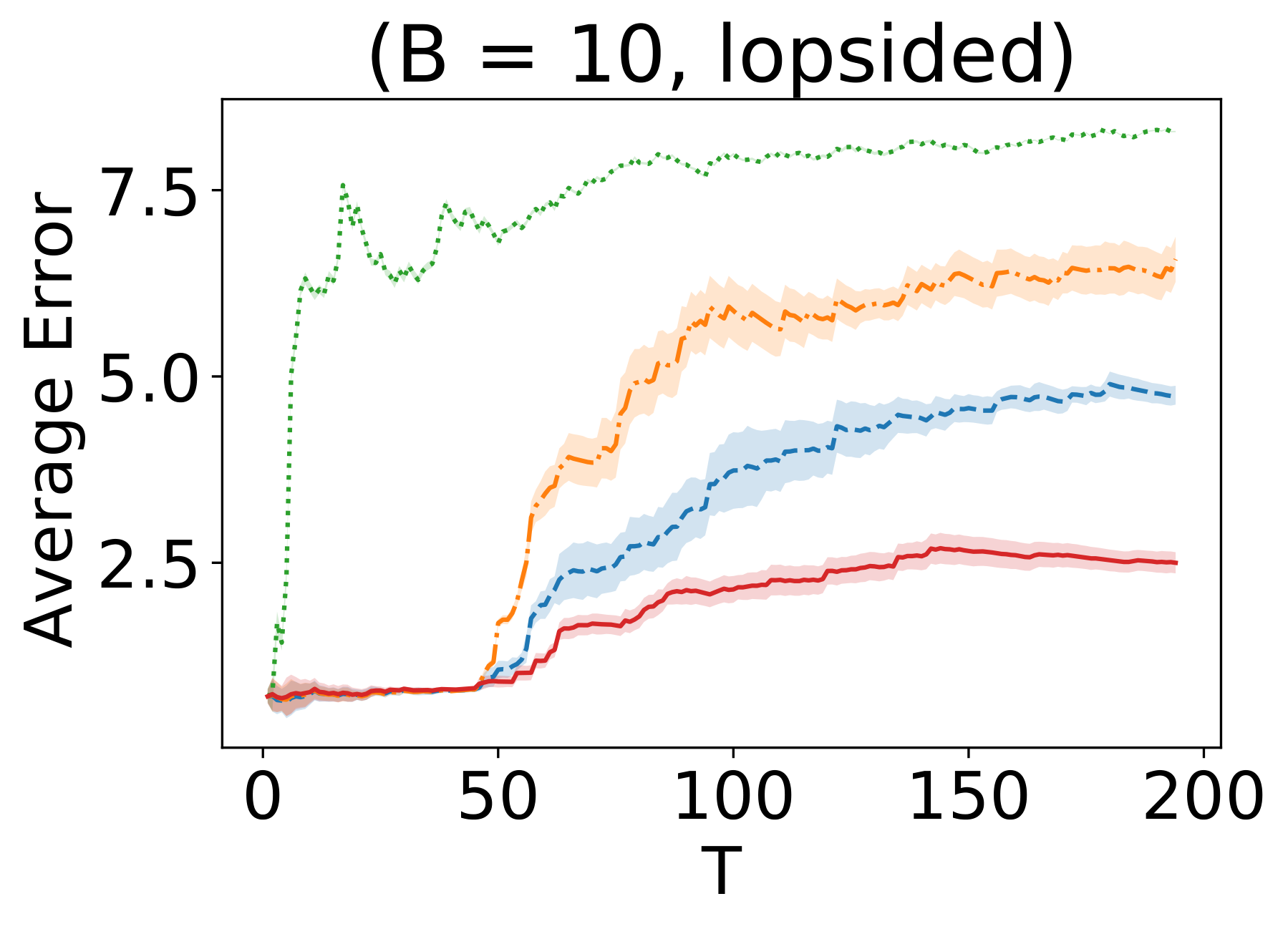} &
     \includegraphics[width = 0.31\textwidth]{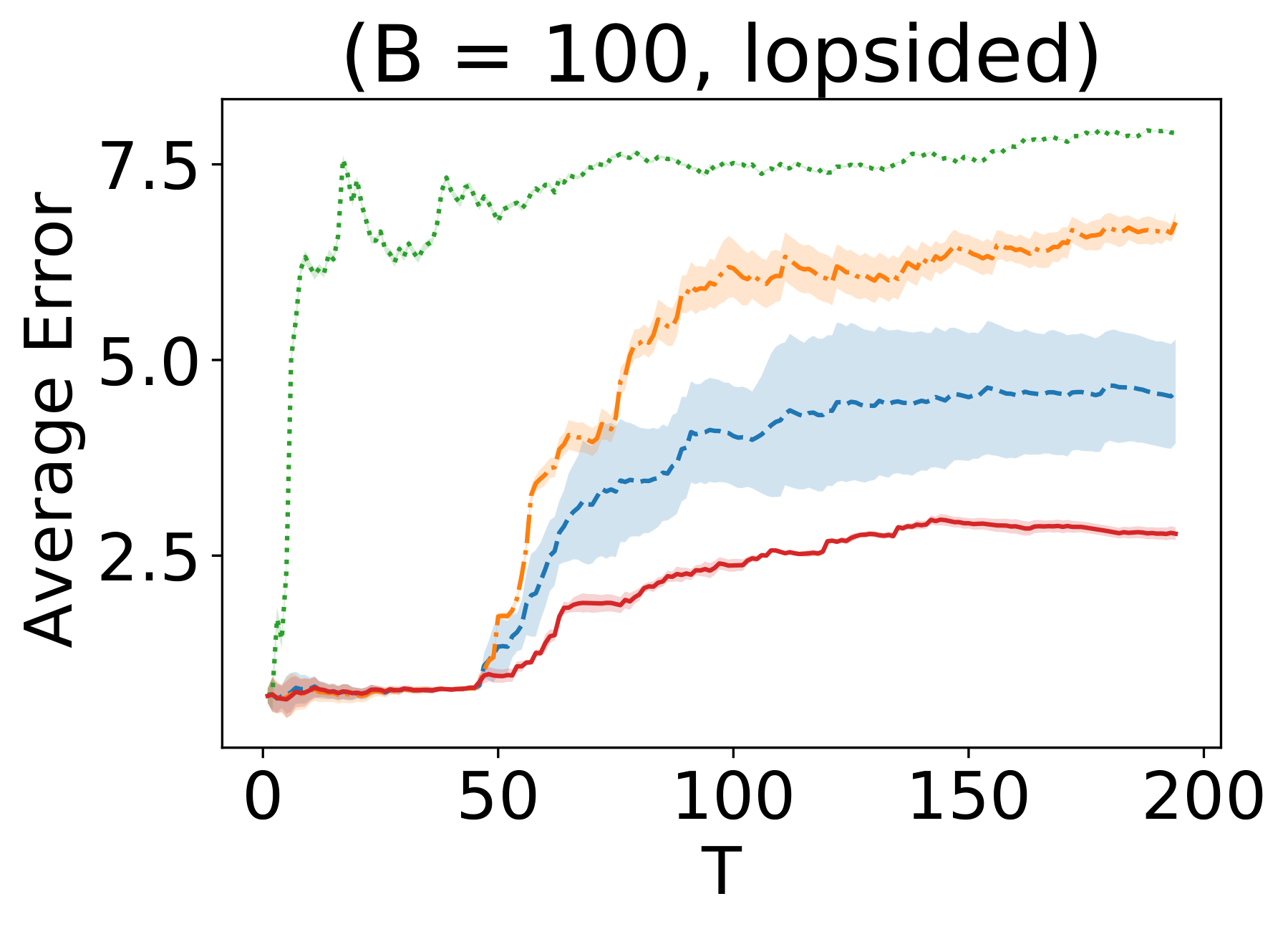}
\end{tabular}
\vspace{-0.1in}
\caption{\small Error plots  comparing the four methods for Parkinsons dataset. The results are averaged over 5 trials and we report the $95\%$ confidence interval of the standard error.}
\label{error:parkinsons}
\end{figure}

\begin{figure}[ht!]
\centering
\begin{tabular}{ccc}
     \includegraphics[width = 0.31\textwidth]{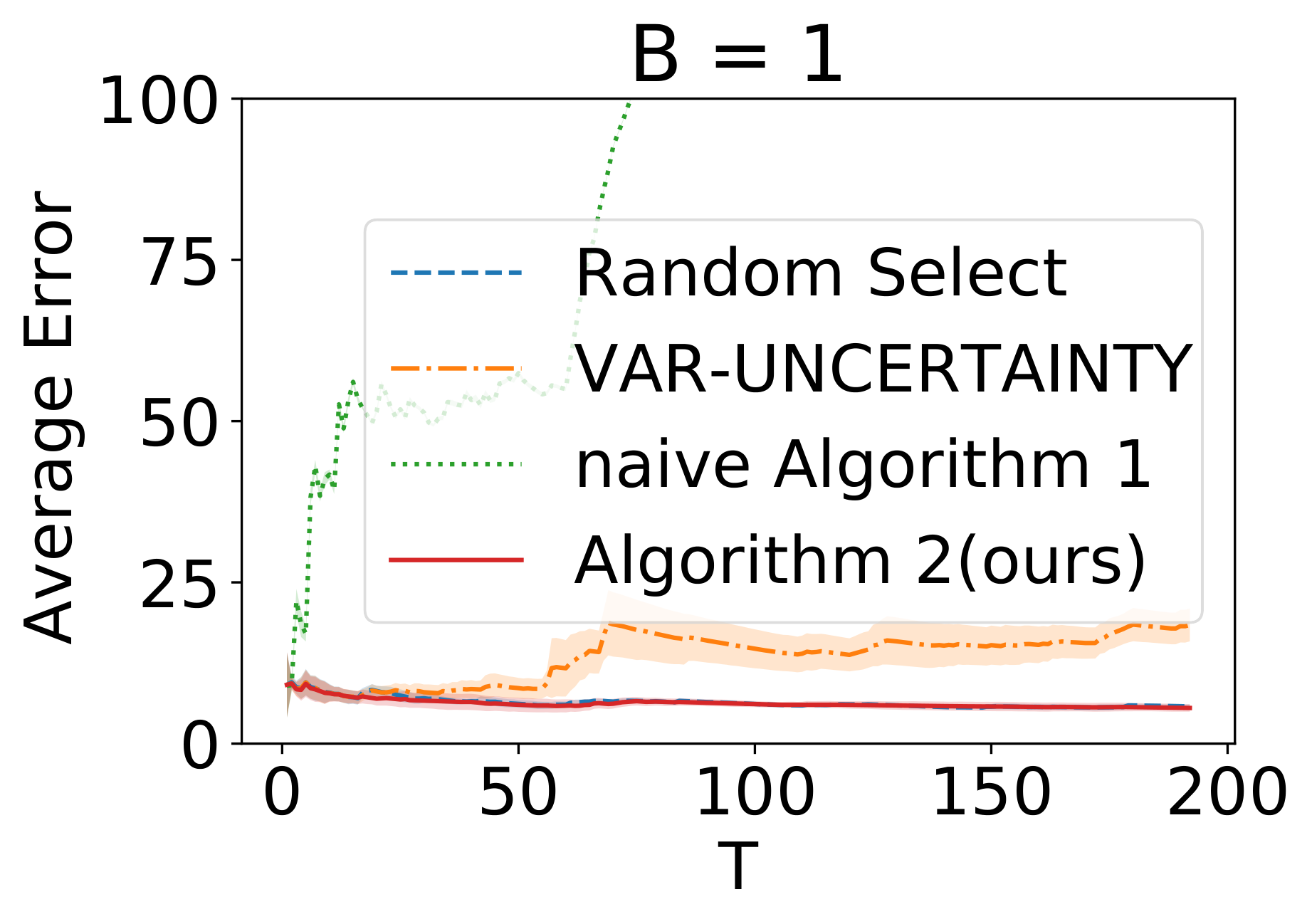} &
     \includegraphics[width = 0.31\textwidth]{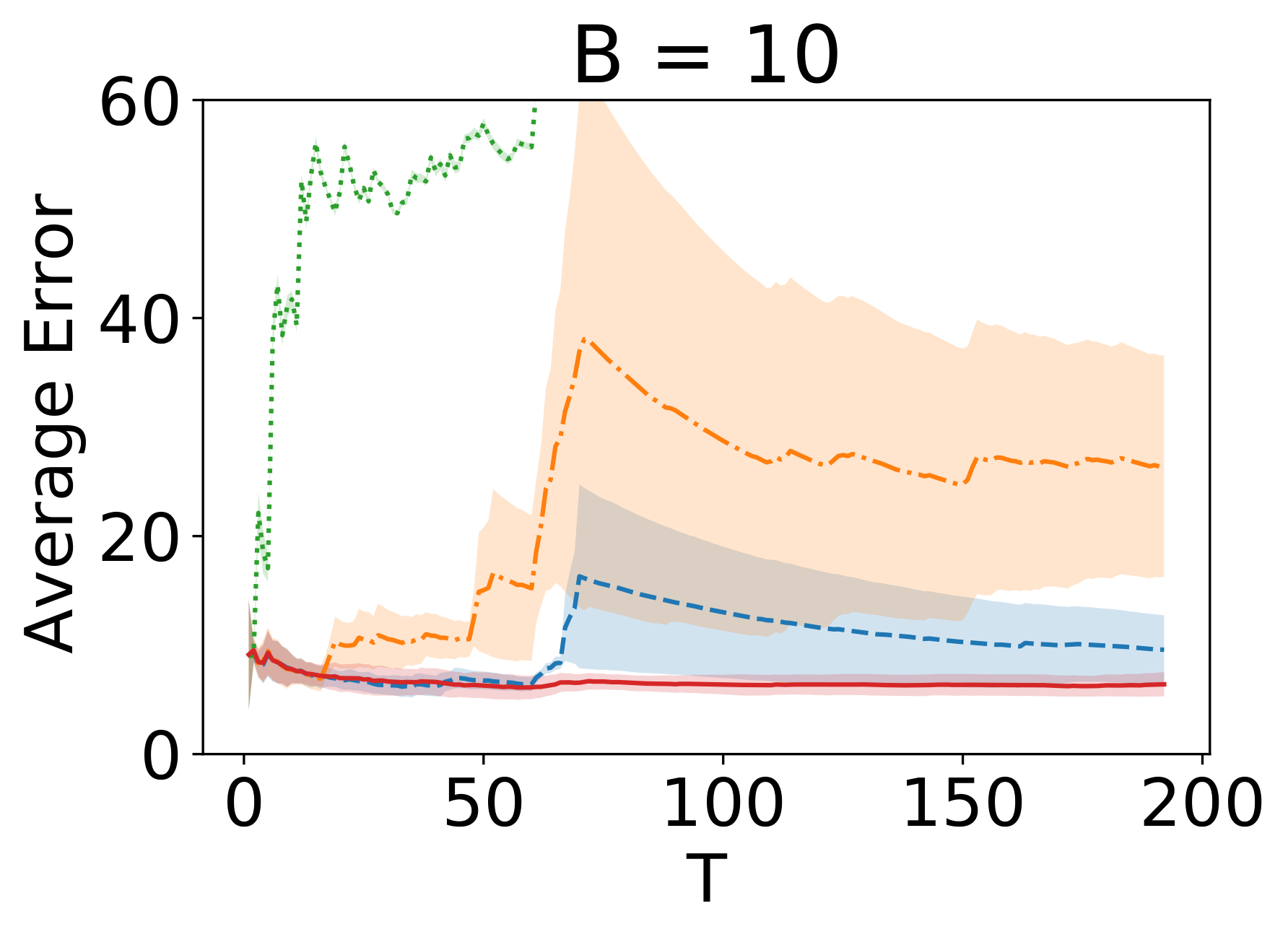} &
     \includegraphics[width = 0.31\textwidth]{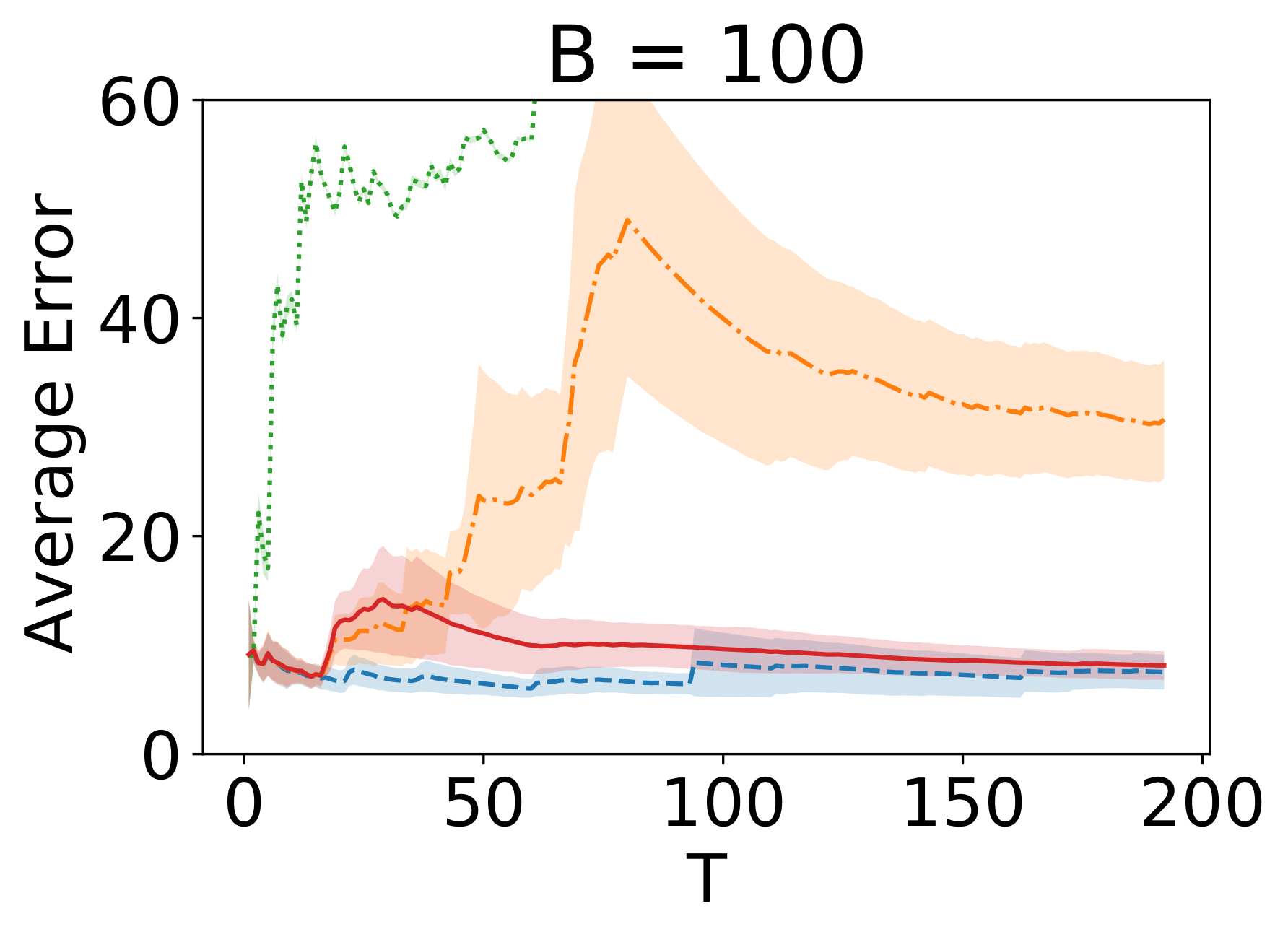}
\end{tabular}
\vspace{-0.1in}
\caption{\small
Error plots comparing the four methods for the supernova dataset. The results are averaged over 5 trials and
we report the $95\%$ confidence interval of the standard error.
The naive Algorithm~\ref{alg:mab} performs poorly so we
limit the $y$-axis to focus on the rest methods.}
\label{error:supernova}
\end{figure}

 The prediction error is defined as the absolute difference between the true function values of $x_t$ and our prediction, which is  $|f(x_t) - p_t|$ at each round $t$. 
 Figures~\ref{error:discrete}, \ref{error:gp_synthetic}, \ref{error:parkinsons}, \ref{error:supernova} provide the average prediction error plots for all the experiments. 
Across all the experiments, Algorithm~\ref{alg:mab} maintains a low prediction error compared to the baselines in the $K$ discrete setting and Algorithm~\ref{alg:gpucb} maintains a low prediction error compared to the baselines in the continuous setting. 
 
 The prediction errors for the naive implementation of Algorithm~\ref{alg:mab} are relatively large compared to other methods in the continuous case, which are normal since we expect the naive prediction by sample mean in the continuous case to perform badly. However, when $B = 100$ for the Parkinsons dataset, naive Algortihm~\ref{alg:mab} has the lowest average loss compared to the other three methods as shown in Figure~\ref{fig:parkinsons}. This indicates that when the cost of experimentation is relatively large and we can tolerate large prediction error, we can choose naive Algorithm~\ref{alg:mab} in the continuous case to reduce the total cost. 
 
 For the Parkinsons dataset, the prediction errors for Algorithm~\ref{alg:gpucb}, \emph{Random Select}, and \emph{VAR-UNCERTAINTY} stay low at the beginning and then increase over time, this is because we allow the three methods to label at the first $5d$ rounds, where $d$ is the dimension of the input data, so they have a low prediction error. Then after this initialization period, for data points that are not labeled but have labels very distinct from the predicted labels, the prediction errors can be large.

\newpage
\section{Ablation study} \label{sup:sec_ablation}
In this section, we compare the performance of Algorithm~\ref{alg:mab} and Algorithm~\ref{alg:gpucb} under different choices of $\lambda$.

\begin{figure}[ht!]
    \centering
    \includegraphics[width = 0.5\textwidth]{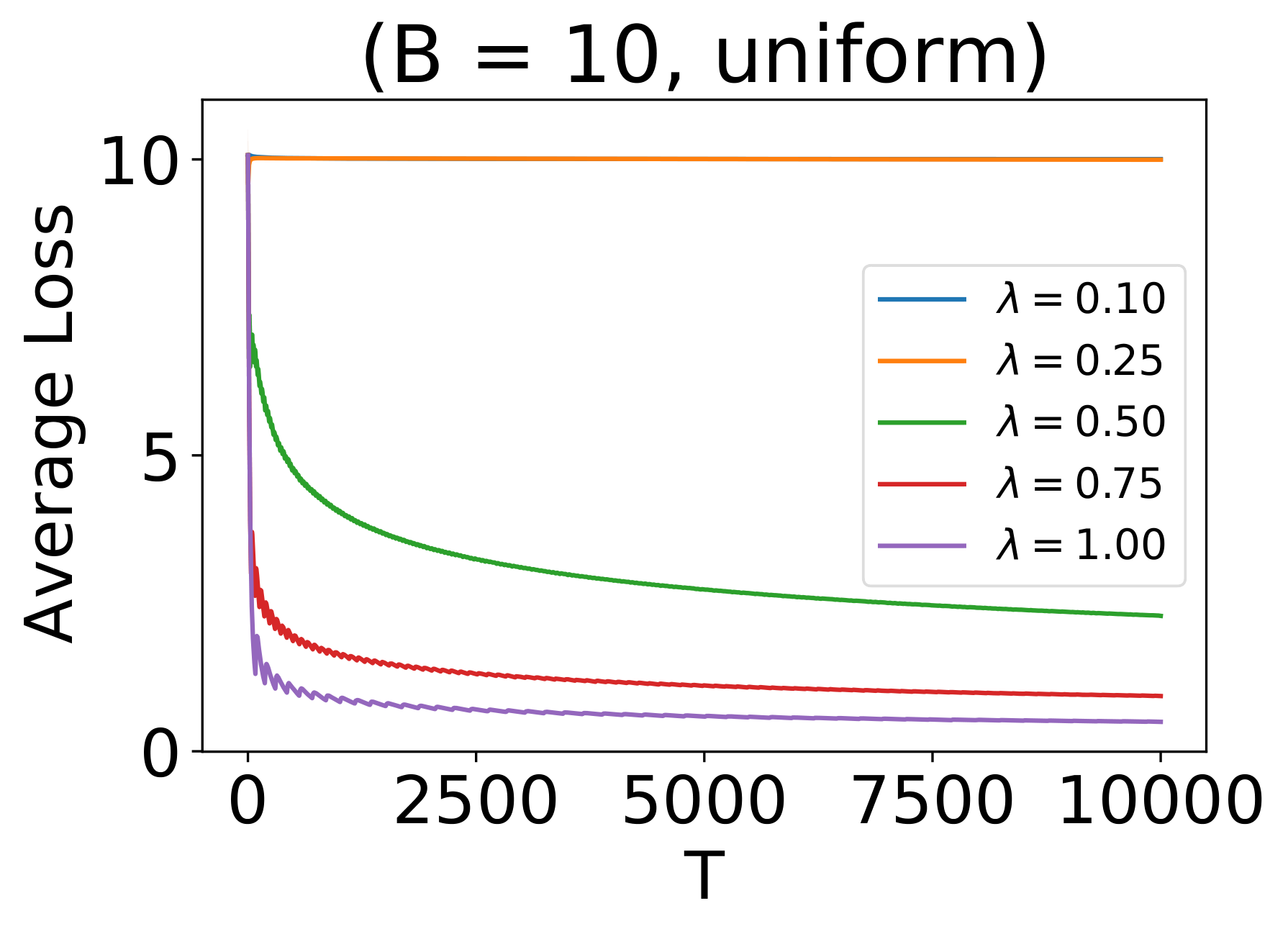}
    \caption{Average loss plots comparing the effects of different $\lambda$ for Algorithm~\ref{alg:mab} when $K = 10, B = 10$ for uniform arrival patterns. The result for $\lambda = 0.1$ is hidden by the result for $\lambda = 0.25$ since a very small value of $\lambda$ prohibits labeling at all. The results are averaged over $10$ trials and we report the $95\%$ confidence interval.}
    \label{fig:ablation_discrete}
\end{figure}

Figure~\ref{fig:ablation_discrete} shows the results of average loss with different values of $\lambda$ for the experiment when $K = 10, B = 10$ and the arrival pattern is uniform. Increasing the value of $\lambda$ can decrease the average loss over time but it may increase the average prediction error since we may not get enough labeling data to do the prediction. 

\begin{figure}[ht!]
    \centering
    \includegraphics[width=0.5\textwidth]{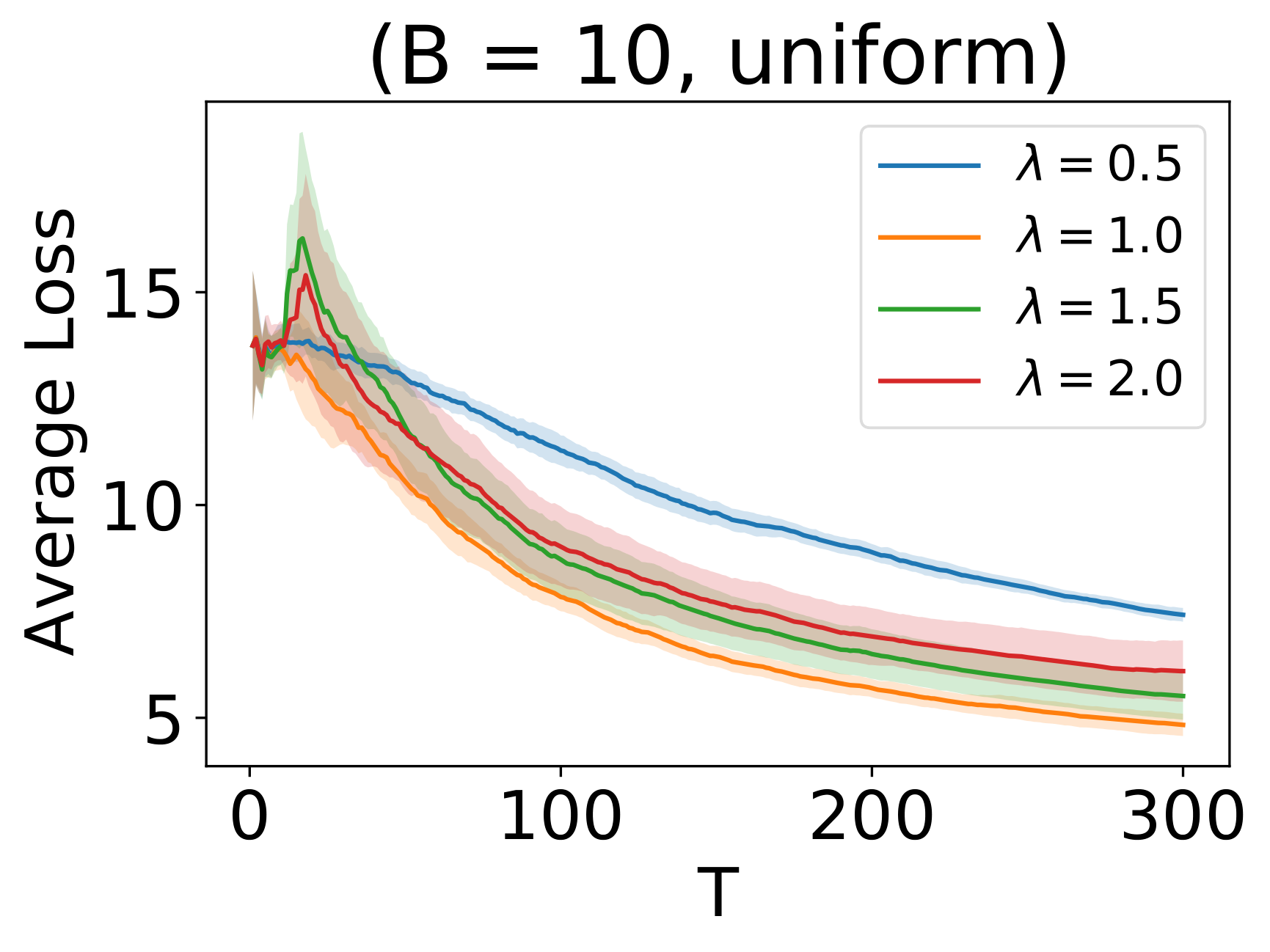}
    \caption{Average loss plots comparing the effects of different $\lambda$ for Algorithm~\ref{alg:gpucb} when $B = 10$ and the arrival pattern is uniform for Branin function. The results are averaged over $10$ trials and we report the $95\%$ confidence interval.}
    \label{fig:abaltion_branin}
\end{figure}

Figure~\ref{fig:abaltion_branin} shows the results of average loss with different values of $\lambda$ when $B = 10$ and the arrival pattern is uniform for the Branin function. $\lambda = 1$ achieves the lowest average loss over time since $\lambda$ being small ($\lambda = 0.5$) limits the number of labeling thus the prediction error is high and $\lambda$ being large ($\lambda \ge 1.5$) labels a lot thus the total cost of labeling  is high.

\newpage
\section{Supplementary tools}
\begin{theorem}[Power Means inequality~\cite{wainwright2019high}] \label{thm:powermean}
For any $p < q$, the following inequality holds
\begin{align*}
    \left (\sum_{i=1}^n w_i x_i^p\right )^{1/p} \le \left (\sum_{i=1}^n w_ix_i^q\right )^{1/q}
\end{align*}
where $w_i \in [0,1]$ and $\sum_{i=1}^n w_i = 1$.
\end{theorem}

\begin{theorem}[Le Cam's Method~\cite{le1960approximation}] \label{thm:lecam}
For any distributions $P_1$ and $P_2$ on $\mathcal{X}$, we have
\[
    \inf_{\Psi} \{P_1(\Psi(X)\ne 1) + P_2 (\Psi(X) \ne 2)\} = 1-\|P_1-P_2\|_{TV},
\]
where the infimum is taken over all tests $\Psi:\mathcal{X}\to\{1,2\}$.
\end{theorem}

\begin{lemma}[Lemma 12 in~\cite{DBLP:journals/corr/KandasamyDOSP16}] \label{lemma}
Let $f \sim \calG \calP(\mathbf{0}, \kappa)$, $f:\calX \to \bbR$ and we observe $y = f(x)+\epsilon$ where $\epsilon \sim \calN(0, \eta^2)$. Let $A\in \calX$ such that its $L_2$ diameter $\mathrm{diam}(A) \le D$. Say we have $n$ queries $(x_t)_{t=1}^n$ of which $s$ points are in $A$. Then the posterior variance of the GP, $\kappa'(x,x)$ at any $x \in A$ satisfies
\begin{equation*}
    \kappa'(x,x) \le 
    \begin{cases}
    D^2/l^2 + \frac{\eta^2}{s} & \mbox{if }\kappa \mbox{ is the SE kernel,} \\
    2L_{\rmM}D + \frac{\eta^2}{s} & \mbox{if }\kappa \mbox{ is the Mat\'ern kernel.}
    \end{cases}
\end{equation*}
Here,  $l$ is the length scale hyperparameter of the SE kernel~\eqref{eqn:kerneldefns},
and $L_{\rm M}$ is a Lipschitz constant of the Mat\'ern kernel.
\end{lemma}

\end{document}


\onecolumn

\section{PROOFS} \label{ax:proofs}
\textit{In this section, we present the detailed proofs of the Theorems and Corollaries in the main paper}.
\subsection{Proof of Theorem~\ref{thm:mab}}

\begin{proof}

Consider the evaluated data and not-evaluated data separately,
\begin{align} \label{eqn:1}
    \nonumber \mathbb{E}L_T &= \sum_{k=1}^K \mathbb{E}\bigg (B \cdot N_{T,k} + \sum_{t=1}^{M_{T,k}} \big|\mu_k - \hat{\mu}_{M_{t,k}, k} \big| \bigg ) \\
    &= \sum_{k=1}^K \mathbb{E}\bigg (B \cdot N_{T,k} + \sum_{\mbox{evaluate}} \big|\mu_k - \hat{\mu}_{M_{t,k}, k} \big| + \sum_{\mbox{not evaluate}} \big|\mu_k - \hat{\mu}_{M_{t,k}, k} \big| \bigg ) 
\end{align}

For any $\delta \in (0,1)$, let $\delta_t = \delta/t^3$.  Define event:
\begin{align*}
    \calE_t = \bigg\{ \hat{\mu}_{M_{t,k},k} - \sqrt{\frac{2\log(2/\delta_{M_{t,k}})}{\max{(N_{t-1,k}},1)}} \le \mu_k \le \hat{\mu}_{M_{t,k},k} + \sqrt{\frac{2\log(2/\delta_{M_{t,k}})}{\max{(N_{t-1,k}},1)}} \quad \forall k \in \calK  \bigg\}
\end{align*}

By Chernoff corollary~\ref{thm:chernoff}, $\bbP \left [|\frac{1}{n}\sum_{i=1}^n y_i - \mu_k| \le \sqrt{\frac{2\log(2/\delta_{M_{t,k}})}{n}}\right] \ge 1-\delta_{M_{t,k}}$. Since $N_{t-1,k}$ is a random variable, union bound on $N_{t-1,k} \in \{1, \cdots, M_{t,k} - 1\}$. We get $\bbP\left[ |\hat{\mu}_{M_{t,k},k}-\mu_k| \le \sqrt{\frac{2\log(2/\delta_{M_{t,k}})}{\max(N_{t-1,k},1)}} \right] \ge 1-M_{t,k}\delta_{M_{t,k}}$. Thus $\bbP(\calE_t) \ge 1-Kt\delta_{t}$. Then  event $\calE = \cap_{t=1}^\infty \calE_t$ holds with probability at least $1-K\pi^2\delta/3$.

Conditioned on event $G$ and by assumption that $\forall k$ $\mu_k \in [0,1]$, \ref{eqn:1} is upper bounded by
\begin{align*}
    \bbE L_T &= \bbE[L_T|\calE]\bbP(\calE) + \bbE[L_T|\neg\calE]\bbP(\neg\calE) \\
    &\le \bbE[L_T|\calE]\bbP(\calE) + (B+1)T\cdot \frac{K\pi^2\delta}{3} \\
    &= \bbE[L_T|\calE]\bbP(\calE) + (B+1)\cdot \frac{K\pi^2}{3} & \left(\delta = \frac{1}{T}\right)
\end{align*}
Hence,
\begin{align*}
    \bbE [L_T|\calE] &\le \sum_{k=1}^K \bbE \bigg( (B+1)\cdot N_{T,k} + \sum_{\mbox{not evaluate}} \sqrt{\frac{2\log(2/\delta_{M_{t,k}})}{\max(N_{t,k},1)}}\bigg) & \\
    &\le \sum_{k=1}^K \bbE \bigg( (B+1)\cdot N_{T,k} + \sum_{\mbox{not evaluate}} B^{\beta}M_{t,k}^\alpha \bigg) &  \mbox{suppose }\tau(M_{t,k}) = B^{\beta}M_{t,k}^\alpha \\
    &\le \sum_{k=1}^K \bbE \bigg( (B+1)\cdot N_{T,k} + \int_0^{M_{T,k}} B^\beta t^\alpha dt \bigg) \\
    &= \sum_{k=1}^K \bigg( (B+1)\cdot \bbE N_{T,k} + \frac{B^\beta}{\alpha+1}M_{T,k}^{\alpha+1} \bigg)
\end{align*}
To bound $\bbE N_{T,k}$, use the same strategy as~\cite{Audibert+MS:2009} Theorem 2. $\forall u \in \bbN_+$, define $S_k = \{t: N_{t,k} \ge u\}$ for the data in $A_k$
\begin{align*}
    N_{T,k} &\le u + \sum_{t \in S_k} \mathds{1}_{\bigg\{  \sqrt{\frac{2\log(2/\delta_{M_{t,k}})}{N_{t,k}}} > \tau(M_{t,k})\bigg\}} \\
    &\le u + \sum_{t \in S_k} \mathds{1}_{\bigg\{N_{t,k} < \frac{2\log(2/\delta_{M_{t,k}})}{\tau^2(M_{t,k})} \bigg\}} \\
    &\le u + \sum_{t \in S_k} \mathds{1}_{\bigg \{u < \frac{2\log(2/\delta_{M_{t,k}})}{\tau^2(M_{t,k})} \bigg\}}
\end{align*}
Hence,
\begin{align*}
    \bbE N_{T,k} \le u + \sum_{t \in S_k} \bbP\bigg(u < \frac{2\log(2/\delta_{M_{t,k}})}{\tau^2(M_{t,k})} \bigg)
\end{align*}
Let $u = B^{\lambda}M_{t,k}^\rho$, choose $u$ such that $\bbP\bigg(u < \frac{2\log(2/\delta_{M_{t,k}})}{\tau^2(M_{t,k})} \bigg) = \bbP \Big(B^\lambda M_{t,k}^\rho\cdot B^{2\beta}M_{t,k}^{2\alpha} < 2\log(2/\delta_{M_{t,k}})\Big) = 0$. Recall that $\delta_t = \delta/t^3$, $\bbP\Big( B^{2\beta + \lambda}M_{t,k}^{\rho+2\alpha} \gtrsim 2\log(2\delta M_{t,k}^3) \Big) = 1$ when $\rho+2\alpha \ge 0$. Choose $\rho = \alpha + 1$, then $\rho = \frac{2}{3}$ and $\alpha = -\frac{1}{3}$. Choose $2\beta+\lambda \ge 0$ and $\lambda+1 = \beta$, then $\lambda = -\frac{2}{3}$ and $\beta = \frac{1}{3}$. We get
\begin{align*}
    \bbE [L_T|\calE] \le \sum_{k=1}^K (B^{\frac{1}{3}}+1+\frac{3B^{\frac{1}{3}}}{2}) M_{T,k}^{\frac{2}{3}}
\end{align*}
Thus
\begin{align*}
    \bbE L_T \le \sum_{k=1}^K (1+\frac{5}{2}B^{\frac{1}{3}}) M_{T,k}^{\frac{2}{3}} + (B+1)\cdot \frac{K\pi^2}{3}.
\end{align*}
    
\end{proof}

\subsection{Proof of Corollary~\ref{col:worstupper}}
In the same setting of Theorem~\ref{thm:mab}, by Power Mean inequality
\begin{align*}
    \Bigg(\frac{M_{T,1}^{\frac{2}{3}}+\cdots + M_{T,K}^{\frac{2}{3}}}{K}\Bigg)^{\frac{3}{2}} \le \Bigg(\frac{M_{T,1}+\cdots+M_{T,k}}{K}\Bigg) = \frac{T}{K}
\end{align*}
which implies $\sum_{k=1}^K M_{T,k}^{\frac{2}{3}} \le T^{\frac{2}{3}}K^{\frac{1}{3}}$. The equality holds when $M_{T,1} = \cdots=M_{T,K}$. Thus the worst case upper bound is $\bbE L_T = \calO(T^{\frac{2}{3}}K^{\frac{1}{3}})$ when data from each sub-spaces comes uniformly.

\subsection{Proof of Theorem~\ref{thm:mablower}}
Recall that
\begin{align*}
    \mathbb{E}L_T &= \sum_{k=1}^K \mathbb{E}\bigg (B \cdot N_{T,k} + \sum_{t=1}^{M_{T,k}} \big|\mu_k - \hat{\mu}_{M_{t,k}, k} \big| \bigg ).
\end{align*}
For each part $k \in \{1,\cdots,K\}$, suppose there are $n_k$ i.i.d. samples $x_1, \cdots, x_{n_k}$ from $\nu_k = \calN(\mu_k, \sigma^2)$. In order to lower bound the minimax risk of estimating $\mu_k$ by the sample mean $\hat{\mu}_k$ under the semi-metric $\rho  = |\mu_k - \hat{\mu}_k|$. Consider two possible distributions for $\nu_k$: fix $\delta \in [0, \frac{1}{4}]$, $P_{1,k} = \calN(\mu_k'+\delta, \sigma^2)$ and $P_{2,k} = \calN(\mu_k'-\delta, \sigma^2)$ where $\mu_k' \sim \mbox{Uniform}(\frac{1}{4}, \frac{3}{4})$. It's easy to see that $\mu_k'+\delta$ and $\mu_k'-\delta$ are $2\delta$-separated. Suppose we have equal probability to choose between $P_{1,k}$ and $P_{2,k}$, by Proposition [] and Le Cam's Method []\textcolor{blue}{need citation here}: consider all tests $\Psi : \calX \to \{1,2\}$
\begin{align*}
    \inf_{\hat{\mu}_k} \sup_{\nu_k} \bbE |\mu_k - \hat{\mu}_k| &\ge \delta \inf_{\Psi} \bigg\{\frac{1}{2}P_{1,k}(\Psi(x_1, \cdots, x_{n_k}) \neq 1) + \frac{1}{2}P_{2,k}(\Psi(x_1, \cdots, x_{n_k}) \neq 2)\bigg\} \\
    &= \frac{\delta}{2} \Big[1-\|P_{1,k}^{n_k} - P_{2,k}^{n_k}\|_{TV}\Big]
\end{align*}
where $P_{i,k}^{n_k}$ is the product distribution for $i = 1,2$ and   $\|P_{1,k}^{n_k} - P_{2,k}^{n_k}\|_{TV}$ is the total variation distance between $P_{1,k}^{n_k}$ and $P_{2,k}^{n_k}$. By Pinsker's inequality and the chain rule of KL-divergence \textcolor{blue}{need citation here}
\begin{align*}
    \|P_{1,k}^{n_k} - P_{2,k}^{n_k}\|_{TV}^2 \le \frac{1}{2} D_{KL}(P_{1,k}^{n_k}\|P_{2,k}^{n_k}) 
    = \frac{n}{2} D_{KL}(P_{1,k}^{n_k}\|P_{2,k}^{n_k}) 
    = \frac{n}{2} \cdot \frac{(2\delta)^2}{2\sigma^2} = \frac{n\delta^2}{\sigma^2}
\end{align*}
Thus $\|P_{1,k}^{n_k}-P_{2,k}^{n_k}\|_{TV} \le \frac{\sqrt{n_k}\delta}{\sigma}$.  Taking $\delta = \frac{\sigma}{2\sqrt{n_k}}$ guarantees that $\|P_{1,k}^{n_k}-P_{2,k}^{n_k} \|\le \frac{1}{2}$. Thus
\begin{align*}
    \inf_{\hat{\mu}_k} \sup_{\nu_k} \bbE |\mu_k - \hat{\mu}_k| \ge \frac{\delta}{2}(1-\frac{1}{2}) = \frac{\delta}{4} = \frac{\sigma}{8\sqrt{n_k}}.
\end{align*}
Hence
\begin{align*}
    \inf\sup \bbE L_T \ge \sum_{k=1}^K\bbE\bigg(B N_{T,k} + M_{T,k}\cdot\frac{\sigma}{8\sqrt{ N_{T,k}}}\bigg)
\end{align*}
Tuning $N_{T,k}$, let $BN_{T,k} = \frac{\sigma M_{T,k}}{8\sqrt{N_{T,k}}}$, $N_{T,k} = (\frac{\sigma}{8B})^{\frac{2}{3}} M_{T,k}^{\frac{2}{3}}$. Thus we get
\begin{align*}
    \inf\sup\bbE L_T \ge \sum_{k=1}^K \frac{1}{4}\sigma^{\frac{2}{3}}B^{\frac{1}{3}} M_{T,k}^{\frac{2}{3}}.
\end{align*}

\subsection{Proof of Corollary~\ref{col:worstlower}}
The worst case lower bound is attained when $M_{T,i} = \frac{T}{K}$, $\forall i \in \calK$.

\subsection{Proof of Theorem~\ref{thm:gp-anytime}}
\begin{align*}
    L_T \le BN_T + \sqrt{C_1\beta_T\gamma_{N_T}N_T} + \frac{C_2}{\alpha+1}T^{\alpha+1}
\end{align*}
Discretize $\calX$ into $K^d$ hypercubes. To bound $N_{T,k}$, $\forall u \in \bbR_+$
\begin{align*}
    N_{T,k} \le u + \sum_{t \in S_k} \mathds{1}_{\{\beta_t^{1/2}\sigma_{N_t}(x_t) > \tau(t)\}}
\end{align*}

\begin{align*}
    R_T &\le B\cdot (N_T + 1) + \sum_{\mbox{query}} |f(x_t) - \mu_{N_t}(x_t)| + \sum_{\mbox{not query}} |f(x_t) - \mu_{N_t}(x_t)| \\
    &\le B\cdot (N_T+1) + \sqrt{C_1\beta_T\gamma_
    {N_T+1} (N_T+1)} + \sum_{\mbox{not query}} \beta_{t}^{1/2}\sigma_{N_t}(x_t) \\
    &\le B\cdot (N_T+1) + \sqrt{C_1\beta_T\gamma_
    {N_T+1} (N_T+1)} + \sum_{\mbox{not query}}\tau(t) \\
    &\le B \cdot (N_T + 1) + \sqrt{C_1\beta_T\gamma_
    {N_T+1} (N_T+1)} + \frac{C_{2,T}}{\alpha +1} T^{\alpha + 1}
\end{align*}
Discretize $\mathcal{X}$ into $K^d$ equal size hypercubes $A_1, \cdots, A_{K^d}$ where their $L_2$ diameters $\mbox{diam}(A_i)= \frac{\sqrt{d}}{K}$, $\forall i \in [K]$.\\
$N_t$: the number of queries until time $t$\\
$N_{t,k}$: the number of queries until time $t$ in $A_k$, $N_{t,1} + \cdots +N_{t,K} = N_t$\\
$M_{t,k}$: the number of points until time $t$ in $A_k$, $M_{t,1} + \cdots +M_{t,K} = t$\\
To bound $\mathbb{E}N_T$, consider bounding $N_{T,k}$ first. $\forall u \in \mathbb{R}_+$,
\begin{align*}
    N_{T,k} \le u + \sum_{t \in S_k} \mathds{1}_{\{N_{t,k} \ge u,\mbox{ } \beta_t^{1/2}\sigma_{N_t}(x_t) > \tau(t)\}}
\end{align*}
where $S_k = \{t: x_t \in A_k, N_{t,k} \ge u\}$.
\begin{align*}
    \mathbb{E}N_{T,k} \le u + \sum_{t \in S_k} \mathbb{P}(N_{t,k} \ge u, \beta_t^{\frac{1}{2}}\sigma_{N_t}(x_t) > \tau(t) )
\end{align*}
To bound $\mathbb{E}N_{T,k}$ by $u$, choose $u$ such that $\sum_{t \in S_k} \mathbb{P}(N_{t,k} \ge u, \beta_t^{\frac{1}{2}}\sigma_{N_t}(x_t) > \tau(t)) = 0$. In other words, choose $u$ such that $\beta_t^{1/2}\sigma_{N_t}(x_t) \le \tau(t)$
\begin{align*}
    \beta_t\sigma_{N_t}^2(x_t) &\le \beta_t \bigg(\frac{C_{SE}d}{K^2} + \frac{\sigma^2}{N_{t,k}} \bigg) &  \mbox{(Lemma 12 in MF-GP-UCB)} \\ 
    &\le \beta_t \bigg( \frac{C_{SE}d}{K^2} + \frac{\sigma^2}{u}\bigg) & (N_{t,k} \ge u) \\
    &\le 2\beta_t \cdot \frac{\sigma^2}{t^\rho} & (\mbox{let }u = t^\rho \mbox{ and } K = \sqrt{\frac{C_{SE}d}{\sigma^2}} \cdot T^{\frac{\rho}{2}})
\end{align*}
So let $\tau(t) = \frac{\sqrt{2\beta_t \sigma^2}}{t^{\rho/2}} $, which indicates $C_{2,t} = \sqrt{2\beta_t \sigma^2}$ and $\alpha = -\frac{\rho}{2}$. Thus $\mathbb{E}N_{t} \le K^d u = \sqrt{\frac{C_{SE}d}{\sigma^2}} T^{\frac{\rho d}{2}+\rho}$
\begin{align*}
    \mathbb{E}R_T \le B \cdot (\sqrt{\frac{C_{SE}d}{\sigma^2}} T^{\frac{\rho d}{2}+\rho} + 1 ) + \sqrt{C'_1\beta_T\gamma_{N_T +1}N_T+1} + \frac{C_{2,T}}{\alpha + 1} T^{-\frac{\rho}{2}+1}
\end{align*}
Let $\frac{\rho d}{2}+\rho = -\frac{\rho}{2}+1$, $\rho = \frac{2}{d+3}$. Thus $\mathbb{E}R_T = \mathcal{O}(T^\frac{d+2}{d+3})$. \textcolor{gray}{For Matern kernel, $\mathbb{E}R_T = \mathcal{O}(T^{\frac{4}{5}})$}.

\section{Addendum of tools}
\begin{manualtheorem}{A.1}[\textbf{Corollary of Chernoff bound}] \label{thm:chernoff}
Let $X_1, \cdots, X_n$ be i.i.d. Random Variables with common mean $\mu \in [0,1]$. For any $\delta \in (0,1)$,
\begin{align*}
    \bbP \bigg[\frac{1}{n}\sum_{i=1}^n X_i - \sqrt{\frac{2\log(2/\delta)}{n}} \le \mu \le \frac{1}{n}\sum_{i=1}^n X_i + \sqrt{\frac{2\log(2/\delta)}{n}} \bigg] \ge 1-\delta.
\end{align*}
\end{manualtheorem}





\vfill

\bibliographystyle{plain}
\bibliography{biblio}